\documentclass{article}

 \usepackage[nonatbib,final]{neurips_2020}

\usepackage[utf8]{inputenc} 
\usepackage[T1]{fontenc}    
\usepackage{hyperref}       
\usepackage{url}            
\usepackage{booktabs}       
\usepackage{amsfonts}       
\usepackage{nicefrac}       
\usepackage{microtype}      

\usepackage{url}
\usepackage{algorithm}
\usepackage{algorithmic}
\usepackage{amsmath}
\usepackage{amsfonts}
\usepackage{graphics}
\usepackage{epsfig}
\usepackage{subfig}
\usepackage{float}
\usepackage{multirow}
\usepackage{appendix}
\usepackage{color}
\usepackage{times}
\usepackage{helvet}
\usepackage{courier}
\usepackage{multirow}
\usepackage{bm}
\usepackage[american]{babel}
\usepackage{enumitem}
\usepackage{caption}
\usepackage{diagbox}

\def\X{\bm{X}}

\def \h{{\bf h}}

\def \w{{\bf w}}

\def \y{{\bf y}}

\def \b{{\bf b}}

\def \W{{\bf W}}

\def \H{{\bf H}}
\def \X{{\bf X}}
\def \E{{\bf E}}

\def \A{{\bf A}}

\def \0{{\bf 0}}

\title{DynaBERT: Dynamic BERT with Adaptive Width and Depth}

\author{Lu Hou$^1$, Zhiqi Huang$^{2}$, Lifeng Shang$^1$, Xin Jiang$^1$, Xiao Chen$^1$, Qun Liu$^1$ \\
	$^1$Huawei Noah's Ark Lab \\
	\texttt{\{houlu3,shang.lifeng,Jiang.Xin,chen.xiao2,qun.liu\}@huawei.com}\\
	$^2$Peking University, China\\
	\texttt{zhiqihuang@pku.edu.cn}\\
}

\begin{document}
	
	\maketitle
	
	\begin{abstract}
		The pre-trained language models like BERT, though powerful in many natural language processing tasks, are both computation and memory expensive.
		To alleviate this problem, one approach is to compress them for specific tasks before deployment.
		However, recent works on BERT compression usually compress the large BERT model to a fixed smaller size. They can not fully satisfy the requirements of different edge devices with various hardware performances.
		In this paper, we propose a novel dynamic BERT model~(abbreviated as DynaBERT), which can flexibly adjust the size and latency by selecting adaptive width and depth.
		The training process of DynaBERT includes first training a  width-adaptive BERT and then allowing both adaptive width and depth, by distilling knowledge from 
		the full-sized model to small sub-networks.
		Network rewiring is also used to  keep the more important attention heads and neurons  shared by more sub-networks.
		Comprehensive experiments under various efficiency constraints demonstrate that our proposed
		dynamic BERT (or RoBERTa) at its largest size has comparable performance as $\text{BERT}_\text{BASE}$ (or  $\text{RoBERTa}_\text{BASE}$), while 
		at smaller widths and depths consistently outperforms existing BERT compression methods. Code is available at \url{https://github.com/huawei-noah/Pretrained-Language-Model/tree/master/DynaBERT}.
		
	\end{abstract}
	\section{Introduction}
	Recently, pre-trained language models based on the Transformer~\cite{vaswani2017attention} structure like BERT~\cite{devlin2019bert} and RoBERTa~\cite{liu2019roberta} have achieved  remarkable results on  natural language processing tasks. However, these models have many parameters,  hindering their deployment on edge devices with limited storage, computation, and energy consumption. 
	The difficulty of deploying BERT to these devices lies in two aspects.
	Firstly,  the hardware performances of various devices vary a lot, and it is infeasible to deploy one single BERT model to all kinds of edge devices. Thus different architectural configurations of the BERT model are desired.
	Secondly, the resource condition of one device under different circumstances can be quite different. For instance, on a mobile phone, when a large number of compute-intensive or storage-intensive programs are running, the resources that can be allocated to the current BERT model will be correspondingly fewer. Thus once the BERT model is deployed, dynamically selecting a part of the model (also referred to as \textit{sub-networks}) for inference based on the device's current resource condition is also desirable. Note that unless otherwise specified, the BERT model mentioned in this paper refers to a task-specific BERT rather than the pre-trained model.

	There have been some attempts to compress and accelerate inference of the Transformer-based models using low-rank approximation~\cite{ma2019tensorized,lan2020ALBERT}, weight-sharing~\cite{dehghani2018universal,lan2020ALBERT}, knowledge distillation~\cite{sanh2019distilbert,sun2019patient,jiao2019tinybert,wang2020minilm}, quantization~\cite{bhandare2019efficient,zafrir2019q8bert,shen2019q,fan2020training} and pruning~\cite{mccarley2019pruning,michel2019sixteen,cui2019fine,voita2019analyzing}.
	However, these methods usually compress the model to a fixed size and can not meet the requirements above. 
	In~\cite{dehghani2018universal,fan2019reducing,elbayad2020depthadaptive,liu2020fastbert,xin2020deebert,zhou2020bert},  Transformer-based models with adaptive depth are proposed to
	dynamically select some of the Transformer layers during inference.
	However, these models only consider compression in the depth direction  and generate a limited number of architectural configurations, which can be restrictive for various deployment requirements. Some studies now show that the width direction also has high redundancy. For example, in \cite{voita2019analyzing,michel2019sixteen}, it is shown that only a small number of attention heads are required to keep comparable accuracy.
	There have been some works that train convolutional neural networks (CNNs) with adaptive width~\cite{yu2018slimmable,yu2019universally,yu2019network}, 
	and even both adaptive width and depth~\cite{Cai2020Once}.
	However,
	since each Transformer layer in the BERT model includes both a Multi-Head Attention (MHA) module and a position-wise Feed-forward Network (FFN) that perform transformations in two different dimensions
	(i.e., the sequence and the feature dimensions),  
	the width of the BERT model can not be simply defined as the number of kernels as in CNNs. Moreover, successive training of first along depth and then width as in \cite{Cai2020Once} can be sub-optimal since these two directions are hard to disentangle. This may also cause the knowledge learned in the depth direction to be forgotten after the width is trained to be adaptive.

	In this work, we propose a novel DynaBERT model that offers flexibility in both width and depth directions of the BERT model.
	Compared to~\cite{dehghani2018universal,fan2019reducing,elbayad2020depthadaptive,liu2020fastbert} where only depth is adaptive, DynaBERT enables a significantly richer number of architectural configurations and better exploration of the balance between model accuracy and size. 
	Concurrently to our work, flexibility in both directions 
		is also proposed in \cite{hanruiwang2020hat}, but on the encoder-decoder Transformer structure~\cite{vaswani2017attention} and on machine translation task.
	Besides the difference in the model and task,
 our proposed DynaBERT 
also advances in the following aspects: 
		(1) We 
		 distill knowledge from the full-sized teacher model to smaller student sub-networks
		to reduce the accuracy drop caused by the lower capacity of smaller size.
	(2)  
Before allowing both  adaptive width and depth, 
	we train an only width-adaptive BERT (abbreviated as $\text{DynaBERT}_\text{W}$) to act as a teacher assistant to bridge the large gap of model size between the student and teacher.
	(3) For $\text{DynaBERT}_\text{W}$, we rewire the connections in each Transformer layer 
	to ensure that the more important heads and neurons are utilized by more sub-networks.
	(4) Once DynaBERT is trained, no further fine-tuning is required for each sub-network.
Extensive experiments  
on the GLUE benchmark and SQuAD under various efficiency constraints
 show that,
our proposed
dynamic BERT (or RoBERTa) at its largest size performs comparably as $\text{BERT}_\text{BASE}$ (or $\text{RoBERTa}_\text{BASE}$), while 
at smaller sizes outperforms other BERT compression methods.
	
	\section{Method}
\begin{figure}[htbp]
		\vspace{-0.15in}
		\centering
		\includegraphics[width=0.8\linewidth]{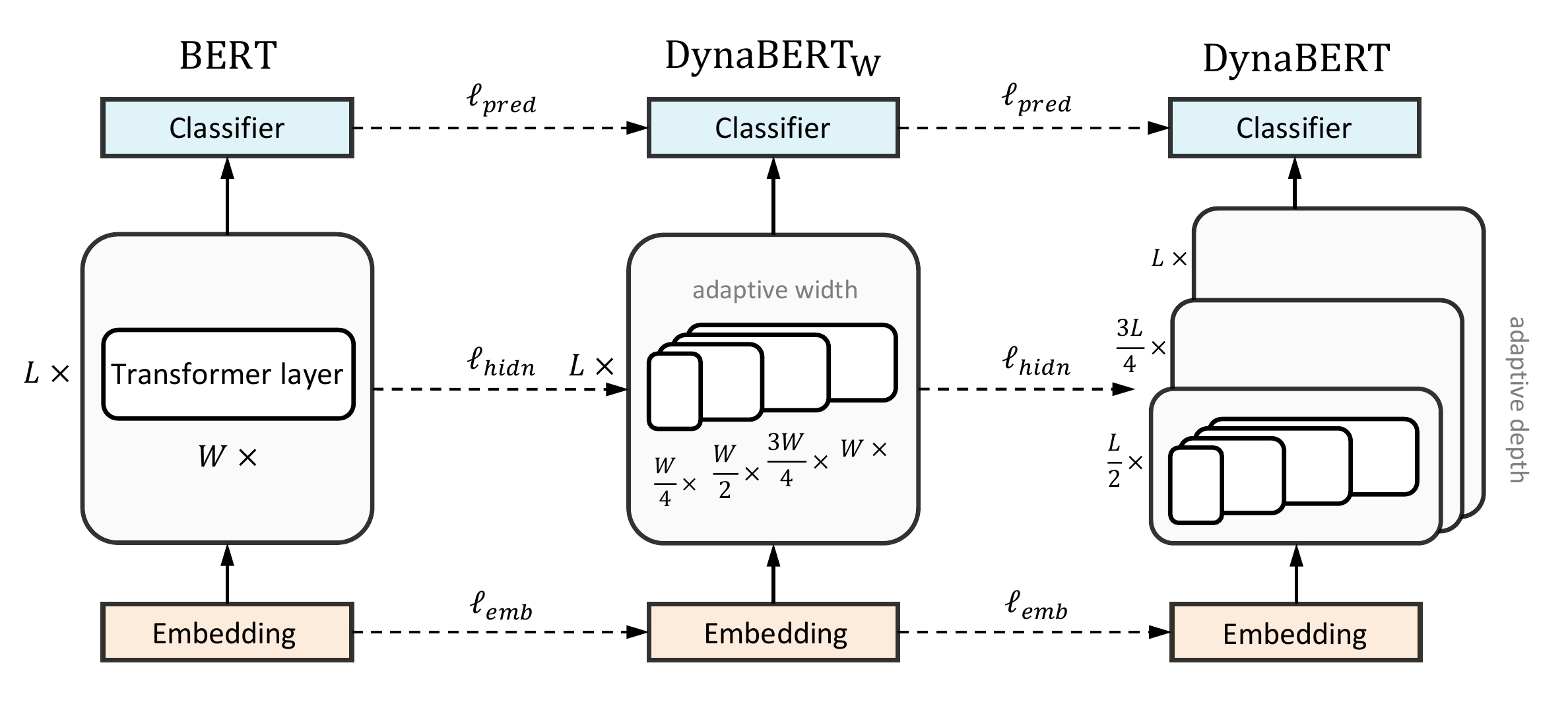}
		\caption{ A two-stage procedure to train $\text{DynaBERT}$. First,
			using knowledge distillation (dashed lines) to transfer the knowledge from a fixed teacher model to student sub-networks with adaptive width in $\text{DynaBERT}_\text{W}$.
			Then,	using knowledge distillation (dashed lines) to transfer the knowledge from a trained $\text{DynaBERT}_\text{W}$ to student sub-networks with adaptive width and depth in $\text{DynaBERT}$.}
		\label{fig:model}
	\end{figure}

	In this section, we elaborate on the training method of our DynaBERT model.
	The training process (Figure~\ref{fig:model}) includes two stages. We first train a width-adaptive  $\text{DynaBERT}_\text{W}$ in Section~\ref{sec:width_adaptive} and then train the both width- and depth-adaptive DynaBERT in Section~\ref{sec:depth_width}.
		Directly using the knowledge distillation to train DynaBERT without $\text{DynaBERT}_\text{W}$, or first train a depth-adaptive BERT and then distill knowledge from it to DynaBERT leads to inferior performance (Details are in Section~\ref{sec:depth_first}).

	\begin{figure*}[htbp]
		\noindent\begin{minipage}{.505\textwidth}
			\centering
			\includegraphics[width=1\linewidth]{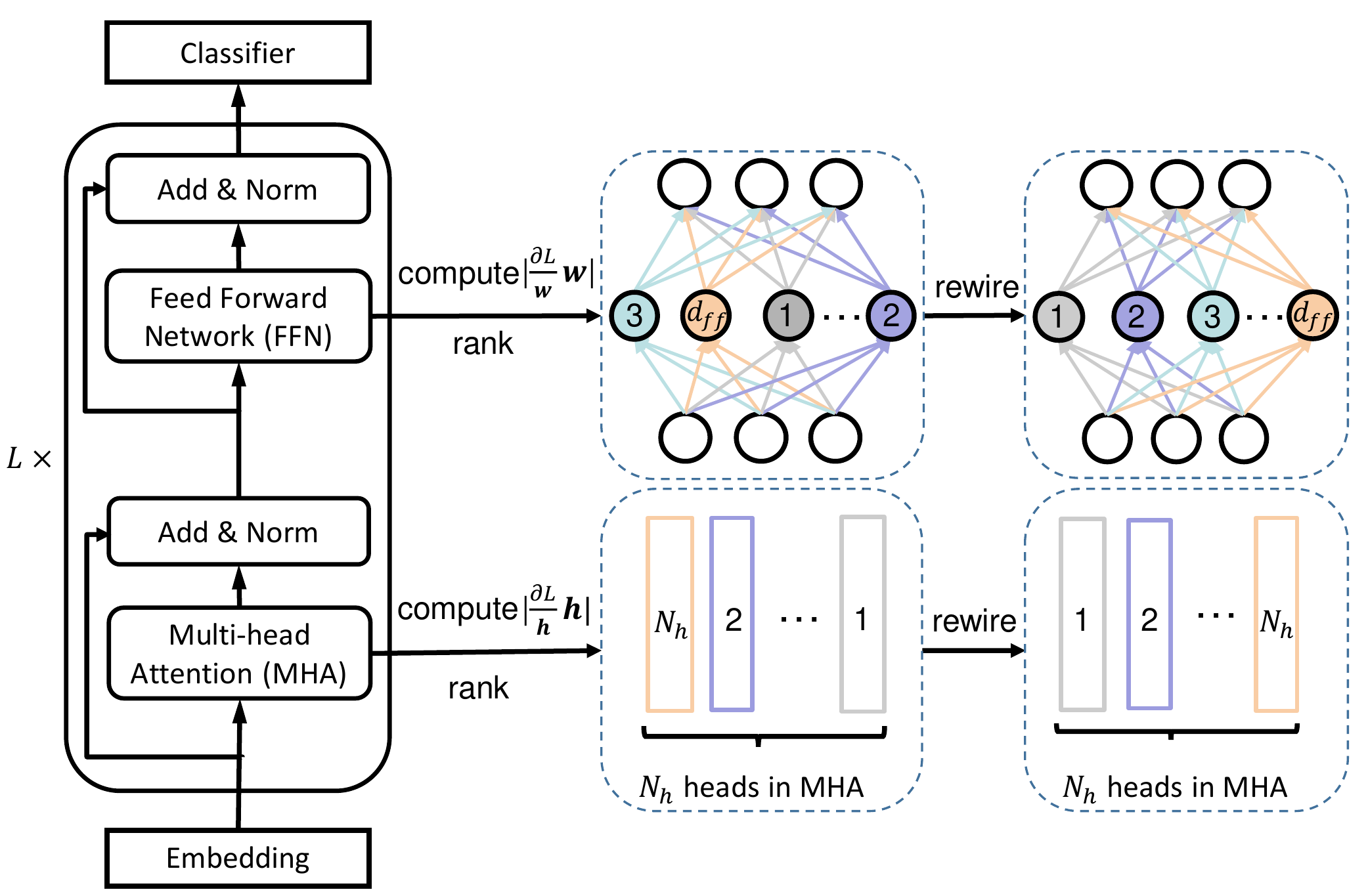}
			\captionof{figure}{Rewire connections in BERT based on the importance of attention heads in MHA and neurons in the intermediate layer of FFN. } \label{fig:rewire}
		\end{minipage}%
		\hspace{0.04in}
		\begin{minipage}{.49\textwidth}
			\vspace{-0.2in}
			\begin{algorithm}[H]
				\caption{Train $\text{DynaBERT}_\text{W}$ or DynaBERT.}\label{alg:adaptive}
				\label{alg:depth_width}
				\small{
					\begin{algorithmic}[1]
						\STATE \textbf{if} training $\text{DynaBERT}_\text{W}$ \textbf{then}
						\STATE \quad $\mathcal{L}\!\!\leftarrow\!\!(\ref{eq:loss})$, $InitM\!\!\leftarrow$rewired net, $depthList\!\!=\!\![1]$.
						\STATE  \textbf{else}:  
						\STATE \quad  $\mathcal{L}\!\!\leftarrow\!\!(\ref{eq:loss1})$, $InitM\!\!\leftarrow\!\!\text{DynaBERT}_\text{W}$.
						
						\STATE {\bf initialize} a fixed teacher model and a trainable student model with  $InitM$.
						\FOR {$iter = 1, \cdots, T_{train} $}
						\STATE Get next mini-batch of training data.
						\STATE Clear gradients  in the student model.
						\STATE  \textbf{for} {$m_d$ in $depthList$}  \textbf{do}
						\STATE \quad \textbf{for} {$m_w$ in $widthList$}  \textbf{do}
						\STATE \quad \quad  		Compute loss  $\mathcal{L}$.
						
						\STATE \quad \quad Accumulate gradient $\mathcal{L}.backward()$.
						\STATE \quad  \textbf{end for}
						\STATE  \textbf{end for} 
						\STATE Update with the accumulated gradients.
						\ENDFOR
					\end{algorithmic}
				}
			\end{algorithm}

		\end{minipage}
	\end{figure*}
	\subsection{Training $\text{DynaBERT}_\text{W}$ with Adaptive Width}
\label{sec:width_adaptive}

	Compared to CNNs stacked with regular convolutional layers, the BERT model is built with Transformer layers, and the width of it can not be trivially determined due to the 
	more complicated computation involved.
	Specifically, a standard Transformer layer contains a Multi-Head Attention (MHA) layer and a Feed-Forward Network (FFN).
 In the following, we rewrite the original formulation  of MHA and FFN in \cite{vaswani2017attention} in a different way to show that, the computation of the attention heads of MHA and the neurons in the intermediate layers of FFN can be performed in parallel.
	Thus we can adjust the width of a Transformer layer by varying the number of attention heads and neurons in the intermediate layer of FFN.
	
	For the $t$-th Transformer layer, suppose the input to it is 
	$\X \in \mathbb{R}^{n \times d}$ where $n$ and $d$ 
	are the sequence length and hidden state size, respectively.
	Following~\cite{michel2019sixteen}, we divide the computation of the MHA into the computations for each attention head.
	suppose there are $N_H$ attention heads in each layer, with head $h$ parameterized by
	$\W^Q_h, \W^K_h, \W^V_h, \W^O_h \in\ \mathbb{R}^{d \times d_h}$ where $d_h = d/N_H$. The 
	output of the head $h$ is computed as
	$
	\text{Attn}_{\W^Q_h, \W^K_h, \W^V_h, \W^O_h}^h(\X) \! =
	\text{Softmax}(\frac{1}{\sqrt{d}}\X \W^Q_h \W^{K\top}_h \X^\top) \X \W^V_h\W^{O\top}_h.
	$
	In multi-head attention, $N_H$ heads are computed in parallel to get the final output~\cite{voita2019analyzing}:
	\begin{equation}
	\label{eq:mha}
	\text{MHAttn}_{\W^Q, \W^K, \W^V, \W^O}(\X) = \sum \nolimits_{h=1}^{N_H} \text{Attn}_{\W^Q_h, \W^K_h, \W^V_h, \W^O_h}^h(\X).
	\end{equation}
		Suppose the two linear layers in FFN are parameterized by $\W^1 \in  \mathbb{R}^{d \times d_{ff}}, \b^1 \in \mathbb{R}^{d_{ff}}$ and $ \W^2 \in \mathbb{R}^{d_{ff} \times d}, \b^2 \in \mathbb{R}^d$,
		where $d_{ff}$ is the number of neurons in the intermediate layer of FFN.
		Denote the input of FFN is $\A \in \mathbb{R}^{n \times d}$,
		the output of FFN  can be divided into computations of $d_{ff}$ neurons:
		\begin{equation}
			\label{eq:ffn}
				\text{FFN}_{\W^1, \W^2,\b^1, \b^2}(\A) = \sum\nolimits_{i=1}^{d_{ff}}\text{GeLU}(\A\W^1_{:,i} + b^1_i)\W^2_{i,:} +\b^2.
		\end{equation}
	Based on (\ref{eq:mha}) and (\ref{eq:ffn}),
	the width of a Transformer layer can be adapted by varying the number of attention heads in MHA and neurons in the intermediate layer of FFN (Figure~\ref{fig:rewire}). 
For
	 width multiplier $m_w$, we retain the leftmost $\lfloor m_w N_H \rfloor$ attention heads in MHA and $\lfloor m_w d_{ff} \rfloor$ neurons in the intermediate layer of FFN. 
	In this case, each Transformer layer is roughly compressed by a ratio of $m_w$. This is not strictly equal as layer normalization and bias in linear layers also have very few parameters.
	The number of neurons in the embedding dimension is not adapted because these neurons are connected through skip connections across all Transformer layers and
	cannot be flexibly scaled.

	\subsubsection{Network Rewiring}
	\label{sec:rewire}
	To fully utilize the network's capacity, the more important heads or neurons should be shared across more sub-networks.
Thus
	before training the width-adaptive network, we rank the attention heads and neurons according to their importance in the fine-tuned BERT model, and then arrange them 
	with descending importance in the width direction 
	(Figure~\ref{fig:rewire}).
	
	Following~\cite{molchanov2016pruning,voita2019analyzing},
	we compute the importance score of a head or neuron
	based on  the  variation in the training loss $\mathcal{L}$ if we remove it. 
	Specifically, 
	for one head with output $\h$,
	its importance $I_h$ can be estimated using the first-order Taylor expansion as
\begin{equation*}
	I_\h = \left|\mathcal{L}_\h - \mathcal{L}_{\h=\0}\right| = \left|\mathcal{L}_\h - (\mathcal{L}_\h - \frac{\partial \mathcal{L}}{\partial \h}(\h-\0) + R_{\h=\0})\right| \approx \left|\frac{\partial \mathcal{L}}{\partial \h}\h\right|
\end{equation*}
	if we ignore the remainder $R_{\h=\0}$.
	Similarly, for a neuron in the intermediate layer of FFN, 
	denote the 
	set of 
	weights  in $\W^1$ and $\W^2$ connected to it
	 as $\w = \{w_1, w_2, \cdots, w_{2d}\}$,
	its importance is estimated by 
	\begin{equation*}
		\left|\frac{\partial \mathcal{L}}{\partial \w}\w\right|
	= \left|\sum\nolimits_{i=1}^{2d}\frac{\partial \mathcal{L}}{\partial w_i}w_i\right|.
	\end{equation*}
	Empirically, we use the development set to calculate the importance of  attention heads and neurons.

	\subsubsection{Training with Adaptive Width}
	\label{sec:width}

	After the connections of the  BERT model are rewired according to Section~\ref{sec:rewire}, we
	use knowledge distillation to train $\text{DynaBERT}_\text{W}$.
	Specifically, we use the rewired BERT model as the fixed teacher network, and to initialize $\text{DynaBERT}_\text{W}$. 
	Then we distill the knowledge from the fixed teacher model to student sub-networks at different widths in   $\text{DynaBERT}_\text{W}$ (First stage in Figure~\ref{fig:model}).

	Following~\cite{jiao2019tinybert},
	we transfer the knowledge in  logits $\y$,  embedding (i.e., the output of the embedding layer) $\E$, 
	and  hidden states (i.e. the output of each Transformer layer) $\H_l$ ($l=1,2,\cdots, L$)
	from the teacher model to $\y^{(m_w)}, \E^{(m_w)}$ and $\H_l^{(m_w)}$ of the student sub-network
	with width multiplier $m_w$.
	Here $\E, \H_l,  \E^{(m_w)}, \H_l^{(m_w)} \in \mathbb{R}^{ n \times d}$.
		Denote 
	$\text{SCE}$ as the soft cross-entropy loss and $\text{MSE}$ as the   mean squared error.
	The three  distillation loss terms are  
	\begin{equation*}
	\ell_{pred}(\y^{(m_w)},\y)= \text{SCE}(\y^{(m_w)},\y), \quad  \ell_{emb}(\E^{(m_w)},\E)=\text{MSE}(\E^{(m_w)}, \E),
	\end{equation*}
	\begin{equation*}
	\ell_{hidn}(\H^{(m_w)}, \H)=\sum \nolimits_{l=1}^{L}\text{MSE}(\H_l^{(m_w)}, \H_l).
	\end{equation*}
	Thus the training objective is
	\begin{equation}
	\label{eq:loss}
	\mathcal{L}=  \lambda_1\ell_{pred}(\y^{(m_w)}, \y) + \lambda_2(\ell_{emb}(\E^{(m_w)}, \E)+\ell_{hidn}(\H^{(m_w)}, \H)),
	\end{equation}
	where
	$\lambda_1$ and $\lambda_2$ are the scaling parameters that control the weights of different loss terms. 
	Note that we use the same scaling parameter for 
	 the distillation loss of the embedding and hidden states, because the two have the same dimension and similar scale.
	Empirically, we choose $(\lambda_1, \lambda_2) = (1, 0.1)$  because $\ell_{emb}+\ell_{hidn}$
	is about one magnitude larger than  $\ell_{pred}$. 
	The detailed  training process of  $\text{DynaBERT}_\text{W}$ is shown in  Algorithm~\ref{alg:adaptive}, where we restrict the depth multiplier $m_d$ to $1$ (i.e., the largest depth) as $\text{DynaBERT}_\text{W}$ is  only adaptive in width.
	 To provide more task-specific data for distillation learning, we use the data augmentation method in TinyBERT~\cite{jiao2019tinybert},
	which uses a pre-trained BERT~\cite{devlin2019bert} trained from the masked language modeling task to generate task-specific augmented samples.

	\subsection{Training DynaBERT with Adaptive Width and Depth}
	\label{sec:depth_width}
	After  $\text{DynaBERT}_\text{W}$ is trained, we use it
	as the fixed teacher model, and to initialize the DynaBERT model. 
	Then we distill the knowledge from the fixed teacher model at the maximum depth to student sub-networks at equal or lower depths (Second stage in Figure~\ref{fig:model}). 
	To avoid catastrophic forgetting of learned elasticity in the width direction, we  still train over different widths in each iteration. 
	For width multiplier $m_w$, 
	the objective of the student sub-network with depth multiplier $m_d$ still contains three terms $\ell'_{pred}, \ell'_{emb}$ and $\ell'_{hidn}$ as in
	(\ref{eq:loss}).
	It
	makes the logits $\y^{(m_w, m_d)}$, embedding $\E^{(m_w, m_d)}$ and hidden states $\H^{(m_w, m_d)}$    mimic  $\y^{(m_w)}$, $\E^{(m_w)}$ and $\H^{(m_w)}$  from the teacher model with the maximum depth. 
	When the depth multiplier $m_d<1$, 
	 the student has fewer layers than the teacher.
	In this case, we use the ``Every Other'' strategy in~\cite{fan2019reducing} and drop  layers evenly to get a balanced network.
	Then we match
	the hidden states of the remaining layers $L_S$ in the student sub-network with 
	those  at depth $d \in L_T$ which satisfies $\text{mod}(d+1, \frac{1}{1-m_d}) \neq 0$ from the teacher model as 	
	\begin{equation*}
	\ell'_{hidn}( \H^{(m_w,m_d)},\H^{(m_w)}) = \sum\nolimits_{l,l'  \in L_S, L_T}\text{MSE}( \H_l^{(m_w,m_d)},\H_{l'}^{(m_w)}).
	\end{equation*}
	We use $d+1$ here 
	because we want to keep the knowledge in the last layer of the teacher model which is shown to be important in \cite{wang2020minilm}. 
	A detailed example can be found at Appendix~\ref{apdx:select_layer}. 
	Thus the distillation objective can still be written as 
\begin{eqnarray}
\mathcal{L}= 	\lambda_1 \ell'_{pred} ( \y^{(m_w,m_d)}, \y^{(m_w)}) + \lambda_2 ( \ell'_{emb}( \E^{(m_w,m_d)},\E^{(m_w)}) + \ell'_{hidn}( \H^{(m_w,m_d)},\H^{(m_w)})). \label{eq:loss1}
\end{eqnarray}
	For simplicity, we do not tune $\lambda_1, \lambda_2$ and  choose $(\lambda_1, \lambda_2) = (1, 1)$ in our experiments.
	The training procedure can be found in Algorithm~\ref{alg:depth_width}. 
    After training with the augmented data and the distillation objective above (Step 1), 
	one can further fine-tune the network using the original data and 
	the cross-entropy loss 
	between the predicted labels
	and  ground-truth labels (Step 2).
	Step 2 further improves the performance on some data sets empirically (Details are in Section~\ref{expt:ablation}). 
	In this work, we report results of the model with higher average validation accuracy 
	of all sub-networks, between  before (Step 1)  and after fine-tuning (Step 2) with the original data.

	\section{Experiment}
	In this section, we evaluate the efficacy of the proposed DynaBERT on the General Language Understanding Evaluation~(GLUE) tasks~\cite{wang2019glue} 
	and the machine reading comprehension task SQuAD v1.1~\cite{rajpurkar2016squad},
	using both  $\text{BERT}_\text{BASE}$~\cite{devlin2019bert}  and $\text{RoBERTa}_\text{BASE}$~\cite{liu2019roberta} as the backbone models.
	The corresponding both width- and depth-adaptive BERT and RoBERTa models are named as DynaBERT and DynaRoBERTa, respectively.
	For $\text{BERT}_\text{BASE}$ and $\text{RoBERTa}_\text{BASE}$, the number of Transformer layers is $L=12$, the hidden state size is $d=768$.
	In each Transformer layer, the number of heads in MHA is $N_H = 12$, and the number of neurons in the intermediate layer in  FFN is $d_{ff} = 3072$. 
	The list of width 
	multipliers is $[1.0,0.75,0.5,0.25]$, and the 
	list of depth multipliers is 
	$[1.0,0.75,0.5]$. There are a total of 
	$4\times 3=12$ 
	different configurations of 
	sub-networks. 
		We use Nvidia V100 GPU for training.
	Detailed hyperparameters
	 for the experiments 
	are 
	in Appendix~\ref{apdx:hyper}.

We compare  the proposed DynaBERT and DynaRoBERTa with  (i) the base models 
 $\text{BERT}_\text{BASE}$~\cite{devlin2019bert} and $\text{RoBERTa}_\text{BASE}$~\cite{liu2019roberta}; and (ii) popular BERT compression methods, including distillation methods
	DistilBERT~\cite{sanh2019distilbert},
	TinyBERT~\cite{jiao2019tinybert},
	and adaptive-depth method LayerDrop~\cite{fan2019reducing}.
	The results of the compared methods are taken from their original paper or official code repository.
	We evaluate the efficacy of our proposed DynaBERT and DynaRoBERTa under different efficiency constraints, including \#parameters, FLOPs, the latency on Nvidia K40 GPU and Kirin 810 A76 ARM CPU (Details can be found in 
	Appendix~\ref{apdx:efficiency_constraint}).

	\subsection{Results on the GLUE benchmark}
	\label{expt:glue}
	\paragraph{Data.}
	The GLUE benchmark~\cite{wang2019glue} is a collection of diverse natural language understanding tasks.
	Detailed descriptions of GLUE data sets are included in Appendix~\ref{apdx:glue_data}.
	Following~\cite{devlin2019bert}, 
	for the development set, we report Spearman correlation for \texttt{STS-B}, Matthews correlation for \texttt{CoLA} and accuracy for the other tasks.
	For the test set of \texttt{QQP} and \texttt{MRPC}, we report ``F1''.
	
	\paragraph{Main Results.}
	\label{expt:main}
	\begin{table}[htbp]
		\caption{Development set results of the GLUE benchmark using DynaBERT and DynaRoBERTa with different width and depth multipliers $(m_w, m_d)$. 
 }
		\label{tbl:main}
		\centering
		\scalebox{0.725}{
			\begin{tabular}{ll|ccc|ccc|ccc|ccc}
				\hline
				Method                           &              &     \multicolumn{3}{c|}{\texttt{CoLA}}     & \multicolumn{3}{c|}{\texttt{STS-B}}  & \multicolumn{3}{c|}{\texttt{MRPC}} &   \multicolumn{3}{c}{\texttt{RTE}}   \\ \hline
				$\text{BERT}_\text{BASE}$        &              &                    &   58.1    &           &               &     89.8      &      &               & 87.7  &            &               &     71.1      &      \\ \hline
				\multirow{5}{*}{DynaBERT}    & \diagbox[width=1.5cm,height=0.5cm]{$m_w$}{$m_d$} &        1.0x        &   0.75x   &   0.5x    &     1.0x      &     0.75x     & 0.5x &     1.0x      & 0.75x &    0.5x    &     1.0x      &     0.75x     & 0.5x \\ \cline{2-14}
				& 1.0x         &        59.7        &   59.1    &   54.6    & \textbf{90.1} &     89.5      & 88.6 &     86.3      & 85.8  &     85.0     &     72.2      &     71.8      & 66.1 \\
				& 0.75x        &   \textbf{60.8}    &   59.6    &   53.2    &     90.0      &     89.4      & 88.5 & \textbf{86.5} & 85.5  &    84.1    &     71.8      & \textbf{73.3} & 65.7 \\
				& 0.5x         &        58.4        &   56.8    &   48.5    &     89.8      &     89.2      & 88.2 &     84.8      & 84.1  &    83.1    &     72.2      &     72.2      & 67.9 \\
				& 0.25x        &        50.9        &   51.6    &   43.7    &     89.2      &     88.3      & 87.0 &     83.8      & 83.8  &    81.4    &     68.6      &     68.6      & 63.2 \\ \hline
				&              & \multicolumn{3}{c|}{\texttt{MNLI-(m/mm)}}  &  \multicolumn{3}{c|}{\texttt{QQP}}   & \multicolumn{3}{c|}{\texttt{QNLI}} &  \multicolumn{3}{c}{\texttt{SST-2}}  \\ \hline
				$\text{BERT}_\text{BASE}$        &              &                    & 84.8/84.9 &           &               &     90.9      &      &               & 92.0  &            &               &     92.9      &      \\ \hline
				\multirow{5}{*}{DynaBERT}    & \diagbox[width=1.5cm,height=0.5cm]{$m_w$}{$m_d$} &        1.0x        &   0.75x   &   0.5x    &     1.0x      &     0.75x     & 0.5x &     1.0x      & 0.75x &    0.5x    &     1.0x      &     0.75x     & 0.5x \\ \cline{2-14}
				& 1.0x         & \textbf{84.9/85.5} & 84.4/85.1 & 83.7/84.6 &     \textbf{91.4}      & \textbf{91.4} & 91.1 &     92.1      & 91.7  &    90.6    &     93.2      &     93.3      & 92.7 \\
				& 0.75x        &     84.7/85.5      & 84.3/85.2 & 83.6/84.4 & \textbf{91.4} &     91.3      & 91.2 &     92.2      & 91.8  &    90.7    &     93.0      &     93.1      & 92.8 \\
				& 0.5x         &     84.7/85.2      & 84.2/84.7 & 83.0/83.6 &     91.3      &     91.2      & 91.0 & \textbf{92.2} & 91.5  &    90.0    & \textbf{93.3} &     92.7      & 91.6 \\
				& 0.25x        &     83.9/84.2      & 83.4/83.7 & 82.0/82.3 &     90.7      &     91.1      & 90.4 &     91.5      & 90.8  &    88.5    &     92.8      &     92.0      & 92.0 \\ \hline \hline
				&              &     \multicolumn{3}{c|}{\texttt{CoLA}}     & \multicolumn{3}{c|}{\texttt{STS-B}}  & \multicolumn{3}{c|}{\texttt{MRPC}} &   \multicolumn{3}{c}{\texttt{RTE}}   \\ \hline
				$\text{RoBERTa}_\text{BASE}$     &              &                    &   65.1    &           &               &     91.2      &      &               & 90.7  &            &               &     81.2      &      \\ \hline
				\multirow{5}{*}{DynaRoBERTa} & \diagbox[width=1.5cm,height=0.5cm]{$m_w$}{$m_d$} &        1.0x        &   0.75x   &   0.5x    &     1.0x      &     0.75x     & 0.5x &     1.0x      & 0.75x &    0.5x    &     1.0x      &     0.75x     & 0.5x \\ \cline{2-14}
				& 1.0x         &        63.6        &   61.0.7    &   59.5    & \textbf{91.3} &     91.0      & 90.0 &     88.7      & 89.7  &    88.5    & \textbf{82.3} &     78.7      & 72.9 \\
				& 0.75x        &   \textbf{63.7}    &   61.4    &   54.9    &     91.0      &     90.7      & 89.7 &     90.0      & 89.2  &    88.2    &     79.4      &     77.3      & 70.8 \\
				& 0.5x         &        61.3        &   58.1    &   52.9    &     90.3      &     90.1      & 88.9 & \textbf{90.4} & 90.0  &    86.5    &     75.1      &     73.6      & 71.5 \\
				& 0.25x        &        54.2        &   46.7    &   39.8    &     89.6      &     89.2      & 87.5 &     88.2      & 88.0  &    84.3    &     70.0      &     70.0      & 66.8 \\ \hline
				&              & \multicolumn{3}{c|}{\texttt{MNLI-(m/mm)}}  &  \multicolumn{3}{c|}{\texttt{QQP}}   & \multicolumn{3}{c|}{\texttt{QNLI}} &  \multicolumn{3}{c}{\texttt{SST-2}}  \\ \hline
				$\text{RoBERTa}_\text{BASE}$     &              &                    & 87.5/87.5 &           &               &     91.8      &      &               & 93.1  &            &               &     95.2      &      \\ \hline
				\multirow{5}{*}{DynaRoBERTa} & \diagbox[width=1.5cm,height=0.5cm]{$m_w$}{$m_d$}&        1.0x        &   0.75x   &   0.5x    &     1.0x      &     0.75x     & 0.5x &     1.0x      & 0.75x &    0.5x    &     1.0x      &     0.75x     & 0.5x \\ \cline{2-14}
				& 1.0x         & \textbf{88.3/87.6} & 87.7/87.2 & 86.2/85.8 &     92.0      & \textbf{92.0} & 91.7 & \textbf{92.9} & 92.5  &    91.4    & \textbf{95.1} &     94.3      & 93.3 \\
				& 0.75x        &     88.0/87.3      & 87.5/86.7 & 85.8/85.4 &     91.9      &     91.8      & 91.6 &     92.8      & 92.4  &    91.3    &     94.6      &     94.3      & 93.3 \\
				& 0.5x         &     87.1/86.4      & 86.8/85.9 & 84.8/84.2 &     91.7      &     91.5      & 91.2 &     92.3      & 91.9  &    90.8    &     93.6      &     94.2      & 92.9 \\
				& 0.25x        &     84.6/84.7      & 84.0/83.7 & 82.1/82.0 &     91.2      &     91.0      & 90.5 &     90.9      & 90.9  &    89.3    &     93.9      &     93.2      & 91.6 \\ \hline
			\end{tabular}
		}
	\end{table}
	
	Table~\ref{tbl:main} shows the results of sub-networks derived from the proposed DynaBERT and DynaRoBERTa.
	The proposed DynaBERT~(or DynaRoBERTa) achieves comparable performances as $\text{BERT}_\text{BASE}$~(or $\text{RoBERTa}_\text{BASE}$) with the same or smaller size.
	For most tasks,
	the sub-network of DynaBERT or DynaRoBERTa with the maximum size does not necessarily have the best performance, indicating that  redundancy exists in the original BERT or RoBERTa model.
	Indeed, with the proposed method, the model's width and depth for most tasks can be reduced without performance drop.
	Another observation is that using one specific width multiplier usually has higher accuracy than using the same depth multiplier.
	This indicates that compared to the depth direction, the width direction is more robust to compression. 
	Sub-networks from DynaRoBERTa, most of the time, perform significantly better than those from DynaBERT under the same depth and width.
			Test set results in Appendix~\ref{apdx:more_glue} also show that DynaBERT (resp. DynaRoBERTa) at its largest size has comparable or better accuracy as $\text{BERT}_\text{BASE}$ (resp. $\text{RoBERTa}_\text{BASE}$).

\begin{figure}[htbp]	
	\centering
	\subfloat{
		\includegraphics[width=0.98\textwidth]{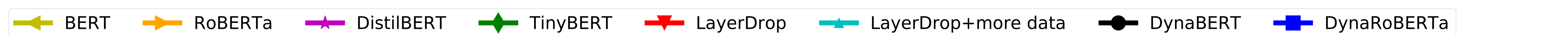}}
	\\
	\addtocounter{subfigure}{-1}
	\subfloat[\#parameters(G).\label{fig:param}]{
		\includegraphics[width=0.245\textwidth]{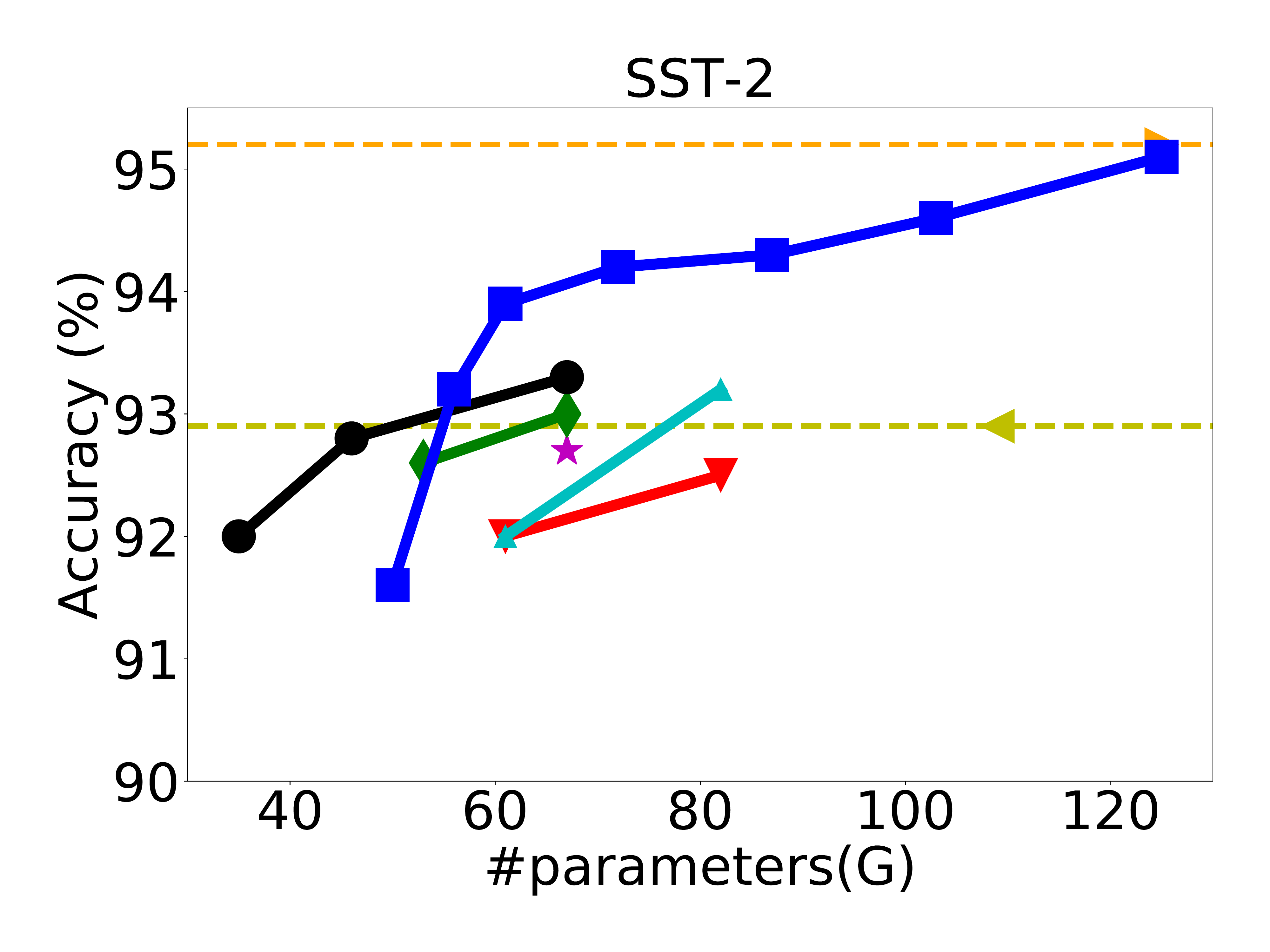}
		\includegraphics[width=0.245\textwidth]{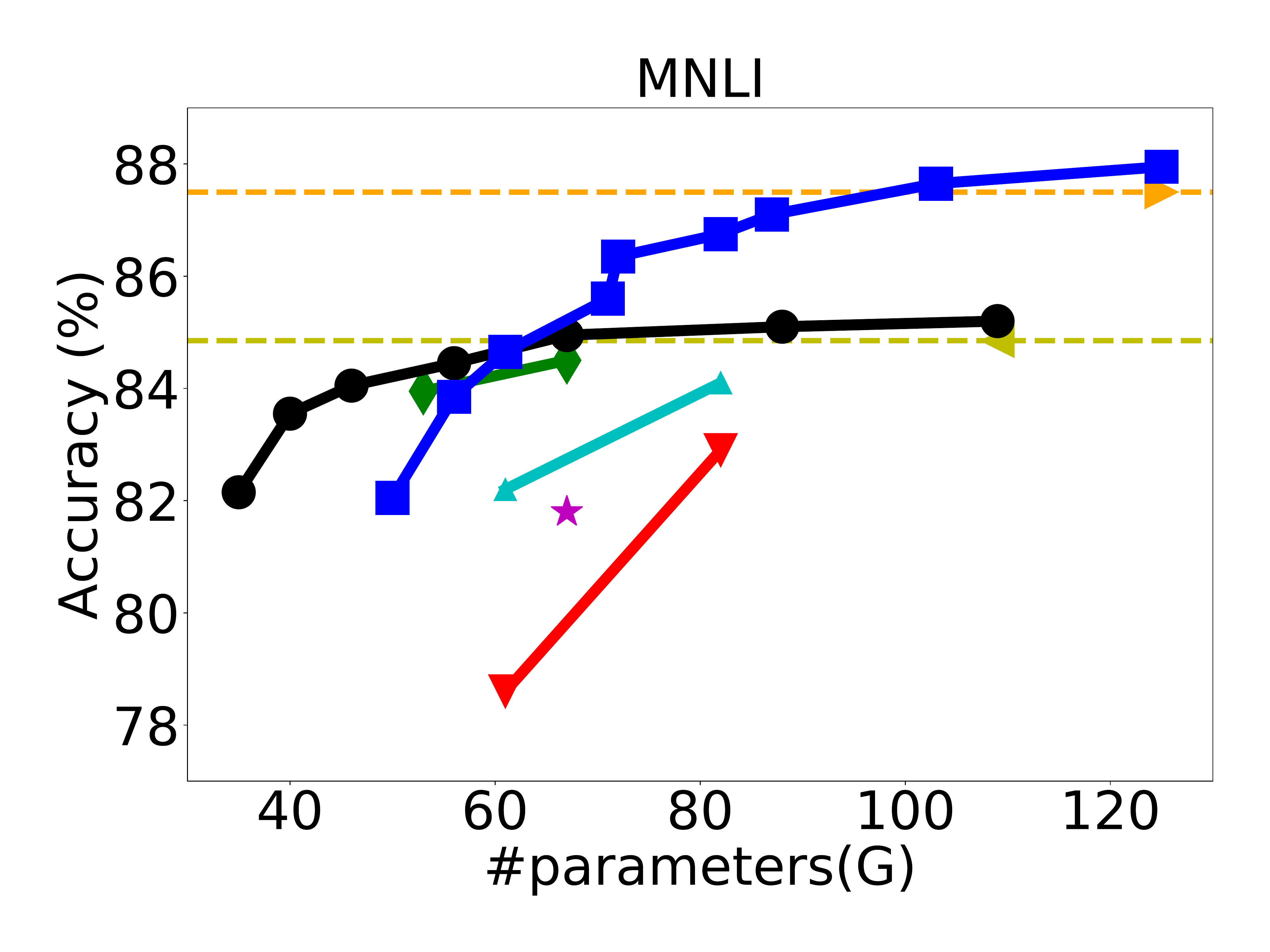}		}
	\subfloat[FLOPs(G).\label{fig:flops}]{
		\includegraphics[width=0.245\textwidth]{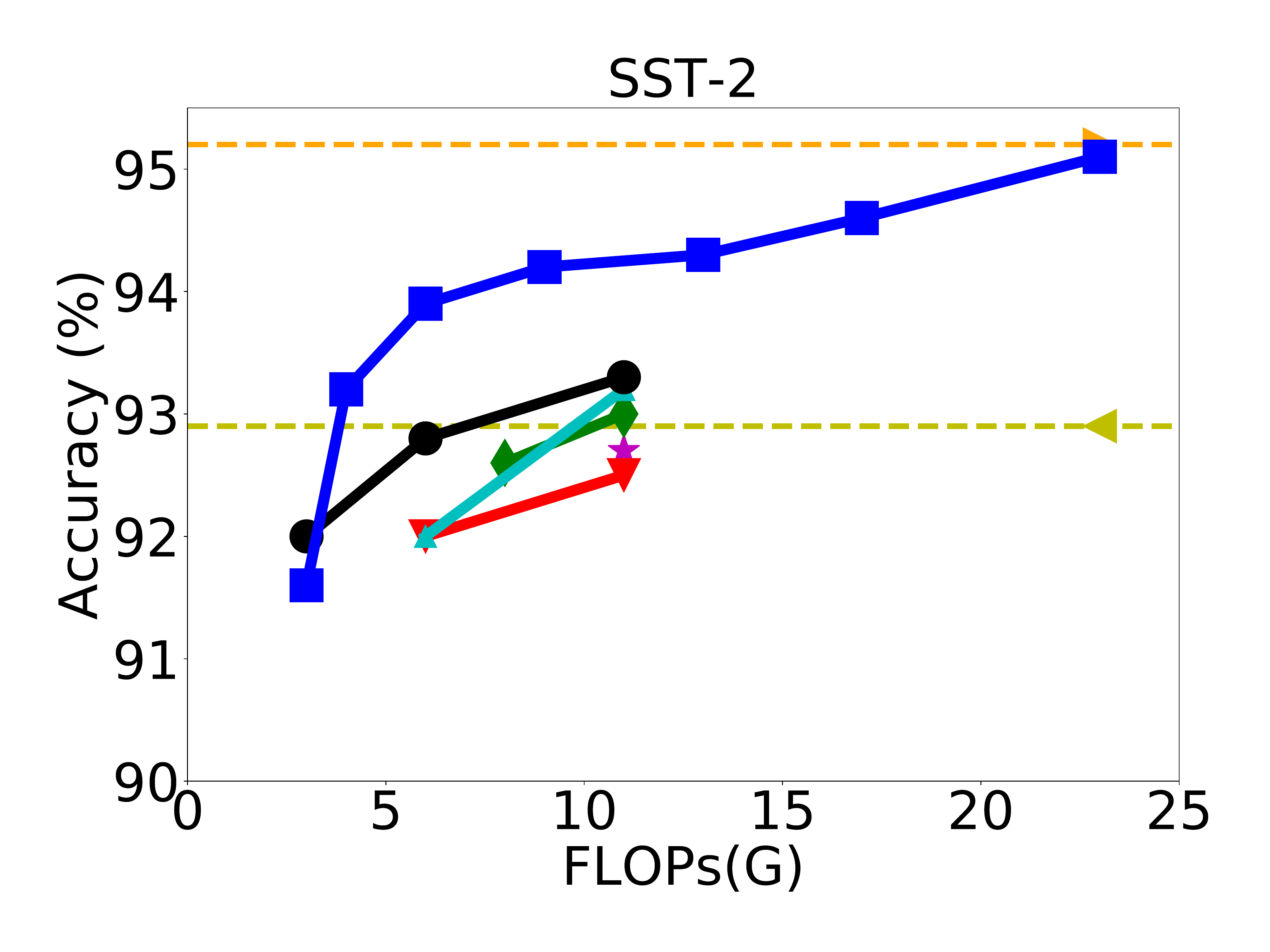}	
		\includegraphics[width=0.245\textwidth]{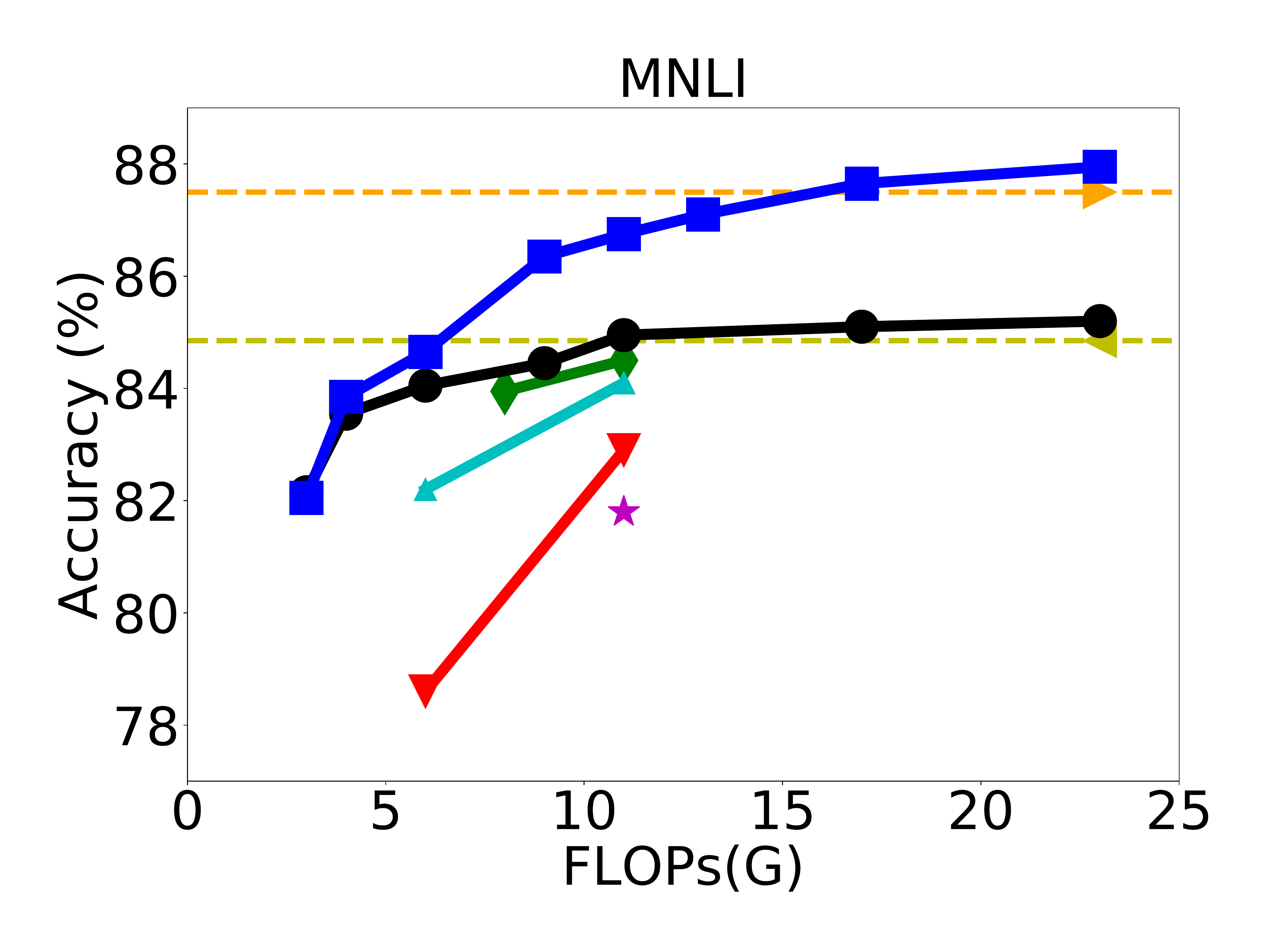}		}
	\vspace{-0.1in}
	\\
	\subfloat[Nvidia K40 GPU latency(s).\label{fig:gpu}]{
		\includegraphics[width=0.245\textwidth]{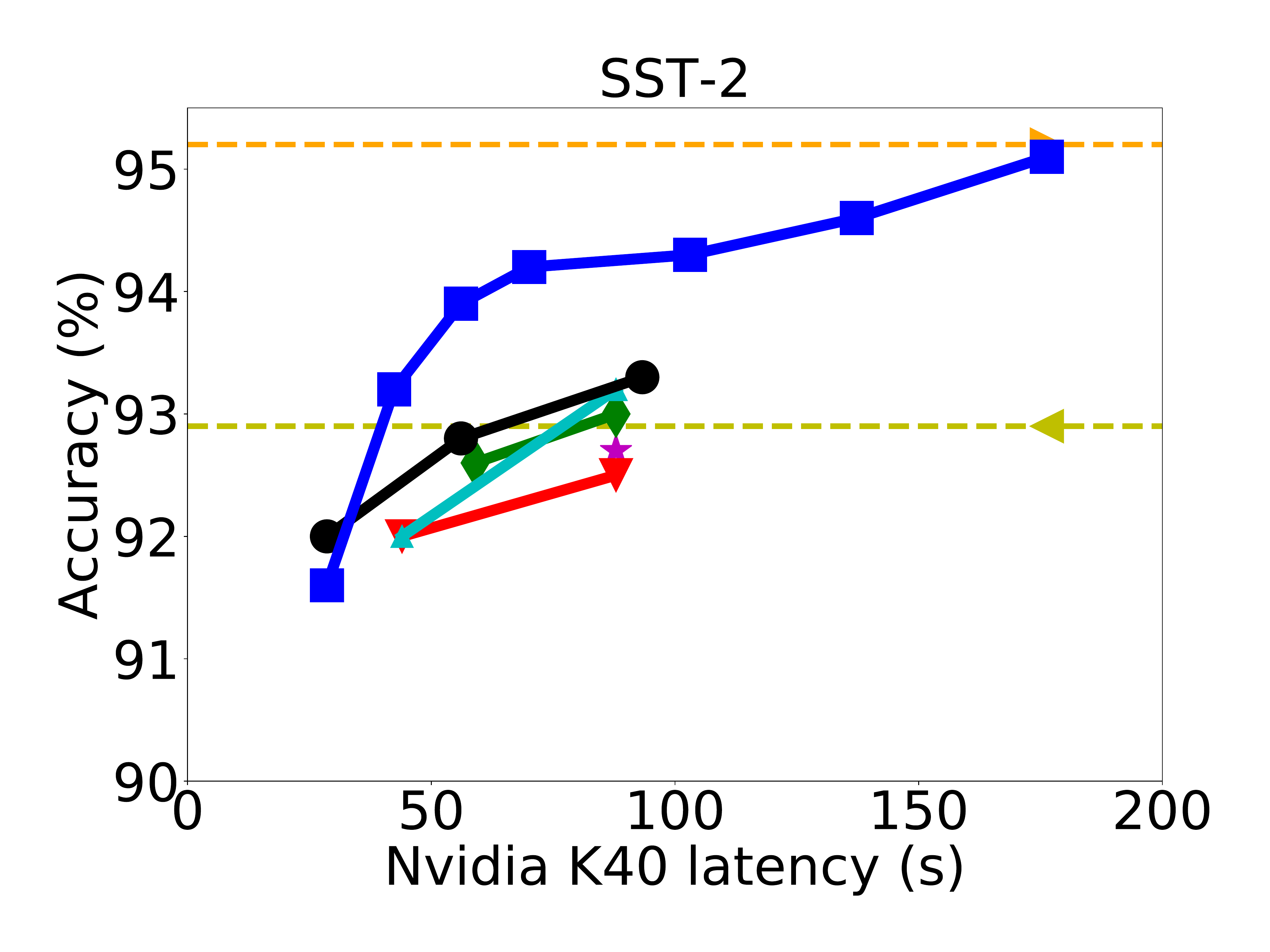}
		\includegraphics[width=0.245\textwidth]{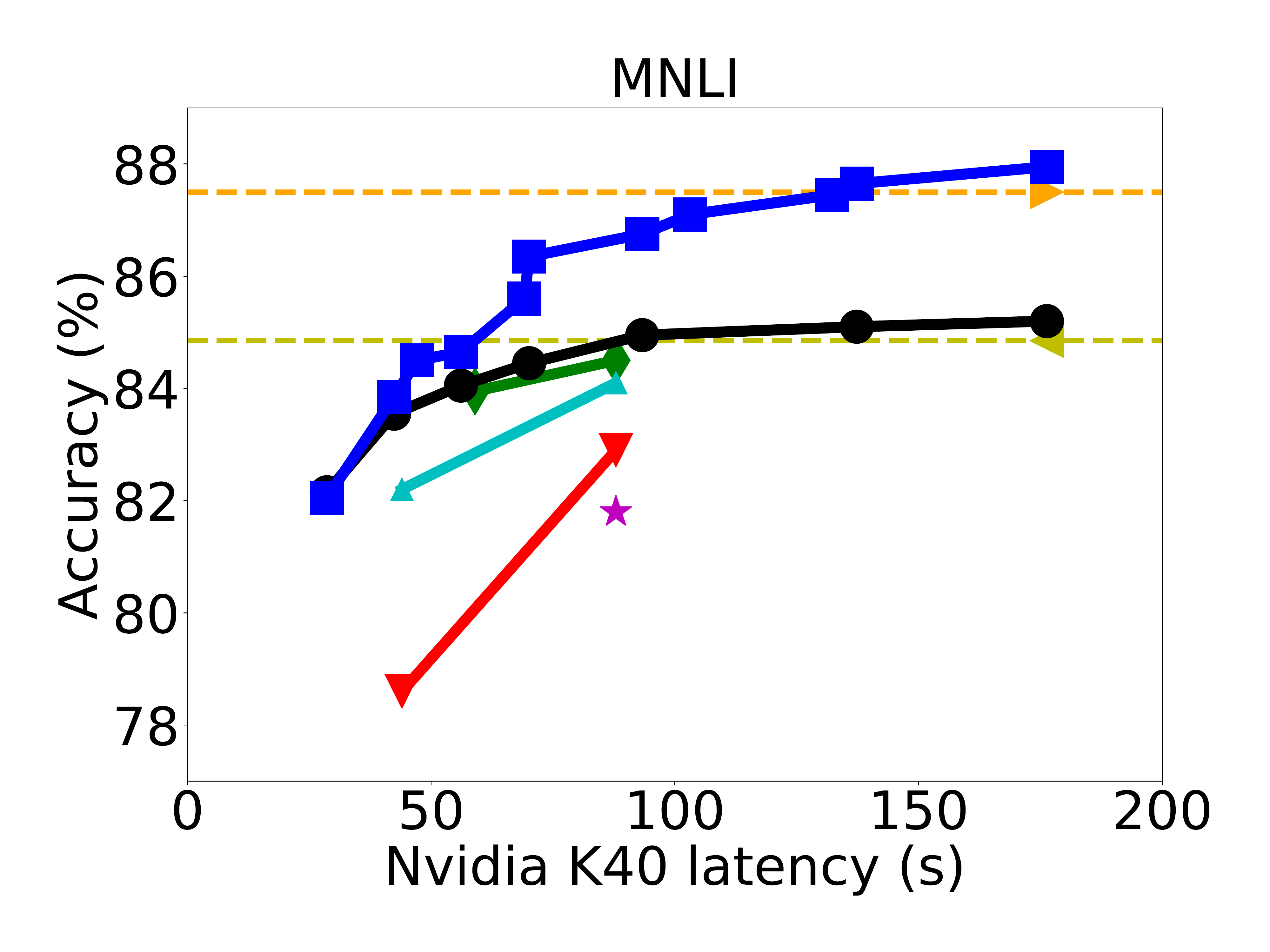}		}	
	\subfloat[Kirin 810 ARM CPU latency(ms).\label{fig:cpu}]{
		\includegraphics[width=0.245\textwidth]{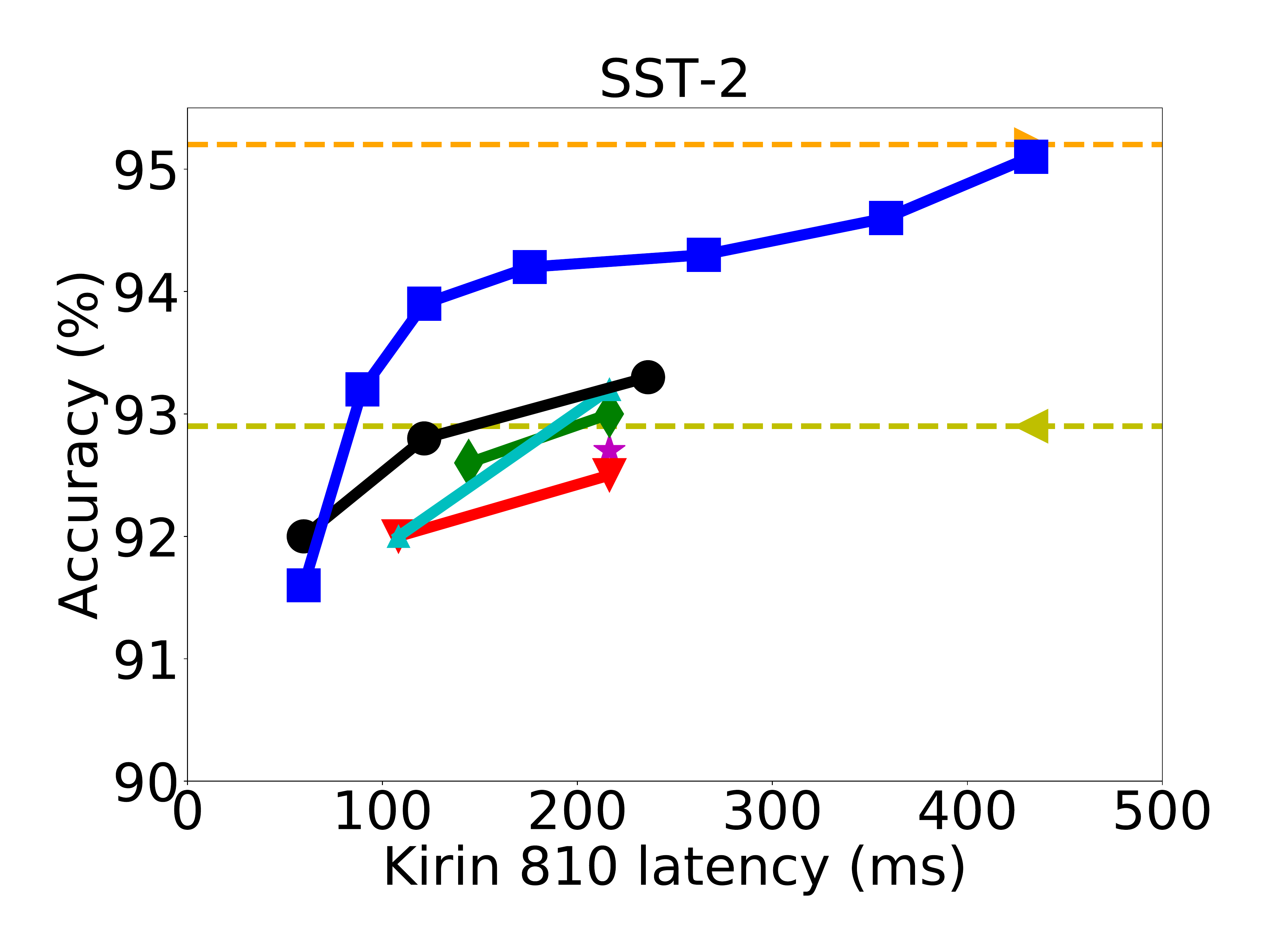}	
		\includegraphics[width=0.245\textwidth]{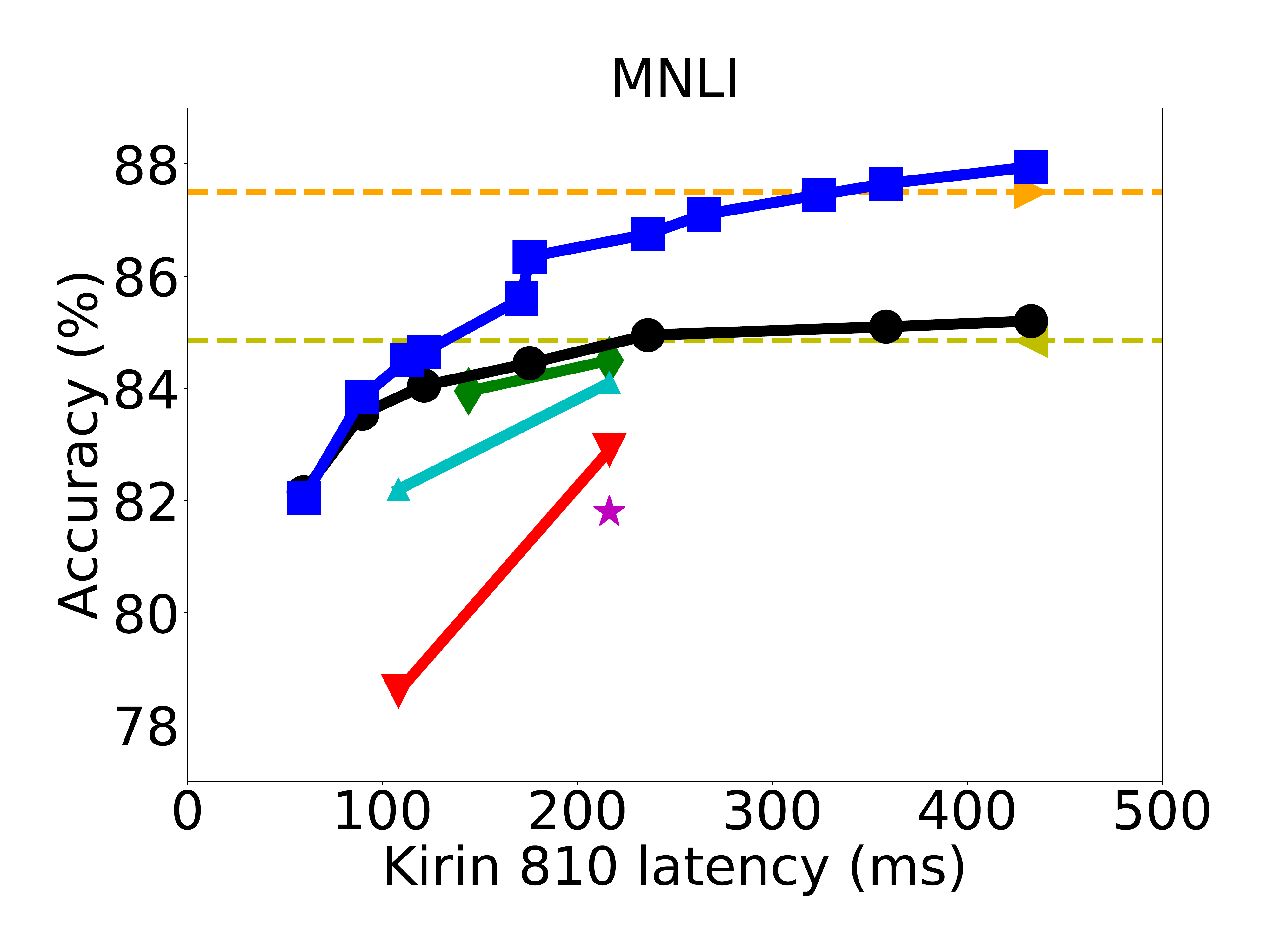}		}
	\caption{ Comparison of \#parameters, FLOPs, latency on GPU and CPU
		between our proposed DynaBERT and DynaRoBERTa and other methods. 
		The GPU latency is the running time of 100 batches with batch size
		128 and sequence length 128. The CPU latency is tested with batch size  1 and sequence length
		128.
		Average accuracy of \texttt{MNLI-m} and \texttt{MNLI-mm} is plotted.}
	\label{fig:comp}
\end{figure}

	\paragraph{Comparison with Other Methods.}
	Figure~\ref{fig:comp} compares DynaBERT and DynaRoBERTa on \texttt{SST-2} and  \texttt{MNLI} with other methods
	under different efficiency constraints, i.e., \#parameters, FLOPs, latency on Nvidia K40 GPU and Kirin 810 ARM CPU. Results of the other data sets are  in Appendix~\ref{apdx:more_glue}.
	Note that each number of TinyBERT and DistilBERT uses a different model, while different numbers of
	our proposed DynaBERT and DynaRoBERTa use different sub-networks within one model.
	
	From Figure~\ref{fig:comp},
	the proposed DynaBERT and DynaRoBERTa  achieve comparable accuracy  as $\text{BERT}_\text{BASE}$ and $\text{RoBERTa}_\text{BASE}$, but often require fewer
	parameters, FLOPs or lower latency.
	Under the same efficiency constraint, sub-networks extracted from 
	DynaBERT outperform DistilBERT and TinyBERT.
	Sub-networks  
	extracted from DynaRoBERTa outperform LayerDrop by  a large margin. 
	They even consistently outperform LayerDrop trained with much more data. 
	We speculate that it is because LayerDrop only allows flexibility in the depth direction.
	On the other hand, ours enables flexibility in both width and depth directions, which generates a significantly larger number of architectural configurations and better explores the balance between model accuracy and size.
	
	\subsection{Results on SQuAD}
	\label{sec:squad}
	SQuAD v1.1 (Stanford Question Answering Dataset)~\cite{rajpurkar2016squad} contains 100k crowd-sourced question/answer pairs.  	
	Given a question and a passage, the task is to extract the start and end of the answer span from the passage. 
	The performance metric used is EM (exact match) and F1. 
	
\begin{figure*}[htbp]
	\noindent\begin{minipage}{.51\textwidth}
	\captionof{table}{Development set results on SQuAD v1.1.}\label{tbl:squad}
				\vspace{-0.05in}
	\scalebox{0.8}{
		\begin{tabular}{ll|ccc}
			\hline
			Method                           &              &     \multicolumn{3}{c}{SQuAD v1.1}  \\ \hline
			$\text{BERT}_\text{BASE}$        &                  &    & 81.5/88.7 &                  \\ \hline
			\multirow{5}{*}{DynaBERT} & \diagbox[width=1.5cm,height=0.5cm]{$m_w$}{$m_d$}     & 1.0x & 0.75x & 0.5x \\ \cline{2-5}
 & 1.0x                 & 82.6/89.7 & 82.1/89.3 & 81.5/88.8 \\
& 0.75x                & 82.3/89.5 & 82.1/89.3 & 80.9/88.5 \\
& 0.5x                 & 81.9/89.2 & 81.7/89.0 & 80.0/87.8 \\
& 0.25x                & 80.7/88.1 & 79.9/87.5 & 76.6/85.0 \\ \hline
		\end{tabular}
	}
	\end{minipage}%
	\hspace{0.08in}
	\noindent\begin{minipage}{.48\textwidth}
		\centering
			\includegraphics[width=0.49\textwidth]{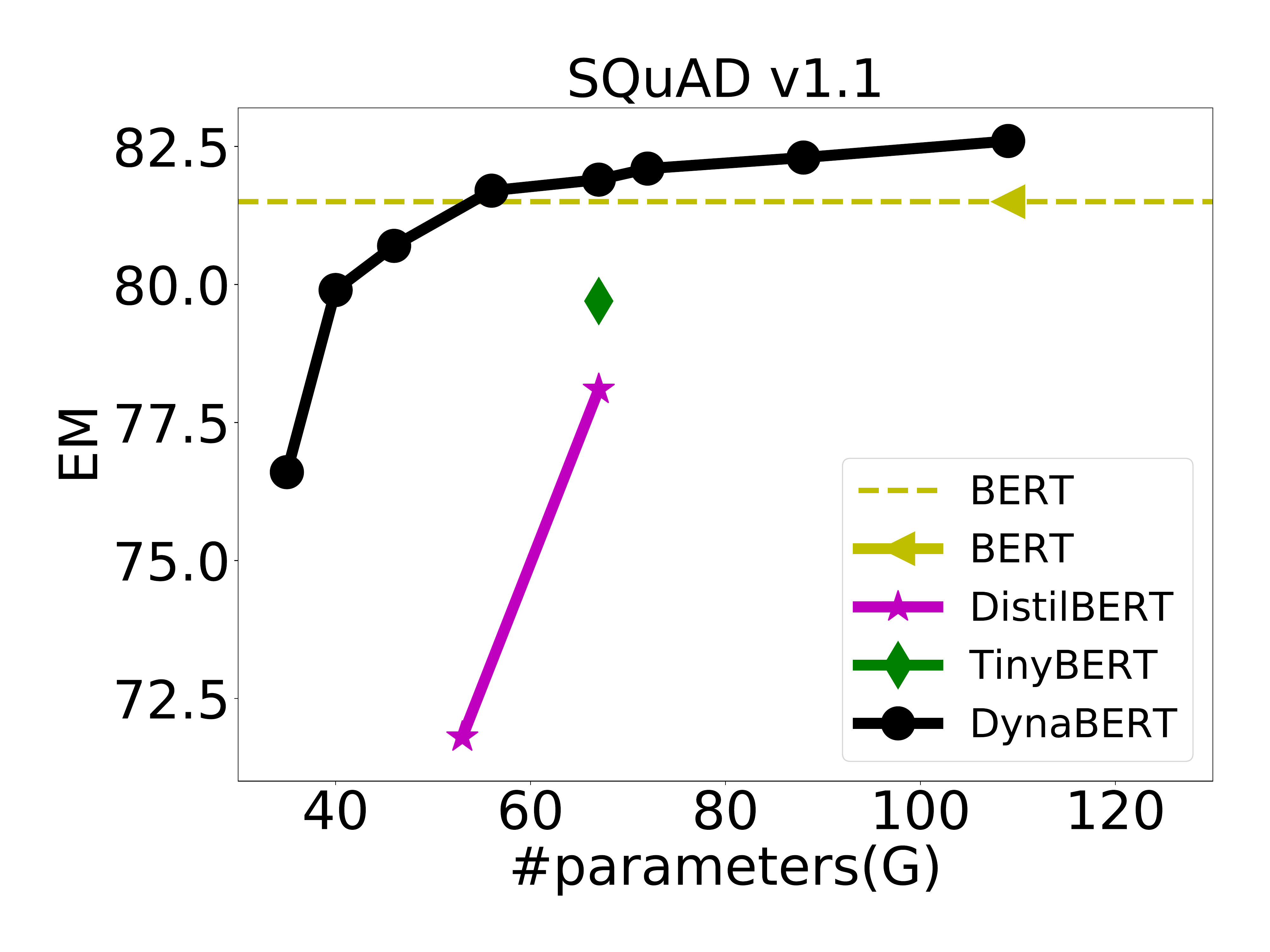}
\includegraphics[width=0.49\textwidth]{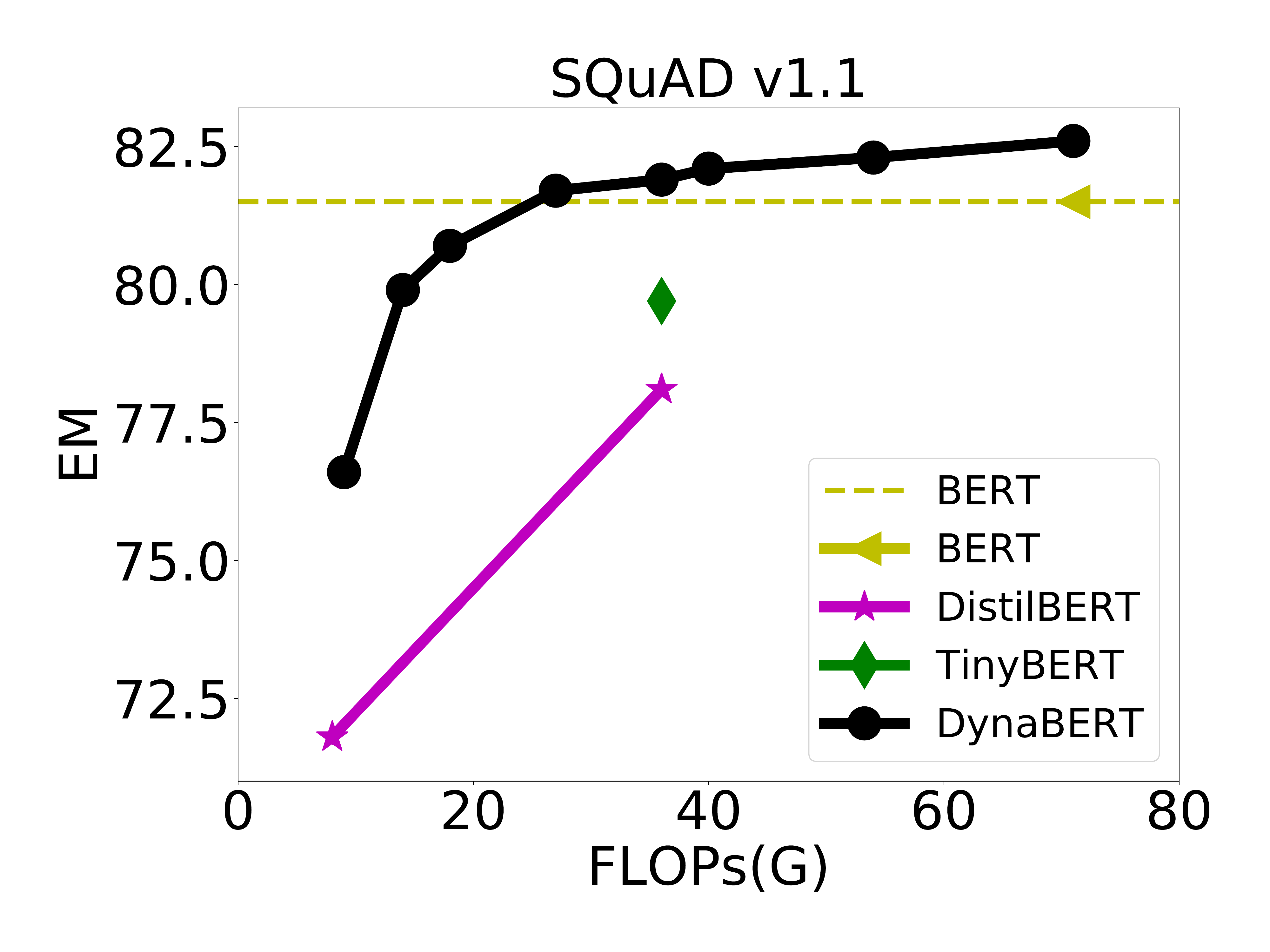}	
\vspace{-0.25in}
\captionof{figure}{ Comparison 
	of \#parameters and FLOPs
	between DynaBERT  and other methods. }
\label{fig:comp_squad}
	\end{minipage}
\end{figure*}

	Table~\ref{tbl:squad} shows the results of sub-networks extracted from DynaBERT. 
	Sub-network with only 1/2 width or depth of $\text{BERT}_\text{BASE}$ already achieves comparable or even better performance than it.
	 Figure~\ref{fig:comp_squad}  shows the comparison of sub-networks of DynaBERT and other methods.
	 We do not compare with LayerDrop~\cite{fan2019reducing} because SQuAD results are not reported in their paper.  
	 From Figure~\ref{fig:comp_squad}, 
	 with the same number of parameters, FLOPs, 
	 sub-networks  extracted from DynaBERT outperform TinyBERT and DistilEBRT by  a large margin.

	\subsection{Ablation Study}
\label{expt:ablation}
\paragraph{Training  $\text{DynaBERT}_\text{W}$ with Adaptive Width.}
In Table~\ref{tbl:width}, we evaluate the importance of network rewiring, knowledge distillation and data augmentation (DA) in the training of 
$\text{DynaBERT}_\text{W}$  using the method in Section~\ref{sec:width_adaptive} on the GLUE benchmark.
Due to space limit, only average accuracy of 4 width multipliers are shown in Table~\ref{tbl:width}. 
Detailed accuracy for each width multiplier can be found  in 
in Appendix~\ref{apdx:ablation}.
$\text{DynaBERT}_\text{W}$ trained without network rewiring, knowledge distillation and data augmentation is called ``vanilla  $\text{DynaBERT}_\text{W}$''.
We also compare 
against the baseline 
of using separate networks, each of which is initialized from the $\text{BERT}_\text{BASE}$ with a certain width multiplier $m_w \in [1.0, 0.75,0.5,0.25]$, and then fine-tuned on the downstream task.
From Table~\ref{tbl:width},  vanilla 
$\text{DynaBERT}_\text{W}$ 
outperforms the separate network baseline. Interestingly, the performance gain is more obvious for smaller data sets \texttt{CoLA}, \texttt{STS-B}, \texttt{MRPC} and \texttt{RTE}.
After network rewiring, 
the average accuracy increases by over 2 points.
The average accuracy further increases by 1.5 points with knowledge distillation and data augmentation.
	
	\begin{table}[htbp]
		\caption{Ablation study in training $\text{DynaBERT}_\text{W}$.
			Average accuracy of 4 width multipliers  is reported.}
		\label{tbl:width}
		\centering
		\scalebox{0.8}{
			\begin{tabular}{l|cccccccccc}
				\hline
				& \texttt{MNLI-m} & \texttt{MNLI-mm} & \texttt{QQP}  & \texttt{QNLI} & \texttt{SST-2} & \texttt{CoLA} & \texttt{STS-B} & \texttt{MRPC} & \texttt{RTE} & avg.\\ \hline
				Separate 	network                      &      82.2       &       82.2       &     90.3      &     87.8      &      91.0      &     39.9      &      84.6      &     78.8      &     61.6   & 77.6  \\ \hline
				Vanilla  $\text{DynaBERT}_\text{W}$    &      82.2       &       82.5       &     90.6      &     89.1      &      91.2      &     44.0      &      87.4      &     80.5      &     64.2    &  79.0\\
				\quad + Network rewiring               &      83.1       &       83.0       &     90.9      &     90.4      &      91.7      &     51.4      &      89.1      &     83.8      & \textbf{69.7} & 81.4\\
					\quad 	+ Distillation and  DA &  \textbf{84.5}  &  \textbf{84.9}   & \textbf{91.0} & \textbf{92.1} & \textbf{92.7}  & \textbf{55.9} & \textbf{89.7}  & \textbf{86.1} &     69.5   &  82.9 \\ \hline
			\end{tabular}
		}
	\end{table}

	\paragraph{Training DynaBERT with Adaptive Width and Depth.}
In Table~\ref{tbl:width_depth}, we evaluate the effect of  knowledge distillation, data augmentation and final fine-tuning in the training of 
DynaBERT described in Section~\ref{sec:depth_width}.
Detailed accuracy for each width and depth multiplier is 
in Appendix~\ref{apdx:ablation}.
The DynaBERT trained without knowledge distillation, data augmentation and final fine-tuning is called ``vanilla DynaBERT''.
From Table~\ref{tbl:width_depth},
with knowledge distillation and data augmentation, 
the average accuracy of 
each task
is significantly improved compared to the vanilla counterpart on all three data sets. 
Additional fine-tuning further improves the performance on 
\texttt{SST-2} and \texttt{CoLA}, but not \texttt{MRPC}.
Empirically, we choose the model with higher average accuracy between 
before and after fine-tuning with the original data using the method described in Section~\ref{sec:depth_width}. 
	
	\paragraph{$\text{DynaBERT}_\text{W}$ as a ``teacher assistant''.}
In Table~\ref{tbl:together}, we also compare with directly distilling the knowledge from the rewired BERT to DynaBERT without $\text{DynaBERT}_\text{W}$.
The average accuracy of 12 configurations of DynaBERT using $\text{DynaBERT}_W$ or not, are reported. 
As can be seen, 
using a width-adaptive $\text{DynaBERT}_\text{W}$ as a ``teacher assistant''
can efficiently bridge the large gap of size between the student  and teacher, and
has better performance on all three data sets investigated.

\begin{figure*}[htbp]
	\noindent\begin{minipage}{.5\textwidth}
		\captionof{table}{Ablation study in training $\text{DynaBERT}$.
			Average accuracy of 12 configurations  is  reported.}	
		\label{tbl:width_depth}
		\vspace{-0.08in}
		\scalebox{0.9}{
			\begin{tabular}{l|cccccccccc}
				\hline
				&     \texttt{SST-2} & \texttt{CoLA}  & \texttt{MRPC} \\ \hline
				Vanilla  $\text{DynaBERT}$                        &      91.3      &     46.0    &     82.1     \\ \hline
				\quad 	+ Distillation and  DA & 92.5 & 52.8  & \textbf{84.5}  \\ \hline
				\quad 	+ Fine-tuning  & \textbf{92.7}  & \textbf{54.8} & 83.2  \\ \hline
			\end{tabular}
		}

	\end{minipage}%
	\hspace{0.2in}
	\noindent\begin{minipage}{.45\textwidth}
	\captionof{table}{Whether using $\text{DynaBERT}_W$ as a ``teacher assistant''. Average accuracy of 12 configurations  is reported.}	
	\label{tbl:together}
	\vspace{-0.09in}
		\scalebox{0.9}{
			\begin{tabular}{l|ccc}
				\hline
				& \texttt{SST-2} & \texttt{CoLA} & \texttt{MRPC}  \\ \hline
				$\text{DynaBERT}$             &      92.7      &     54.8      &     84.5      \\ \hline
				\quad 	- 	$\text{DynaBERT}_W$ &      92.3          &     54.1      &     84.4       \\ \hline
			\end{tabular}
		}
	\end{minipage}
\end{figure*}

		\paragraph{Adaptive Depth First or Adaptive Width First?}
		\label{sec:depth_first}
		To finally obtain both width- and depth-adaptive DynaBERT, 
		one can also 
		 train a only depth-adaptive model $\text{DynaBERT}_\text{D}$ first as the ``teacher assistant'', and then distill knowledge from it to DynaBERT. 
		 	 Table~\ref{tbl:depth_first} shows the accuracy of $\text{DynaBERT}_\text{W}$ and $\text{DynaBERT}_\text{D}$
		 	  under different compression rates of 
		 	the Transformer layers. 
		As can be seen, $\text{DynaBERT}_\text{W}$ performs significantly better than $\text{DynaBERT}_\text{D}$ for smaller width/depth multiplier  $0.5$.
		This may because
		unlike the width direction where the computation of attention heads and neurons are in parallel (Equations (\ref{eq:mha}) and (\ref{eq:ffn}) in Section~\ref{sec:width_adaptive}),
		the depth direction computes layer by layer consecutively.
		Thus 
		we can not rewire the connections based on the importance of layers in $\text{DynaBERT}_\text{D}$, 
		leading to severe accuracy drop of sub-networks with smaller depth in $\text{DynaBERT}_\text{D}$.
	
		\begin{table}[htbp]
			\centering
	\caption{Comparison of $\text{DynaBERT}_\text{W}$ and $\text{DynaBERT}_\text{D}$.}
	\label{tbl:depth_first}
	\centering
	\scalebox{0.74}{
		\begin{tabular}{l|ccc|ccc|ccc|ccc|ccc}
			\hline
			                           & \multicolumn{3}{c|}{\texttt{QNLI}} & \multicolumn{3}{c|}{\texttt{SST-2}} & \multicolumn{3}{c|}{\texttt{CoLA}} & \multicolumn{3}{c|}{\texttt{STS-B}} & \multicolumn{3}{c}{\texttt{MRPC}} \\ \hline
			$m_w$ or $m_d$             & 1.0x & 0.75x &        0.5x         & 1.0x & 0.75x &         0.5x         & 1.0x & 0.75x &        0.5x         & 1.0x & 0.75x &         0.5x         & 1.0x & 0.75x &        0.5x        \\ \hline
			$\text{DynaBERT}_\text{W}$ & 92.5 & 92.4  &        92.3         & 92.9 & 93.1  &         93.0         & 59.0 & 57.9  &        56.7      & 90.0 & 90.0  &         89.9            & 86.0 & 87.0  &        87.3        \\ \hline
			$\text{DynaBERT}_\text{D}$ & 92.4 & 91.9  &        90.6         & 92.9 & 92.8  &         92.1         & 58.3 & 58.3  &        52.2     & 89.9 & 89.0  &         88.3             & 87.3 & 85.8  &        84.6        \\ \hline
		\end{tabular}
	}
\end{table}

	\subsection{Looking into DynaBERT}
	\label{sec:attention_vis}
\begin{figure}[h!]	
		\centering
		\includegraphics[width=0.95\textwidth]{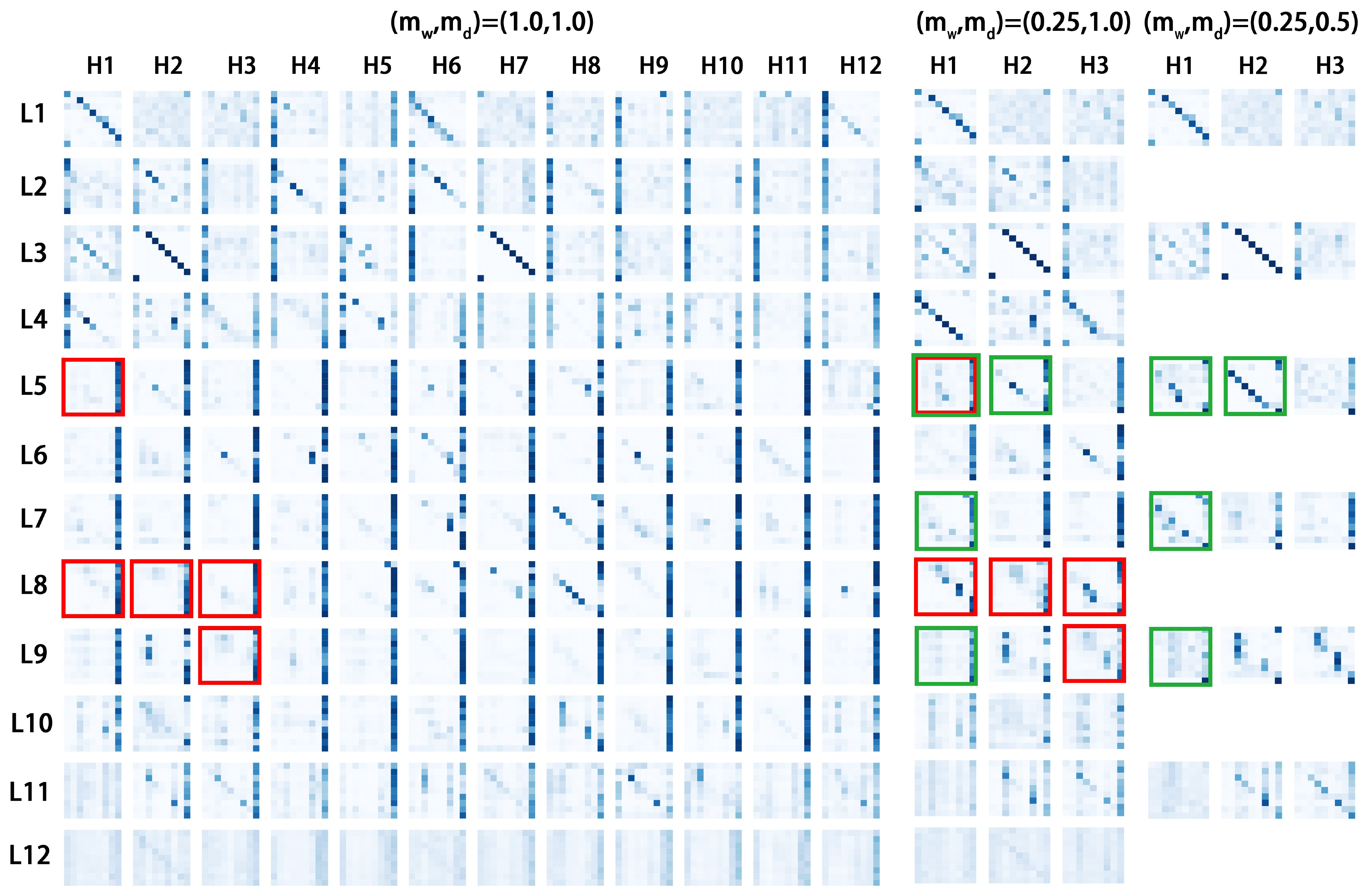}
		\caption{Attention maps of sub-networks with different widths and depths in DynaBERT 
			trained on \texttt{CoLA}. The sentence used is ``the cat sat on the mat.''}
		\label{fig:cola_orig}
	\end{figure}

	We  conduct a case study on the DynaBERT trained on \texttt{CoLA} by visualizing the attention distributions in Figure~\ref{fig:cola_orig}.
	The sentence used is ``the cat sat on the mat.''
	In \cite{voita2019analyzing,kovaleva2019revealing,rogers2020primer}, the attention heads
	for single-sentence task 
	are found to mainly play  ``positional'', ``syntactic/semantic'' functions.
	The positional head points to itself, adjacent tokens, [CLS], or [SEP] tokens, forming vertical or diagonal lines in the attention maps.
	The syntactic/semantic head points to tokens in a specific syntactic relation, and the attention maps do not have specific patterns.
	
	From Figure~\ref{fig:cola_orig}, 
	the  attention patterns in the first three layers in the sub-network with $(m_w,m_d)\!=\!(0.25,1.0)$ are quite similar to those in the full-sized model, while 
	those at intermediate layers show a clear function fusion. 
	For instance,
	 H1 in L5, H1-3 in L8, H3 in L9 (marked with red squares) in the sub-network with $(m_w,m_d)\!=\!(0.25,1.0)$  start to exhibit more syntactic or semantic patterns than their positional counterparts in the full-sized model.
	 This observation is consistent with the finding
	in \cite{jawahar2019does} that linguistic information is encoded in BERT's intermediate layers.
	Similarly, by comparing the attention maps in sub-networks with $(m_w,m_d)\!=\!(0.25,1.0)$ and $(m_w,m_d)\!=\!(0.25,0.5)$, functions (marked with green squares) also start  to fuse when the depth is compressed.
	
	Interestingly, we also find that  DynaBERT improves the ability of distinguishing linguistic acceptable and non-acceptable sentences for \texttt{CoLA}. This is consistent with 
	the superior performance of DynaBERT than $\text{BERT}_\text{BASE}$ in Table~\ref{tbl:main}.
	The attention patterns of \texttt{SST-2} also explain why  it can be compressed by a large rate without severe accuracy drop.
	Details can be found in Appendix~\ref{apdx:attention_vis}.

		\subsection{Discussion}
	
	\paragraph{Comparison of Conventional and Inplace Distillation.}
To train width-adaptive CNNs, in \cite{yu2019universally}, inplace distillation  is used to boost the performance.
Inplace distillation uses sub-network with the maximum width as the teacher while sub-networks with smaller widths in the same model as students.
Training loss includes the loss from both the teacher 
network and the student network.
Here we also adapt inplace distillation to train  $\text{DynaBERT}_\text{W}$ 
in Table~\ref{tbl:two_distillation},
 and compare it with the conventional distillation used in Section~\ref{sec:width_adaptive}.
For inplace distillation,
the student  mimics the teacher and the teacher mimics 
a fixed fine-tuned task-specific BERT, via  distillation loss over logits, embedding, and hidden states.
From Table~\ref{tbl:two_distillation},
inplace distillation has higher average accuracy on \texttt{MRPC} and \texttt{RTE}  
in training  $\text{DynaBERT}_\text{W}$, 
but performs worse on three data sets after  training DynaBERT.
	
	\begin{table}[htbp]
		\caption{Comparison of 
			conventional distillation and inplace distillation.  Average accuracy of 4 width multipliers ($\text{DynaBERT}_\text{W}$) or 12 configurations (DynaBERT) is reported.
		}
		\label{tbl:two_distillation}
		\centering
		\scalebox{0.81}{
			\begin{tabular}{l|c|cccc|c|cccc}
				\hline
				Distillation type                     &                                             & \texttt{SST-2} & \texttt{CoLA} & \texttt{MRPC} & \texttt{RTE} &                                    & \texttt{SST-2} & \texttt{CoLA} & \texttt{MRPC} & \texttt{RTE} \\ \hline
				Conventional                          & \multirow{2}{*}{$\text{DynaBERT}_\text{W}$} &      92.7      &     55.9      &     86.1      &     69.5     & \multirow{2}{*}{$\text{DynaBERT}$} &      92.7      &     54.8      &     84.5      &     69.5     \\
				Inplace~\cite{yu2019universally}                             &                                             &      92.6      &     55.9      &     87.0      &     70.0     &                                    &      92.5      &     54.5      &     84.8    &     69.0     \\ \hline
			\end{tabular}
		}
	\end{table}

	\paragraph{Different Methods to Train  $\text{DynaBERT}_\text{W}$.}
	
	For  $\text{DynaBERT}_\text{W}$ in Section~\ref{sec:width_adaptive}, we rewire the network only once, and train
by alternating over four different width multipliers.
In Table~\ref{tbl:pr_us}, 
we also adapt the following  two methods in training width-adaptive CNNs to BERT:
(1) using progressive rewiring (PR) as in \cite{Cai2020Once} which progressively rewires the network as more width multipliers are supported; and
(2) universally slimmable training (US) \cite{yu2019universally}, which randomly samples some width multipliers in each iteration.
The detailed setting of these two methods is in Appendix~\ref{apdx:pr_us}.
By comparing with Table~\ref{tbl:width}, using 
PR or US has no significant difference from using the method  in Section~\ref{sec:width_adaptive}.
	\begin{table}[htbp]
		\caption{Training $\text{DynaBERT}_\text{W}$ with PR and US.
			Average accuracy of 4 width multipliers is reported.}
		\label{tbl:pr_us}
		\centering
		\scalebox{0.92}{
			\begin{tabular}{l|ccccccccc|c}
				\hline
			  &     \texttt{MNLI-m}       &      \texttt{MNLI-mm}       &      \texttt{QQP}      &     \texttt{QNLI}      &     \texttt{SST-2}     &    \texttt{CoLA}      &     \texttt{STS-B}     &     \texttt{MRPC}      &      \texttt{RTE}      & avg. \\ \hline
				PR~\cite{Cai2020Once}      &   82.3   &  82.8  &  90.9   & 90.4 & 91.6 & 52.8  & 89.1 & 84.5  & 70.3 & 81.6   \\ \hline
				US~\cite{yu2019universally}   &      82.6       &       82.9       &     90.6     &     90.3      &      91.5      &      51.2      &      89.1      &     83.8      &     69.6     & 81.3 \\ \hline
			\end{tabular}
		}
	\end{table}

	\section{Conclusion}
	In this paper, we propose DynaBERT which can flexibly adjust its size and latency by selecting 
	sub-networks with different widths and depths. 
	DynaBERT is trained by knowledge distillation.
	We adapt the width of the BERT model by varying the number of attention heads in MHA and neurons in the intermediate layer in FFN, and adapt the depth by varying the number of Transformer layers.
	Network rewiring is also used to make the more important attention heads and neurons shared by more sub-networks.
		Experiments on various tasks
	show that under the same efficiency constraint, sub-networks extracted from the proposed DynaBERT consistently achieve better performance than the other BERT compression methods.

	
	\section*{Broader Impact}
Traditional machine learning computing relies on mobile perception and cloud computing. However, considering the speed, reliability, and cost of the data transmission process, cloud-based machine learning may cause delays in inference, user privacy leakage, and high data transmission costs. In such cases, in addition to end-cloud collaborative computing, it becomes increasingly important to run deep neural network models directly on edge.
Recently, pre-trained language models like BERT  have achieved impressive results in various natural language processing tasks. However, the BERT model contains tons of parameters,  hindering its deployment to devices with limited resources. 
The difficulty of deploying BERT to these devices lies in two aspects. Firstly,  the performances of various devices are different, and it is unclear how to deploy a BERT model suitable for each edge device based on its resource constraint.
Secondly, the resource condition of the same device under different circumstances can be quite different.  
Once the BERT model is deployed to a specific device, dynamically selecting a part of the model for inference based on the device's current resource condition is also desirable.

Motivated by this, we propose DynaBERT. Instead of compressing the BERT model to a fixed size like existing BERT compression methods, the proposed DynaBERT can adjust its size and latency by selecting a sub-network with adaptive width and depth. By allowing both adaptive width and depth, the proposed DynaBERT also enables a large number of architectural configurations of the BERT model. Moreover, once the DynaBERT is trained, no further fine-tuning is required for each sub-network, and the benefits are threefold. Firstly, we only need to train one DynaBERT model, but can deploy different sub-networks to different hardware platforms based on their performances. Secondly, once one sub-network is deployed to a specific device, this device can select the same or smaller sub-networks for inference based on its dynamic efficiency constraints.
Thirdly, different sub-networks sharing weights in one single model dramatically reduces the training and inference cost, compared to using different-sized models separately for different hardware platforms. This can reduce carbon emissions, and is thus more environmentally friendly.

Though not originally developed for compression, sub-networks of the proposed DynaBERT outperform other BERT compression methods under the same efficiency constraints like \#parameters, FLOPs, GPU and CPU latency. 
Besides, the proposed DynaBERT at its largest size often achieves better performances as $\text{BERT}_\text{BASE}$ with the same size.     A possible reason is that allowing adaptive width and depth increases the training difficulty and acts as
regularization, and so contributes positively to the performance. In this way, the proposed training method of DynaBERT also acts as a regularization method that can boost the generalization performance.

Meanwhile, we also find that the compressed sub-networks of the learned DynaBERT have good interpretability.
In order to maintain the representation power,
the attention patterns of sub-networks with smaller width or depth of the trained DynaBERT exhibit function fusion, compared to the full-sized model. Interestingly, these attention patterns even
explain the enhanced performance of DynaBERT on some tasks, e.g., enhanced ability of distinguishing linguistic acceptable and non-acceptable sentences for \texttt{CoLA}.

Besides the positive broader impacts above, since DynaBERT
enables easier deployment of BERT, it also makes the negative impacts of BERT more severe. For instance, application in dialogue systems replaces help-desks and can cause job loss. Extending our method to generative models like GPT also faces the risk of generating offensive, biased or unethical outputs.

\bibliography{paper}

\begin{thebibliography}{10}

\bibitem{bhandare2019efficient}
A.~Bhandare, V.~Sripathi, D.~Karkada, V.~Menon, S.~Choi, K.~Datta, and
  V.~Saletore.
\newblock Efficient 8-bit quantization of transformer neural machine language
  translation model.
\newblock Preprint arXiv:1906.00532, 2019.

\bibitem{Cai2020Once}
H.~Cai, C.~Gan, T.~Wang, Z.~Zhang, and S.~Han.
\newblock Once for all: Train one network and specialize it for efficient
  deployment.
\newblock In {\em International Conference on Learning Representations}, 2020.

\bibitem{clark2019electra}
K.~Clark, M.~Luong, Q.~V. Le, and C.~D. Manning.
\newblock Electra: Pre-training text encoders as discriminators rather than
  generators.
\newblock In {\em International Conference on Learning Representations}, 2019.

\bibitem{cui2019fine}
B.~Cui, Y.~Li, M.~Chen, and Z.~Zhang.
\newblock Fine-tune {BERT} with sparse self-attention mechanism.
\newblock In {\em Conference on Empirical Methods in Natural Language
  Processing}, 2019.

\bibitem{dehghani2018universal}
M.~Dehghani, S.~Gouws, O.~Vinyals, J.~Uszkoreit, and L.~Kaiser.
\newblock Universal transformers.
\newblock In {\em International Conference on Learning Representations}, 2019.

\bibitem{devlin2019bert}
J.~Devlin, M.~Chang, K.~Lee, and K.~Toutanova.
\newblock Bert: Pre-training of deep bidirectional transformers for language
  understanding.
\newblock In {\em North American Chapter of the Association for Computational
  Linguistics}, pages 4171--4186, 2019.

\bibitem{elbayad2020depthadaptive}
M.~Elbayad, J.~Gu, E.~Grave, and M.~Auli.
\newblock Depth-adaptive transformer.
\newblock In {\em International Conference on Learning Representations}, 2020.

\bibitem{fan2019reducing}
A.~Fan, E.~Grave, and A.~Joulin.
\newblock Reducing transformer depth on demand with structured dropout.
\newblock In {\em International Conference on Learning Representations}, 2019.

\bibitem{fan2020training}
A.~Fan, P.~Stock, B.~Graham, E.~Grave, R.~Gribonval, H.~Jegou, and A.~Joulin.
\newblock Training with quantization noise for extreme model compression.
\newblock Preprint arXiv:2004.07320, 2020.

\bibitem{jawahar2019does}
G.~Jawahar, B.~Sagot, and D.~Seddah.
\newblock What does bert learn about the structure of language?
\newblock In {\em Annual Meeting of the Association for Computational
  Linguistics}, 2019.

\bibitem{jiao2019tinybert}
X.~Jiao, Y.~Yin, L.~Shang, X.~Jiang, X.~Chen, L.~Li, F.~Wang, and Q.~Liu.
\newblock Tinybert: Distilling bert for natural language understanding.
\newblock Preprint arXiv:1909.10351, 2019.

\bibitem{jozefowicz2016exploring}
R.~Jozefowicz, O.~Vinyals, M.~Schuster, N.~Shazeer, and Y.~Wu.
\newblock Exploring the limits of language modeling.
\newblock Preprint arXiv:1602.02410, 2016.

\bibitem{kovaleva2019revealing}
O.~Kovaleva, A.~Romanov, A.~Rogers, and A.~Rumshisky.
\newblock Revealing the dark secrets of bert.
\newblock In {\em Conference on Empirical Methods in Natural Language
  Processing}, pages 4356--4365, 2019.

\bibitem{lan2020ALBERT}
Z.~Lan, M.~Chen, S.~Goodman, K.~Gimpel, P.~Sharma, and R.~Soricut.
\newblock Albert: A lite bert for self-supervised learning of language
  representations.
\newblock In {\em International Conference on Learning Representations}, 2020.

\bibitem{liu2020fastbert}
W.~Liu, P.~Zhou, Z.~Zhao, Z.~Wang, H.~Deng, and Q.~Ju.
\newblock Fastbert: a self-distilling bert with adaptive inference time.
\newblock In {\em Annual Meeting of the Association for Computational
  Linguistics}, 2020.

\bibitem{liu2019roberta}
Y.~Liu, M.~Ott, N.~Goyal, J.~Du, M.~Joshi, D.~Chen, O.~Levy, M.~Lewis,
  L.~Zettlemoyer, and V.~Stoyanov.
\newblock Roberta: A robustly optimized bert pretraining approach.
\newblock Preprint arXiv:1907.11692, 2019.

\bibitem{ma2019tensorized}
X.~Ma, P.~Zhang, S.~Zhang, N.~Duan, Y.~Hou, D.~Song, and M.~Zhou.
\newblock A tensorized transformer for language modeling.
\newblock In {\em Advances in Neural Information Processing Systems}, 2019.

\bibitem{mccarley2019pruning}
J.S. McCarley.
\newblock Pruning a bert-based question answering model.
\newblock Preprint arXiv:1910.06360, 2019.

\bibitem{melis2018state}
G.~Melis, C.~Dyer, and P.~Blunsom.
\newblock On the state of the art of evaluation in neural language models.
\newblock In {\em International Conference on Learning Representations}, 2018.

\bibitem{michel2019sixteen}
P.~Michel, O.~Levy, and G.~Neubig.
\newblock Are sixteen heads really better than one?
\newblock In {\em Advances in Neural Information Processing Systems}, pages
  14014--14024, 2019.

\bibitem{molchanov2016pruning}
P.~Molchanov, S.~Tyree, T.~Karras, T.~Aila, and J.~Kautz.
\newblock Pruning convolutional neural networks for resource efficient
  inference.
\newblock In {\em International Conference on Learning Representations}, 2017.

\bibitem{rajpurkar2016squad}
P.~Rajpurkar, J.~Zhang, K.~Lopyrev, and P.~Liang.
\newblock Squad: 100,000+ questions for machine comprehension of text.
\newblock In {\em Conference on Empirical Methods in Natural Language
  Processing}, pages 2383--2392, 2016.

\bibitem{rogers2020primer}
A.~Rogers, O.~Kovaleva, and A.~Rumshisky.
\newblock A primer in bertology: What we know about how bert works.
\newblock Preprint arXiv:2002.12327, 2020.

\bibitem{sanh2019distilbert}
V.~Sanh, L.~Debut, J.~Chaumond, and T.~Wolf.
\newblock Distilbert, a distilled version of bert: smaller, faster, cheaper and
  lighter.
\newblock Preprint arXiv:1910.01108, 2019.

\bibitem{shen2019q}
S.~Shen, Z.~Dong, J.~Ye, L.~Ma, Z.~Yao, A.~Gholami, M.~W. Mahoney, and
  K.~Keutzer.
\newblock Q-bert: Hessian based ultra low precision quantization of bert.
\newblock In {\em AAAI Conference on Artificial Intelligence}, 2020.

\bibitem{subramani2019can}
N.~Subramani, S.~Bowman, and K.~Cho.
\newblock Can unconditional language models recover arbitrary sentences?
\newblock In {\em Advances in Neural Information Processing Systems}, pages
  15258--15268, 2019.

\bibitem{sun2019patient}
S.~Sun, Y.~Cheng, Z.~Gan, and J.~Liu.
\newblock Patient knowledge distillation for bert model compression.
\newblock In {\em Conference on Empirical Methods in Natural Language
  Processing}, pages 4314--4323, 2019.

\bibitem{sun2020mobilebert}
Z.~Sun, H.~Yu, X.~Song, R.~Liu, Y.~Yang, and D.~Zhou.
\newblock Mobilebert: Task-agnostic compression of bert by progressive
  knowledge transfer.
\newblock In {\em Annual Meeting of the Association for Computational
  Linguistics}, 2020.

\bibitem{vaswani2017attention}
A.~Vaswani, N.~Shazeer, N.~Parmar, J.~Uszkoreit, L.~Jones, A.~N. Gomez,
  {\L}.~Kaiser, and I.~Polosukhin.
\newblock Attention is all you need.
\newblock In {\em Advances in neural information processing systems}, pages
  5998--6008, 2017.

\bibitem{voita2019analyzing}
E.~Voita, D.~Talbot, F.~Moiseev, R.~Sennrich, and I.~Titov.
\newblock Analyzing multi-head self-attention: Specialized heads do the heavy
  lifting, the rest can be pruned.
\newblock In {\em Annual Meeting of the Association for Computational
  Linguistics}, 2019.

\bibitem{wang2019glue}
A.~Wang, A.~Singh, J.~Michael, F.~Hill, O.~Levy, and S.~R. Bowman.
\newblock Glue: A multi-task benchmark and analysis platform for natural
  language understanding.
\newblock In {\em International Conference on Learning Representations}, 2019.

\bibitem{hanruiwang2020hat}
H.~Wang, Z.~Wu, Z.~Liu, H.~Cai, L.~Zhu, C.~Gan, and S.~Han.
\newblock Hat: Hardware-aware transformers for efficient natural language
  processing.
\newblock In {\em Annual Meeting of the Association for Computational
  Linguistics}, 2020.

\bibitem{wang2020minilm}
W.~Wang, F.~Wei, L.~Dong, H.~Bao, N.~Yang, and M.~Zhou.
\newblock Minilm: Deep self-attention distillation for task-agnostic
  compression of pre-trained transformers.
\newblock Preprint arXiv:2002.10957, 2020.

\bibitem{xin2020deebert}
Ji~Xin, R.~Tang, J.~Lee, Y.~Yu, and J.~Lin.
\newblock Deebert: Dynamic early exiting for accelerating bert inference.
\newblock In {\em Annual Meeting of the Association for Computational
  Linguistics}, 2020.

\bibitem{yu2019network}
J.~Yu and T.~S. Huang.
\newblock Network slimming by slimmable networks: Towards one-shot architecture
  search for channel numbers.
\newblock Preprint arXiv:1903.11728, 2019.

\bibitem{yu2019universally}
J.~Yu and T.~S. Huang.
\newblock Universally slimmable networks and improved training techniques.
\newblock In {\em IEEE International Conference on Computer Vision}, pages
  1803--1811, 2019.

\bibitem{yu2018slimmable}
J.~Yu, L.~Yang, N.~Xu, J.~Yang, and T.~Huang.
\newblock Slimmable neural networks.
\newblock In {\em International Conference on Learning Representations}, 2018.

\bibitem{zafrir2019q8bert}
O.~Zafrir, G.~Boudoukh, P.~Izsak, and M.~Wasserblat.
\newblock Q8bert: Quantized 8bit bert.
\newblock Preprint arXiv:1910.06188, 2019.

\bibitem{zhou2020bert}
W.~Zhou, C.~Xu, T.~Ge, J.~McAuley, K.~Xu, and F.~Wei.
\newblock Bert loses patience: Fast and robust inference with early exit.
\newblock In {\em Advances in Neural Information Processing Systems}, 2020.

\end{thebibliography}
\bibliographystyle{plain}	

\clearpage
\appendix 
\section{Layer Pruning Strategy and Hidden States Matching}
\label{apdx:select_layer}
In the training of DynaBERT with adaptive width and depth, 
when $m_d<1$,
we use the  ``Every Other'' strategy in \cite{fan2019reducing} and drop layers evenly to get a balanced network.
Specifically, 
for depth multiplier $m_d$ (i.e., prune layers with a rate $1-m_d$),
we drop layers at depth $d$ which satisfies $\text{mod}(d, \frac{1}{1-m_d}) \equiv 0$,
because the lower layers in the student network which are found to change less from pre-training to fine-tuning~\cite{kovaleva2019revealing}.
We then match the hidden states of the remaining layers
with those
from all layers in the teacher model except those at depth $d$ which satisfies $\text{mod}(d+1, \frac{1}{1-m_d}) \equiv 0$.
In this way, we  
keep the knowledge learned in the last layer of the teacher network which is shown to be important in \cite{wang2020minilm}.
For a BERT model with 12 Transformer layers indexed by $1,2,3,\cdots, 12$, when $m_d=0.75$, we 
drop the layers with indices $4,8,12$ of the student model.
Then we match hidden states of the remaining 9 layers $L_S = \{1,2,3,4,5,6,7,8,9\}$  in  the student with those indexed  $L_T=\{1,2,4,5,6,8,9,10,12\}$ from the teacher network.
When  $m_d=0.5$, we 
drop the layers indexed $2,4,6,8,10,12$ of the student model.
Then we match hidden states of the kept 6 layers $L_S = \{1,2,3,4,5,6\}$  in  the student with those indexed $L_T=\{2,4,6,8,10,12\}$ from the teacher network.
The loss $\ell_{hidn}$  is computed as
\[
\ell'_{hidn}( \H^{(m_w,m_d)},\H^{(m_w)}) = \sum\nolimits_{l,l'  \in L_S, L_T} \text{MSE}( \H_l^{(m_w,m_d)},\H_{l'}^{(m_w)}).
\]

\section{More Experiment Settings}

\subsection{Description of Data sets in  the GLUE benchmark}
\label{apdx:glue_data}
The GLUE benchmark~\cite{wang2019glue} is a collection of diverse natural language understanding tasks,
including textual entailment (\texttt{RTE} and \texttt{MNLI}), question answering (\texttt{QNLI}), similarity and paraphrase (\texttt{MRPC}, \texttt{QQP}, \texttt{STS-B}), sentiment analysis (\texttt{SST-2}) and linguistic acceptability (\texttt{CoLA}).
For \texttt{MNLI}, we use both the matched (\texttt{MNLI-m}) and mismatched (\texttt{MNLI-mm}) sections.
We do not experiment on Winograd Schema (\texttt{WNLI}) because even  a majority baseline outperforms many methods on it.
%
We use the default train/validation/test splits from the official website\footnote{\url{https://gluebenchmark.com/tasks}}.

\subsection{Hyperparameters}
\label{apdx:hyper}
\paragraph{GLUE benchmark.}
On the GLUE benchmark, the detailed hyperparameters for training $\text{DynaBERT}_\text{W}$  in Section~\ref{sec:width_adaptive} and DynaBERT in Section~\ref{sec:depth_width} are shown in Table~\ref{tbl:hyperparam}. 
The same hyperparameters as in  Table~\ref{tbl:hyperparam} are used for DynaRoBERTa.

\begin{table}[htbp]
	\centering
	\caption{Hyperparameters for different stages in training DynaBERT and DynaRoBERTa on  the GLUE benchmark.}
	\label{tbl:hyperparam}
	\scalebox{0.9}{
		\begin{tabular}{c|c|c|c}
			\hline
			&      $\text{DynaBERT}_\text{W}$ & \multicolumn{2}{|c}{DynaBERT} \\ \cline{2-4}            
			&    Width-adaptive    & Width- and depth-adaptive &  Final fine-tuning   \\ \hline
			Width mutipliers           & [1.0, 0.75,0.5,0.25] &   [1.0, 0.75,0.5,0.25]    & [1.0, 0.75,0.5,0.25] \\
			Depth multipliers          &          1           &      [1.0, 0.75,0.5]      &   [1.0, 0.75,0.5]    \\
			Batch Size              &          32          &            32             &          32          \\
			Learning Rate            &        $2e-5$        &          $2e-5$           &        $2e-5$        \\
			Warmup Steps             &          0           &             0             &          0           \\
			Learning Rate Decay         &        Linear        &          Linear           &        Linear        \\
			Weight Decay             &          0           &             0             &          0           \\
			Gradient Clipping          &          1           &             1             &          1           \\
			Dropout               &         0.1          &            0.1            &         0.1          \\
			Attention Dropout          &         0.1          &            0.1            &         0.1          \\
			Distillation             &          y           &             y             &          n           \\
			$\lambda_1, \lambda_2$        &       1.0, 0.1       &          1.0,1.0          &          -           \\
			Data augmentation          &          y           &             y             &          n           \\
			Training Epochs (MNLI, QQP)        &          1           &             1             &          3           \\
			Training Epochs (other data sets)     &          3           &             3             &          3           \\ \hline
		\end{tabular}
	}
\end{table}

\paragraph{SQuAD.}
Since $\mathcal{L}_{emb}+\mathcal{L}_{hidn}$ is several magnitudes larger than $\mathcal{L}_{pred}$ in this task, 
for both  $\text{DynaBERT}_\text{W}$ and DynaBERT,
we separate the training into two stages, i.e.,  first using $\mathcal{L}_{emb}+\mathcal{L}_{hidn}$ as the objective and then $\mathcal{L}_{pred}$.
When training with objective $\mathcal{L}_{emb}+\mathcal{L}_{hidn}$, we use the augmented data from \cite{jiao2019tinybert} and train for 2 epochs.
When training with objective $\mathcal{L}_{pred}$, we use the original training data and train for 10 epochs.
The batch size is 12
throughout the training process.
The other hyperparameters are the same as the GLUE benchmark in Table~\ref{tbl:hyperparam}. 

\subsection{FLOPs and Latency}
\label{apdx:efficiency_constraint}
To count the floating-point operations (FLOPs), we follow the setting in  \cite{clark2019electra} and infer FLOPs with batch size 1 and sequence length 128.
Unlike \cite{clark2019electra}, we do
not count the operations in the embedding lookup because the inference time in this part is negligible compared to that in the Transformer layers \cite{sun2020mobilebert}.
To evaluate the inference speed on GPU, we
follow~\cite{jiao2019tinybert},	and experiment on the \texttt{QNLI} training set with batch size  128 and sequence length 128. The numbers are the average running time of 100 batches on an Nvidia K40 GPU.
To evaluate the inference speed on CPU, we 
experiment on Kirin 810 A76 ARM CPU with batch size  1 and sequence length  128.  

\section{More Experiment Results}

\subsection{More Results on the GLUE Benchmark}
\label{apdx:more_glue}

\paragraph{Test Set Results.}
Table~\ref{tbl:glue_test} shows the test set results.
Again, the proposed DynaBERT achieves comparable accuracy as $\text{BERT}_\text{BASE}$ with the same size.
Interestingly, the proposed DynaRoBERTa outperforms $\text{RoBERTa}_\text{BASE}$ on seven out of eight tasks.
A possible reason is that allowing adaptive width and depth increases the training difficulty and acts as
regularization, and so contributes positively to the performance.

\begin{table}[htbp]
	\vspace{-0.15in}
	\caption{Test set results of the GLUE benchmark. }
	\centering
	\label{tbl:glue_test}
	\scalebox{0.78}{
		\begin{tabular}{l|ccccccccc}
			\hline
			& \texttt{MNLI-m} & \texttt{MNLI-mm} & \texttt{QQP}  & \texttt{QNLI} & \texttt{SST-2} & \texttt{CoLA} & \texttt{STS-B} & \texttt{MRPC} & \texttt{RTE}   \\ \hline
			$\text{BERT}_\text{BASE}$                        &  84.6  &  83.6   & 71.9 & 90.7 & 93.4  & 51.5 & 85.2  &     87.5      & 69.6 \\
			DynaBERT ($m_w,m_d=1,1$)    &  84.5  &  84.1   & 72.1 & 91.3 &  93.0   & 54.9 & 84.4  &     87.9      & 69.9 \\
			$\text{RoBERTa}_\text{BASE}$                     &  86.0  &  85.4   & 70.9 & 92.5 & 94.6  & 50.5 & 88.1  &     90.0      &  73.0  \\
			DynaRoBERTa ($m_w,m_d=1,1$) &  86.9  &  86.7   & 71.9 & 92.5 & 94.7  & 54.1 & 88.4  &     90.8      & 73.7 \\ \hline
		\end{tabular}
	}
	\vspace{-0.1in}
\end{table}

\begin{figure}[htbp]
	\centering
	\subfloat{
		\includegraphics[width=0.95\textwidth]{figures/comparison/legend.pdf}}
	\\
	\addtocounter{subfigure}{-1}
	\vspace{-0.1in}
	\subfloat{
		\includegraphics[width=0.23\textwidth]{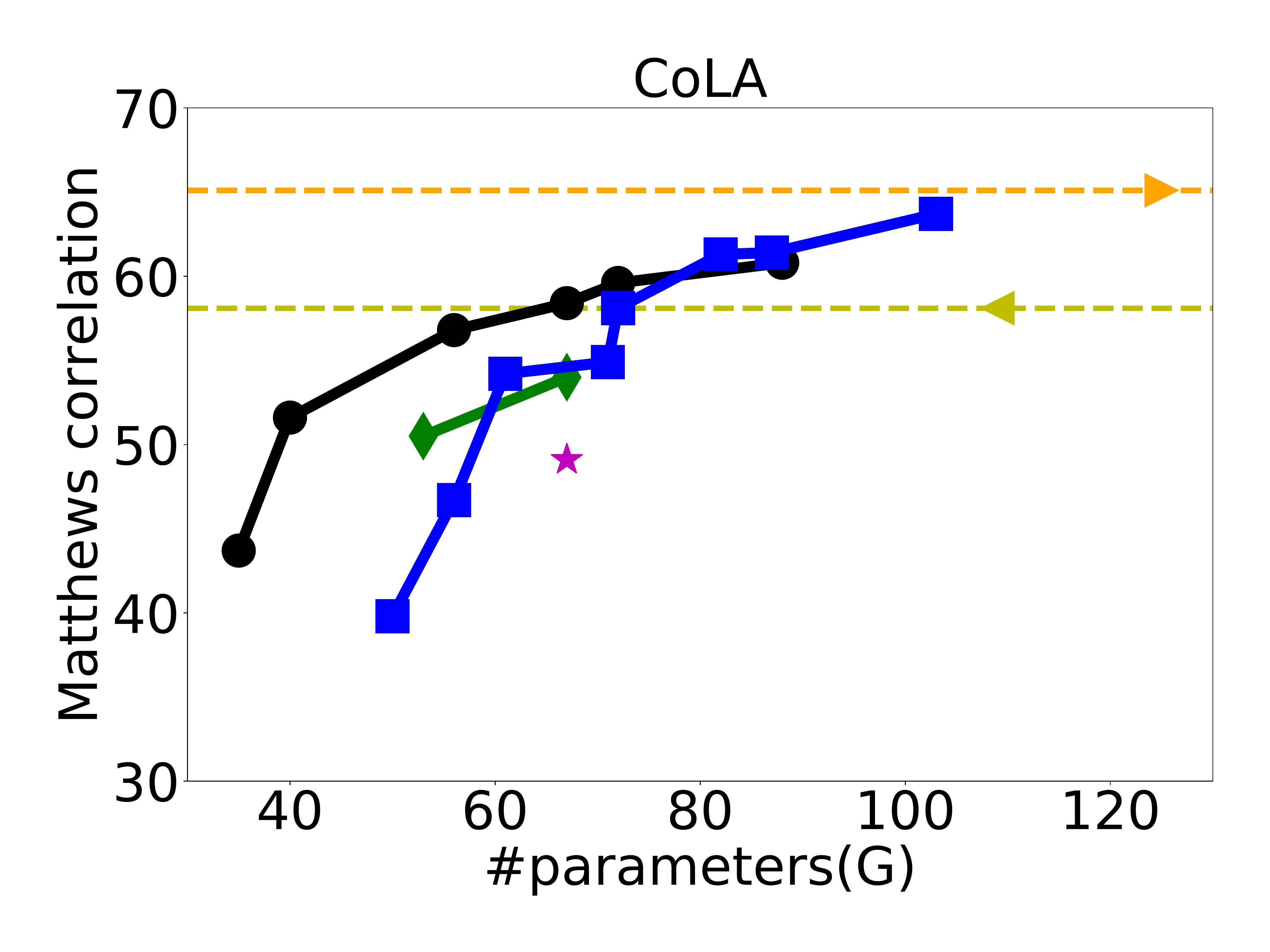}
		\includegraphics[width=0.23\textwidth]{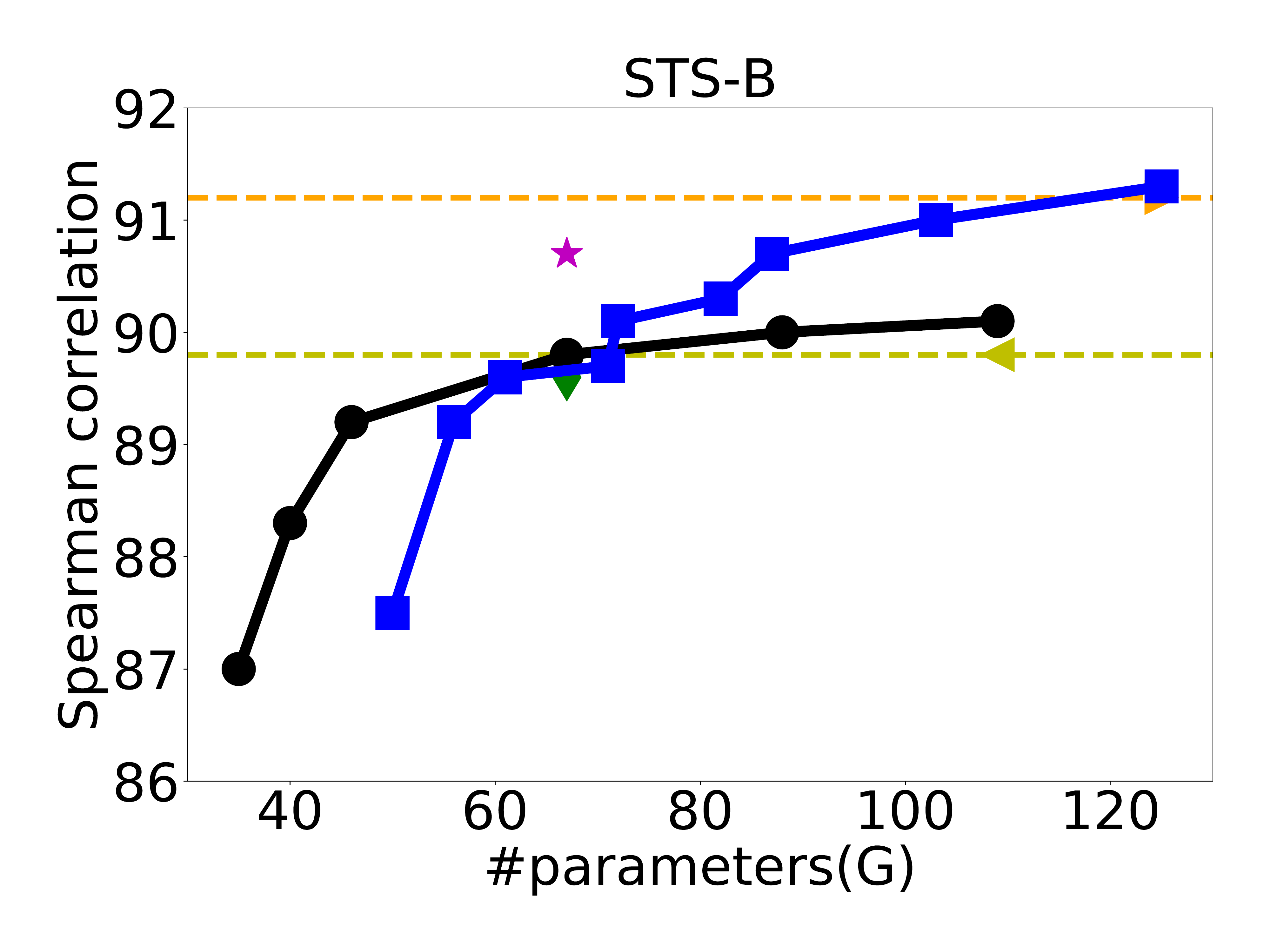}
		\includegraphics[width=0.23\textwidth]{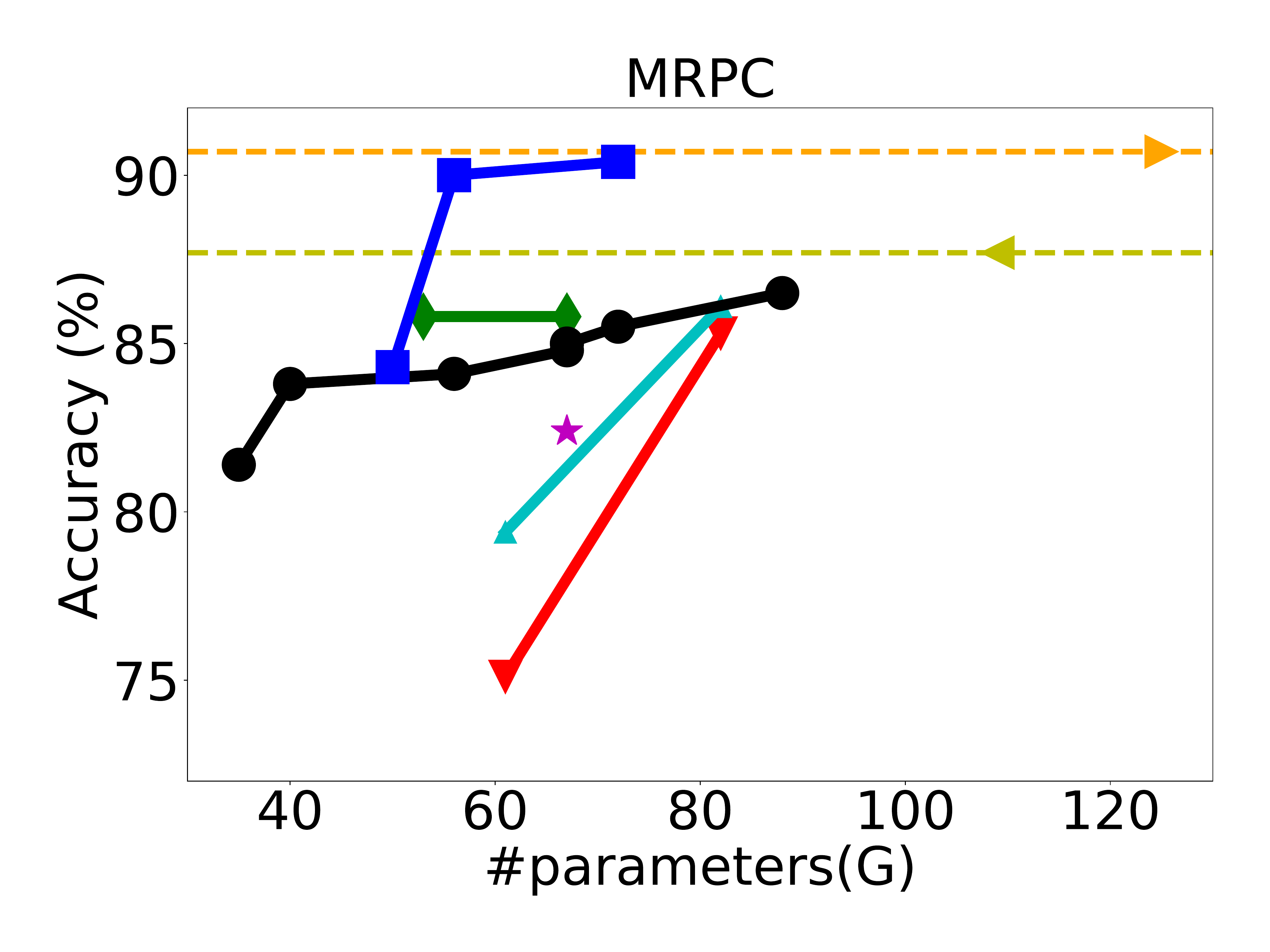}
		\includegraphics[width=0.23\textwidth]{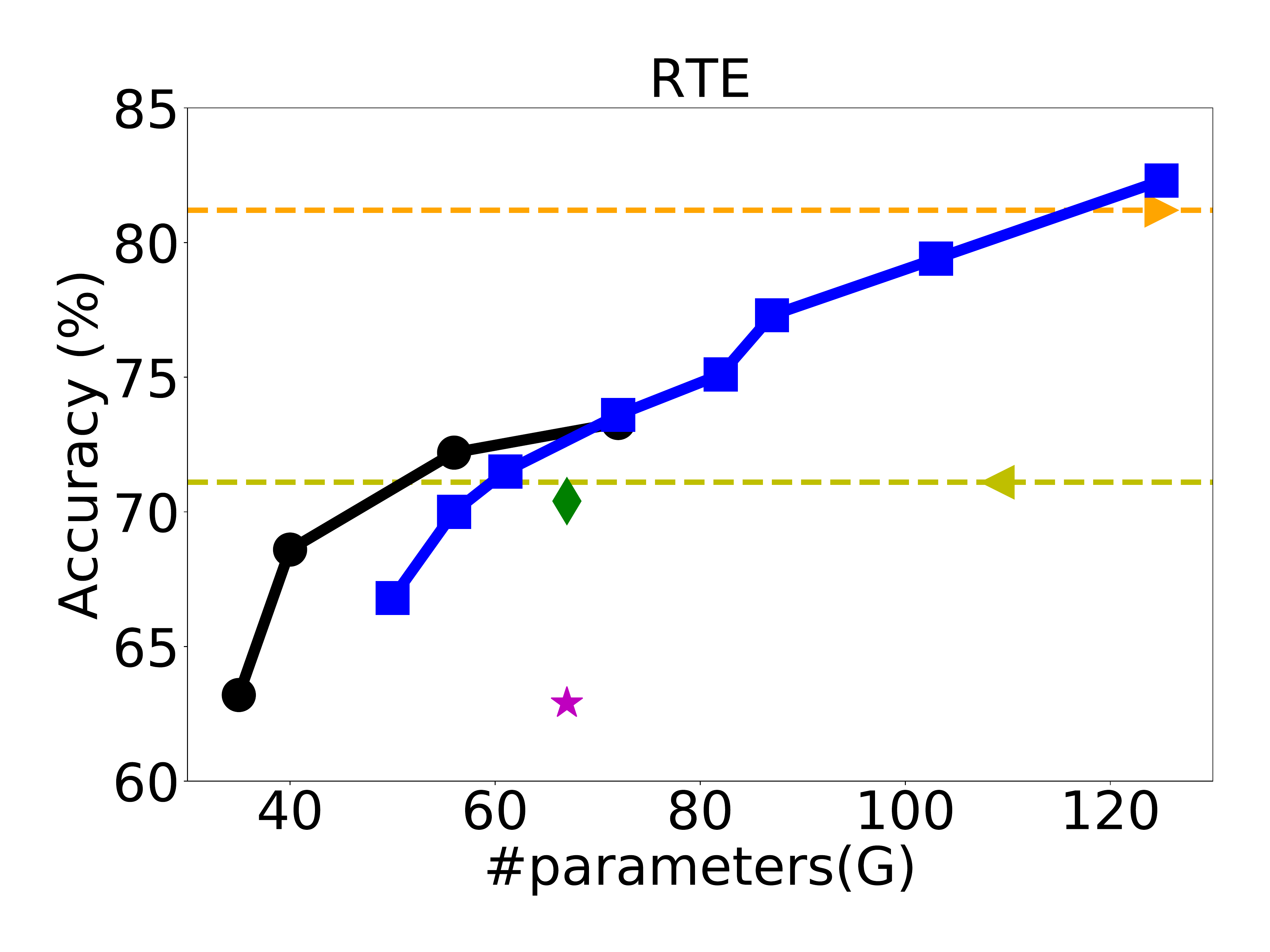}
	}
	\addtocounter{subfigure}{-1}
	\vspace{-0.15in}
	\\
	\subfloat[\#Parameters(G).\label{fig:cola_param}]
	{
		\includegraphics[width=0.23\textwidth]{figures/comparison/mnli_val_param.pdf}
		\includegraphics[width=0.23\textwidth]{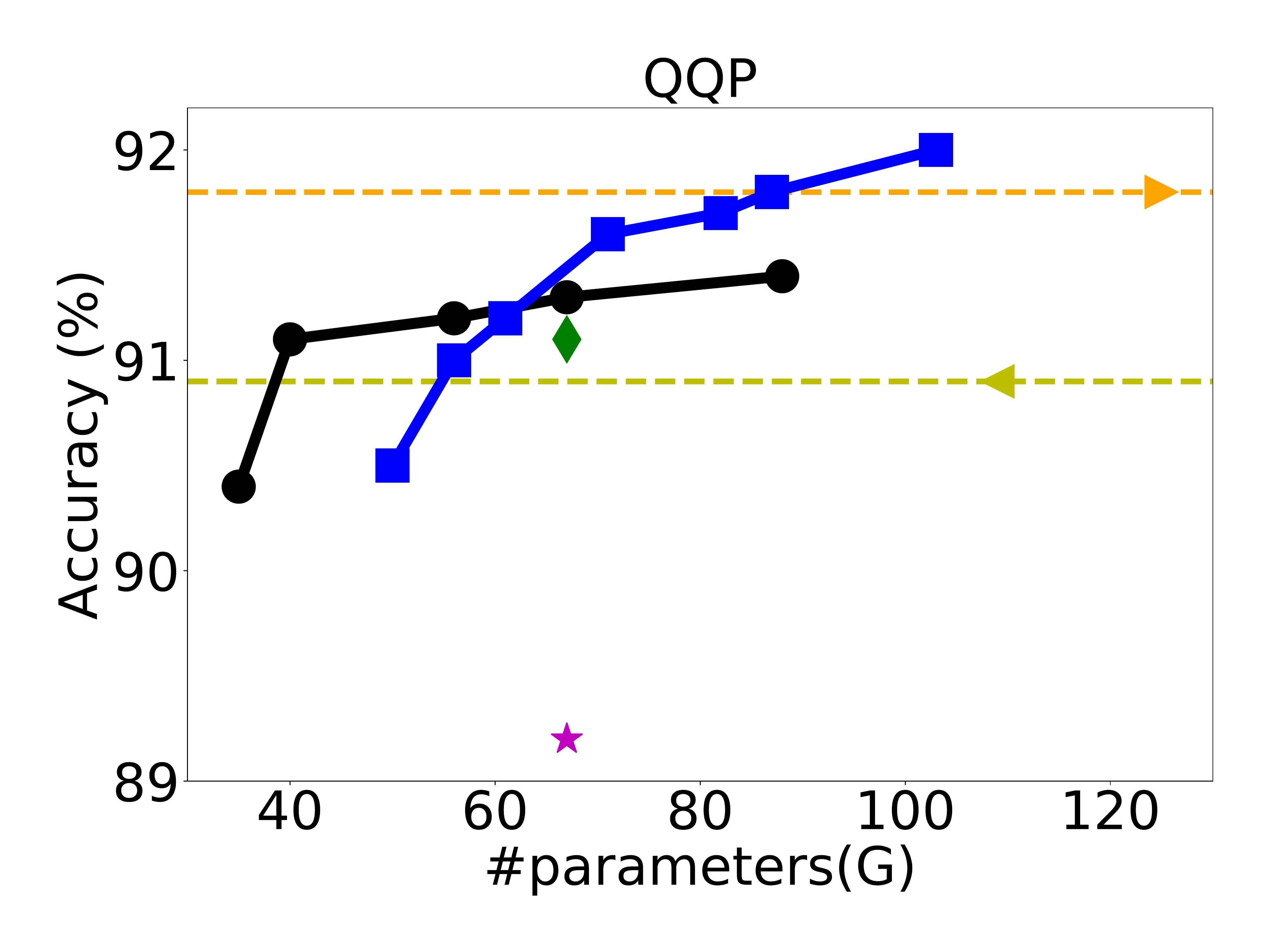}
		\includegraphics[width=0.23\textwidth]{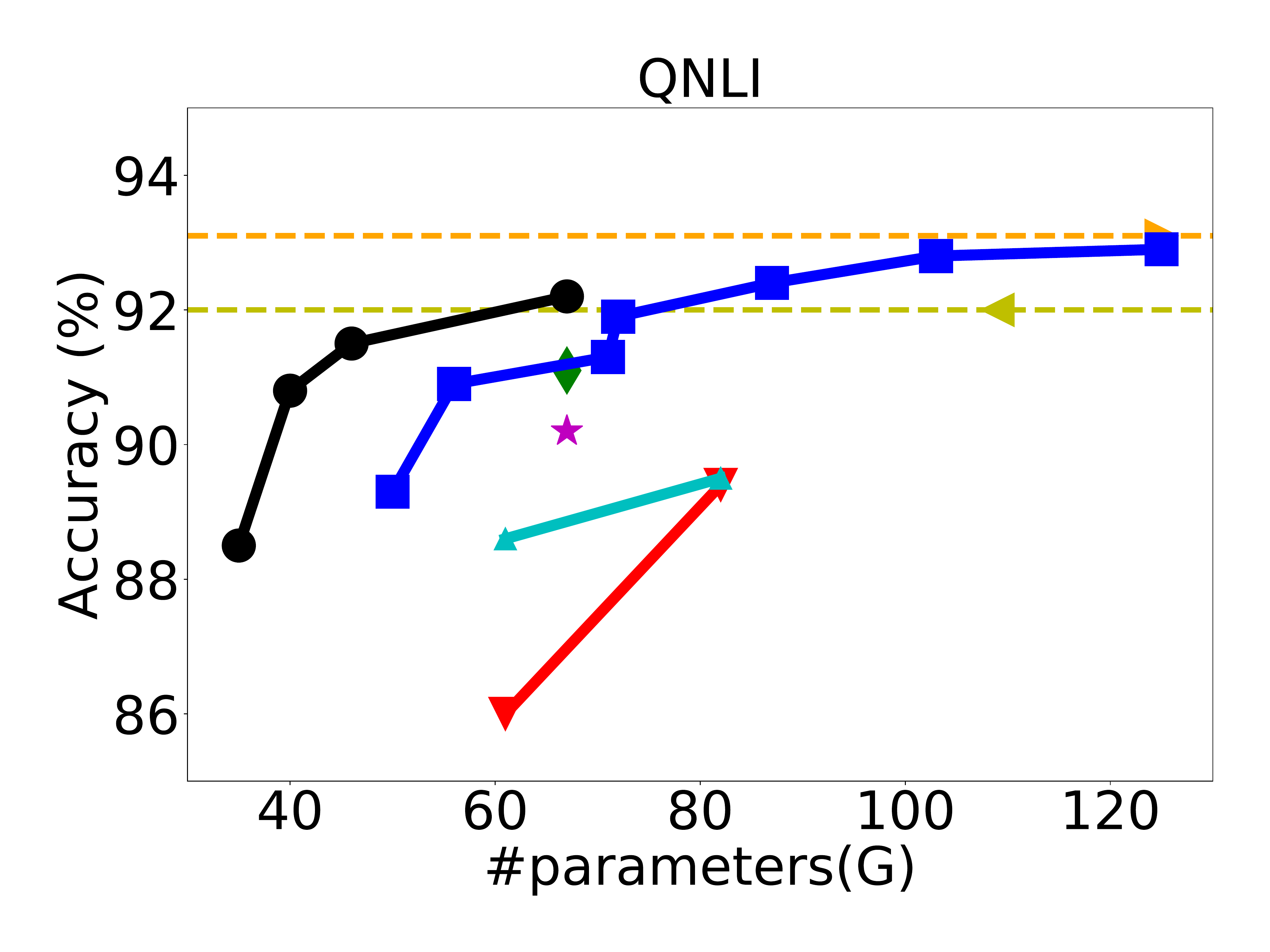}
		\includegraphics[width=0.23\textwidth]{figures/comparison/sst2_val_param.pdf}
	}
	\vspace{-0.15in}
	\\
	\subfloat{
		\includegraphics[width=0.23\textwidth]{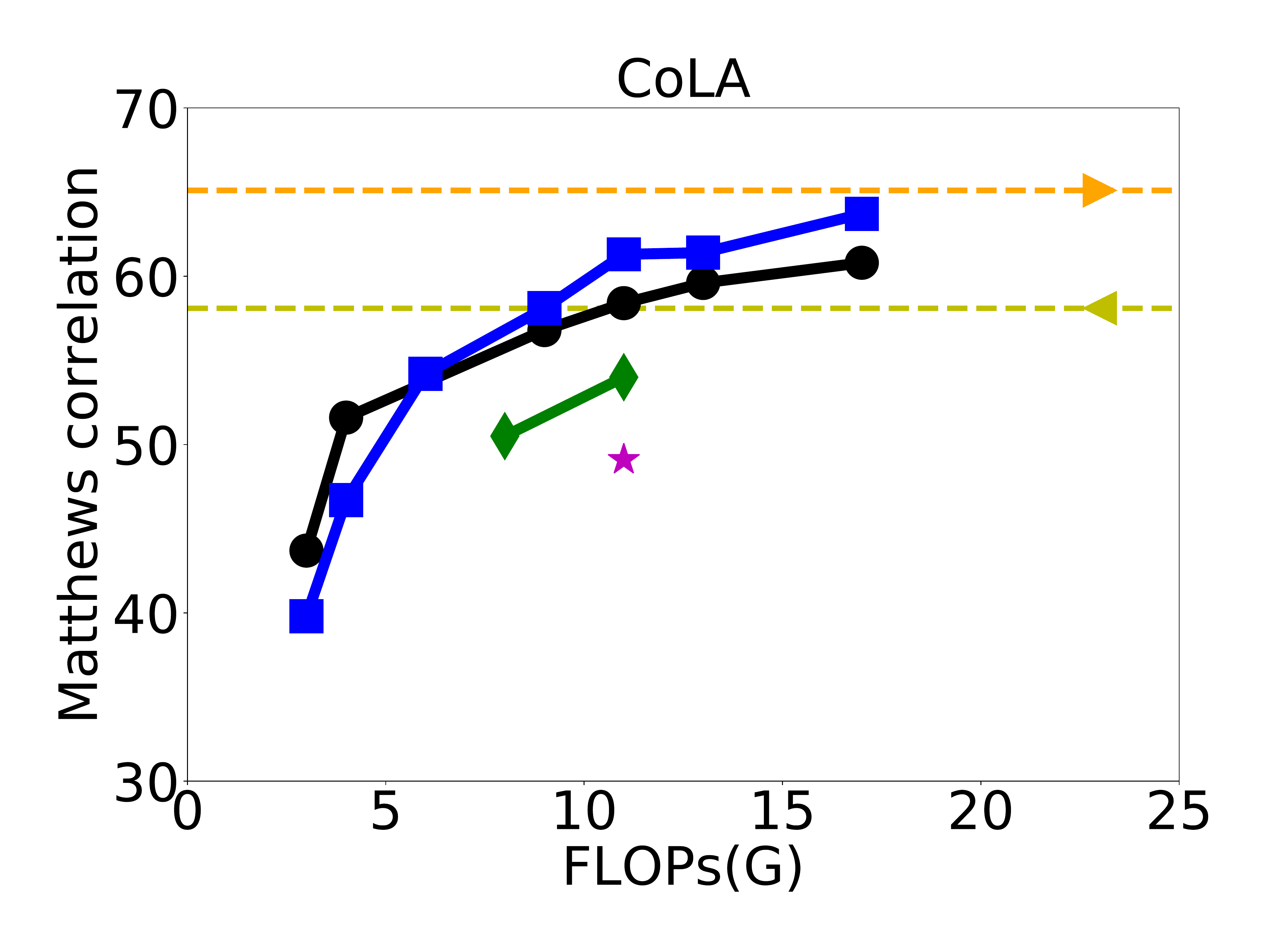}
		\includegraphics[width=0.23\textwidth]{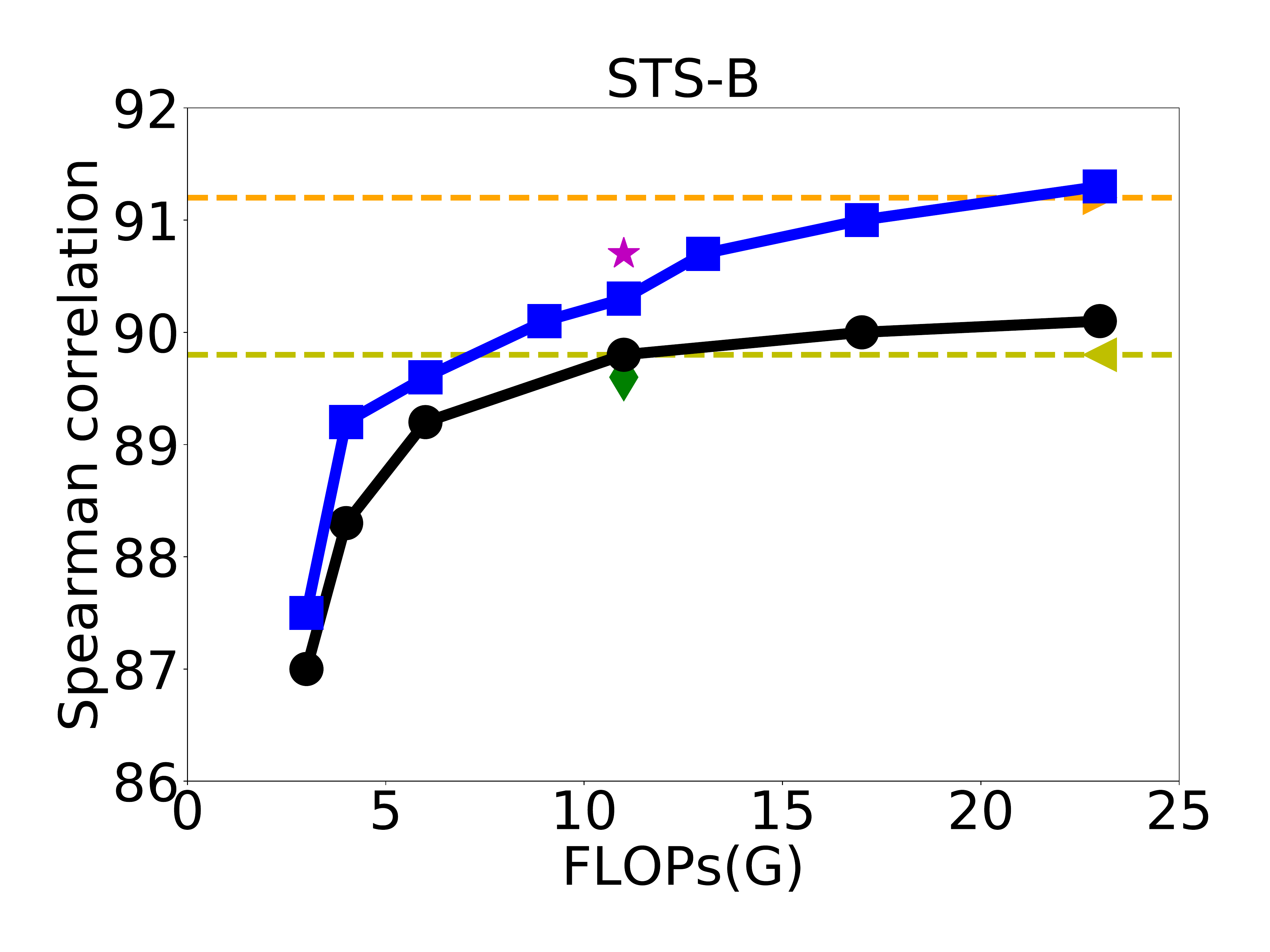}
		\includegraphics[width=0.23\textwidth]{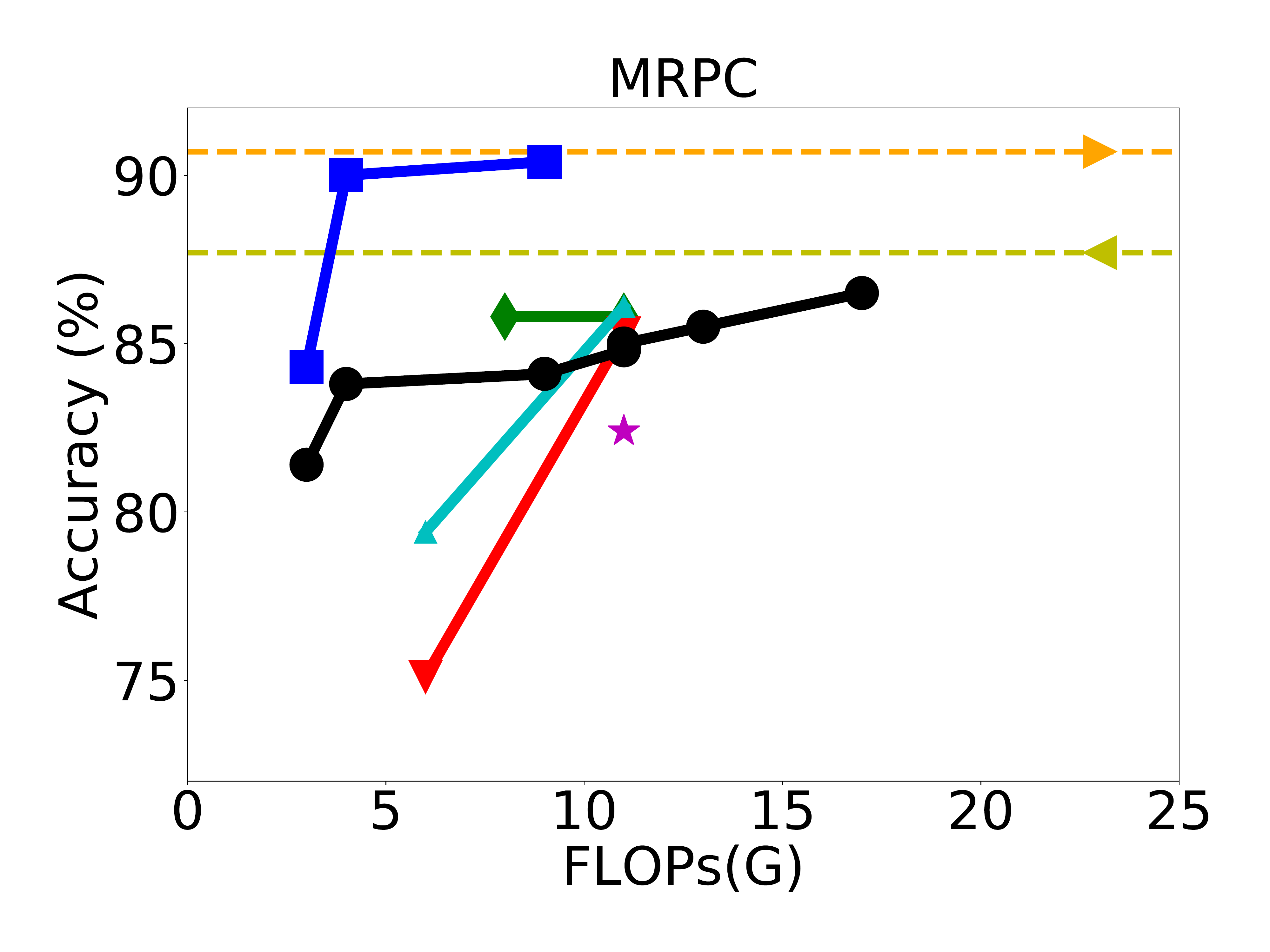}
		\includegraphics[width=0.23\textwidth]{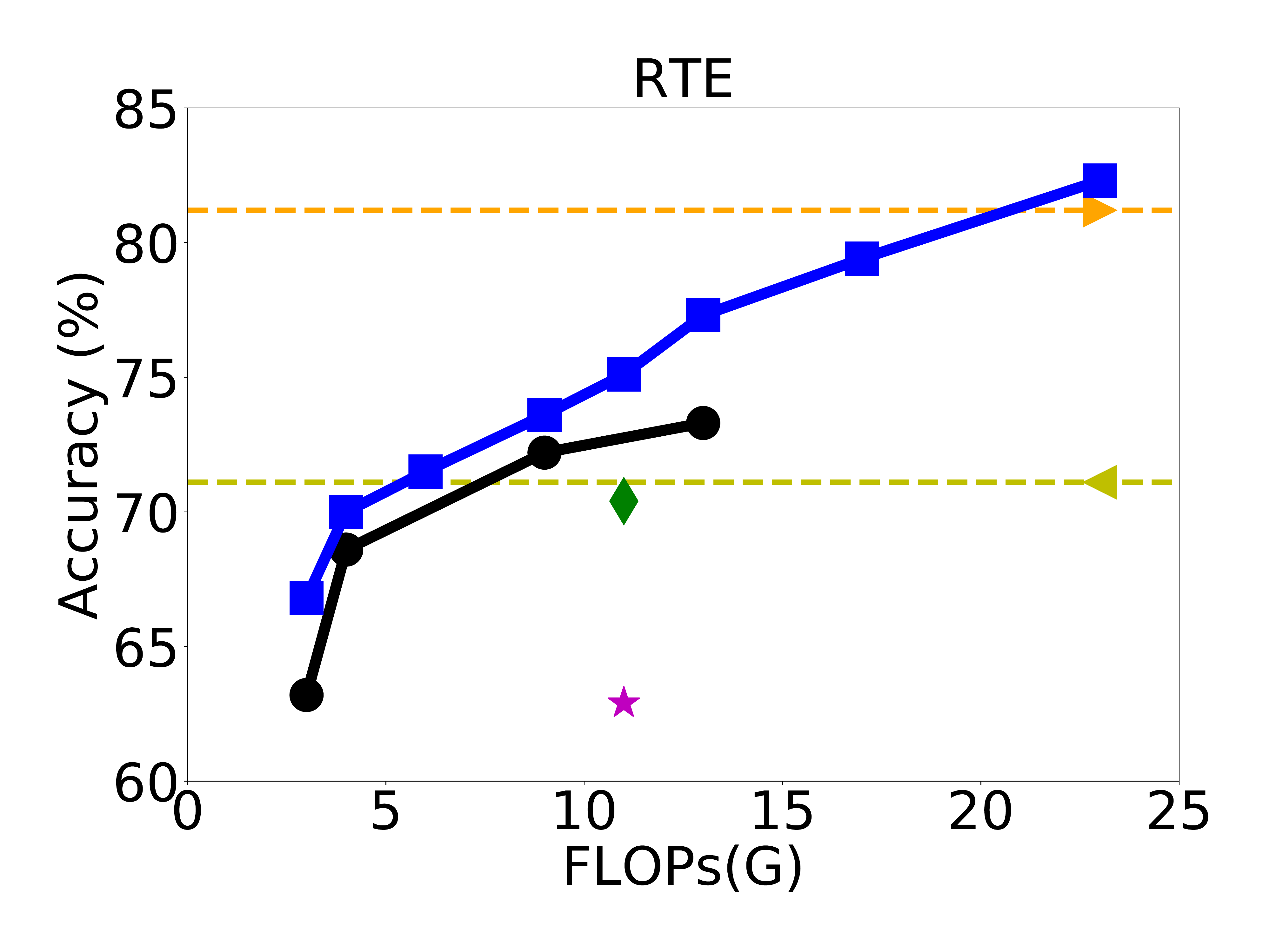}
	}
	\addtocounter{subfigure}{-1}
	\vspace{-0.15in}
	\\
	\subfloat[FLOPs(G).\label{fig:cola_flops}]
	{
		\includegraphics[width=0.23\textwidth]{figures/comparison/mnli_val_flops.pdf}
		\includegraphics[width=0.23\textwidth]{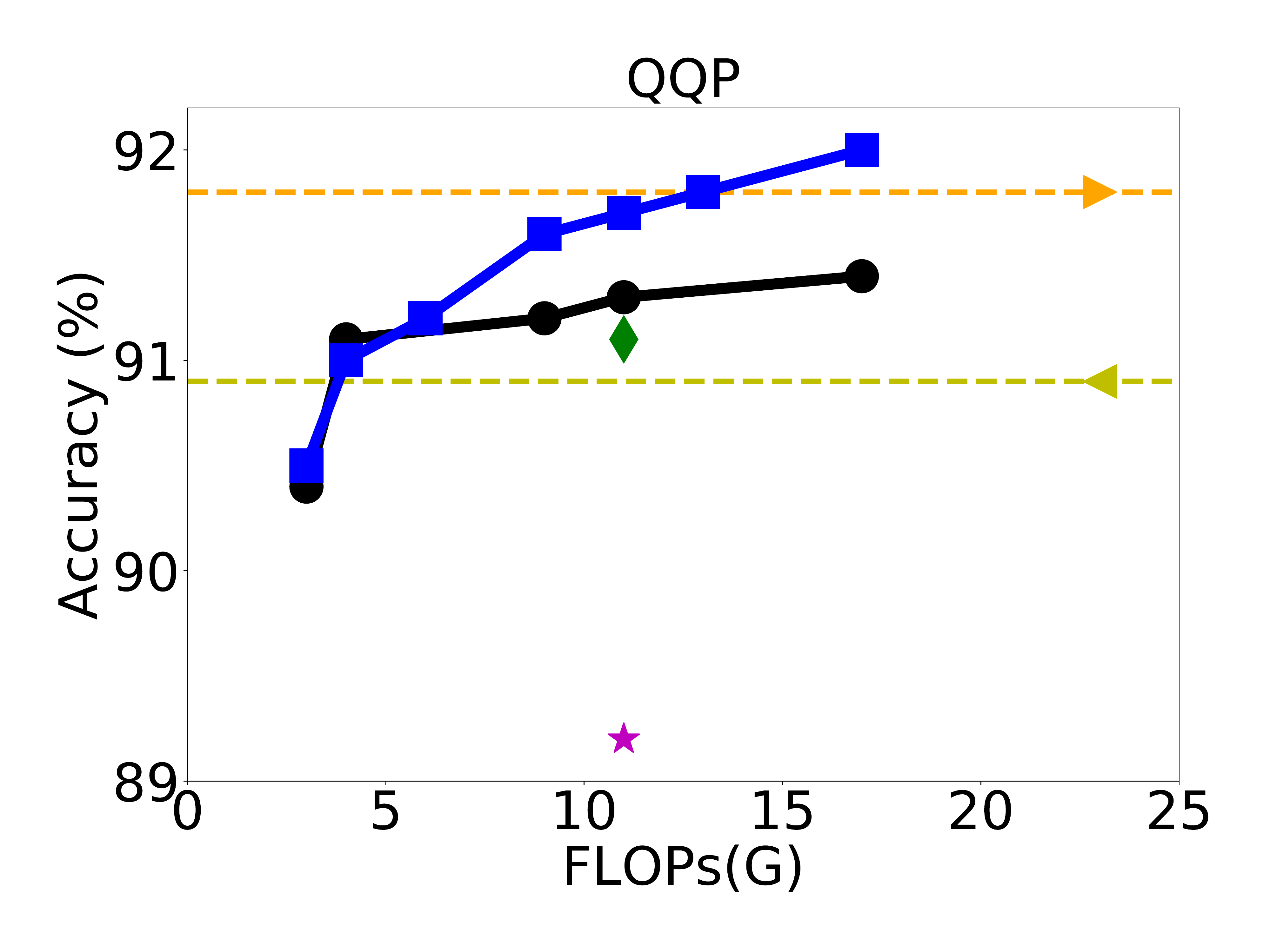}
		\includegraphics[width=0.23\textwidth]{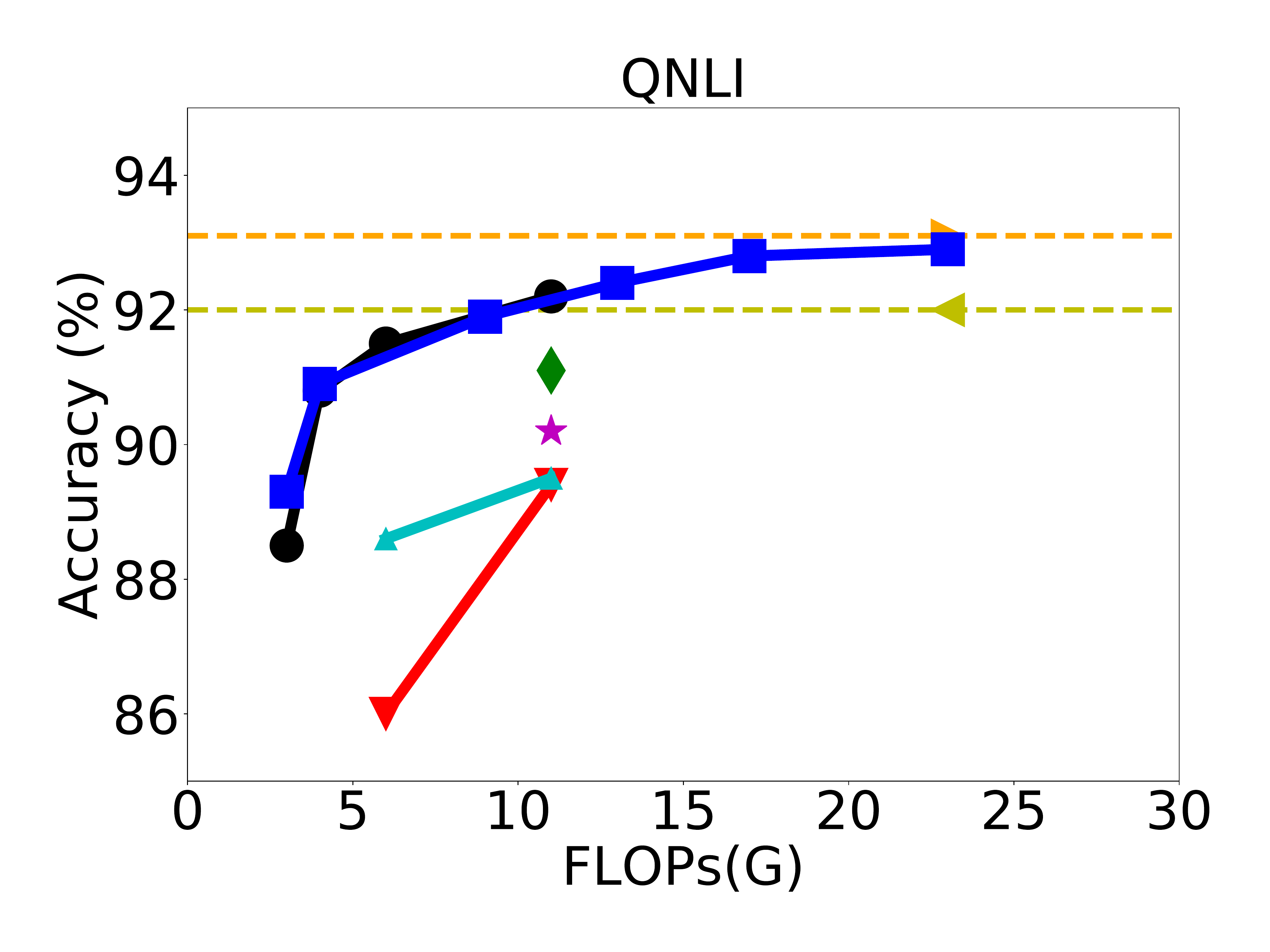}
		\includegraphics[width=0.23\textwidth]{figures/comparison/sst2_val_flops.pdf}
	}
	\vspace{-0.15in}
	\addtocounter{subfigure}{-1}
	\\
	\subfloat{
		\includegraphics[width=0.23\textwidth]{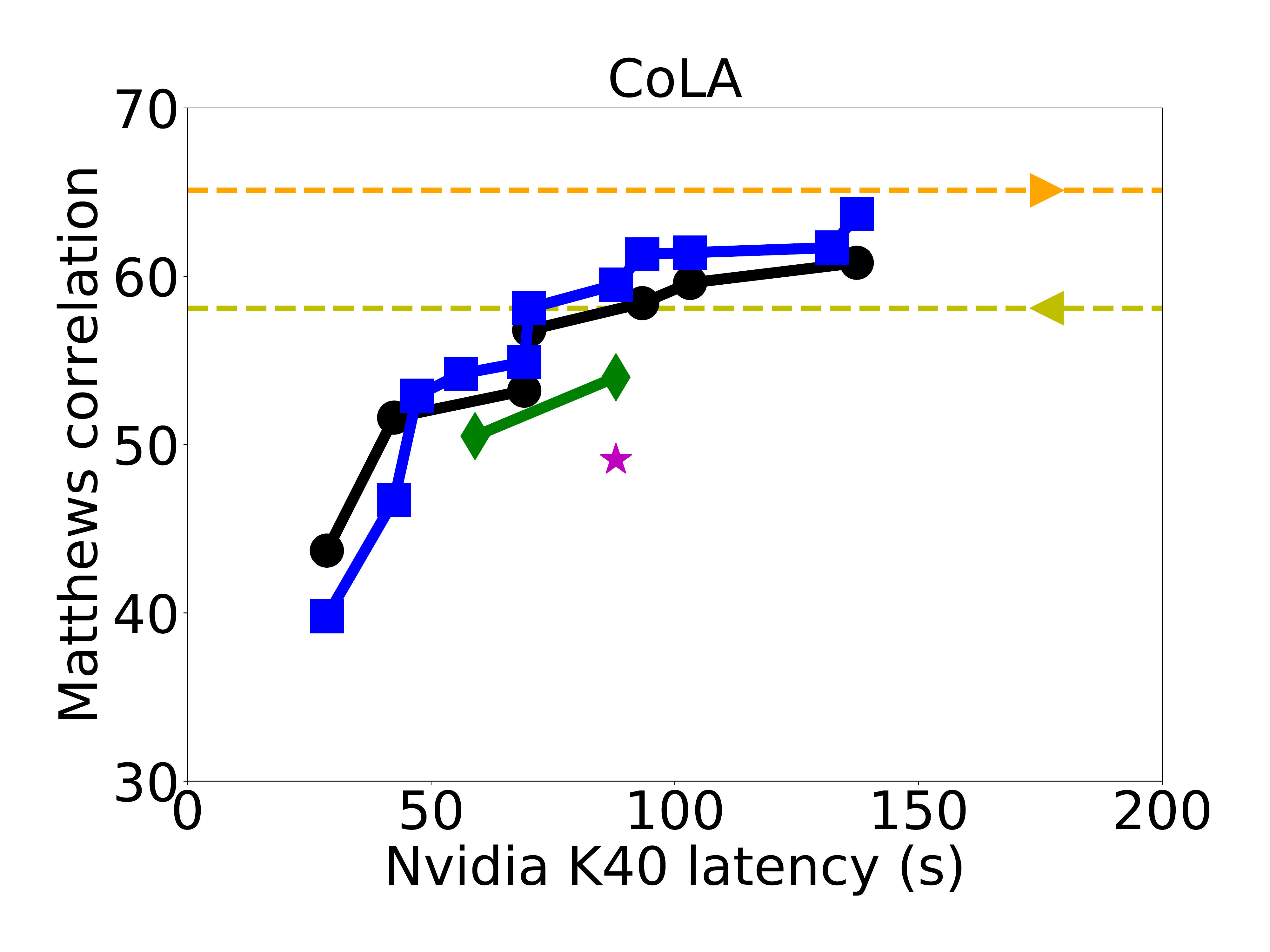}
		\includegraphics[width=0.23\textwidth]{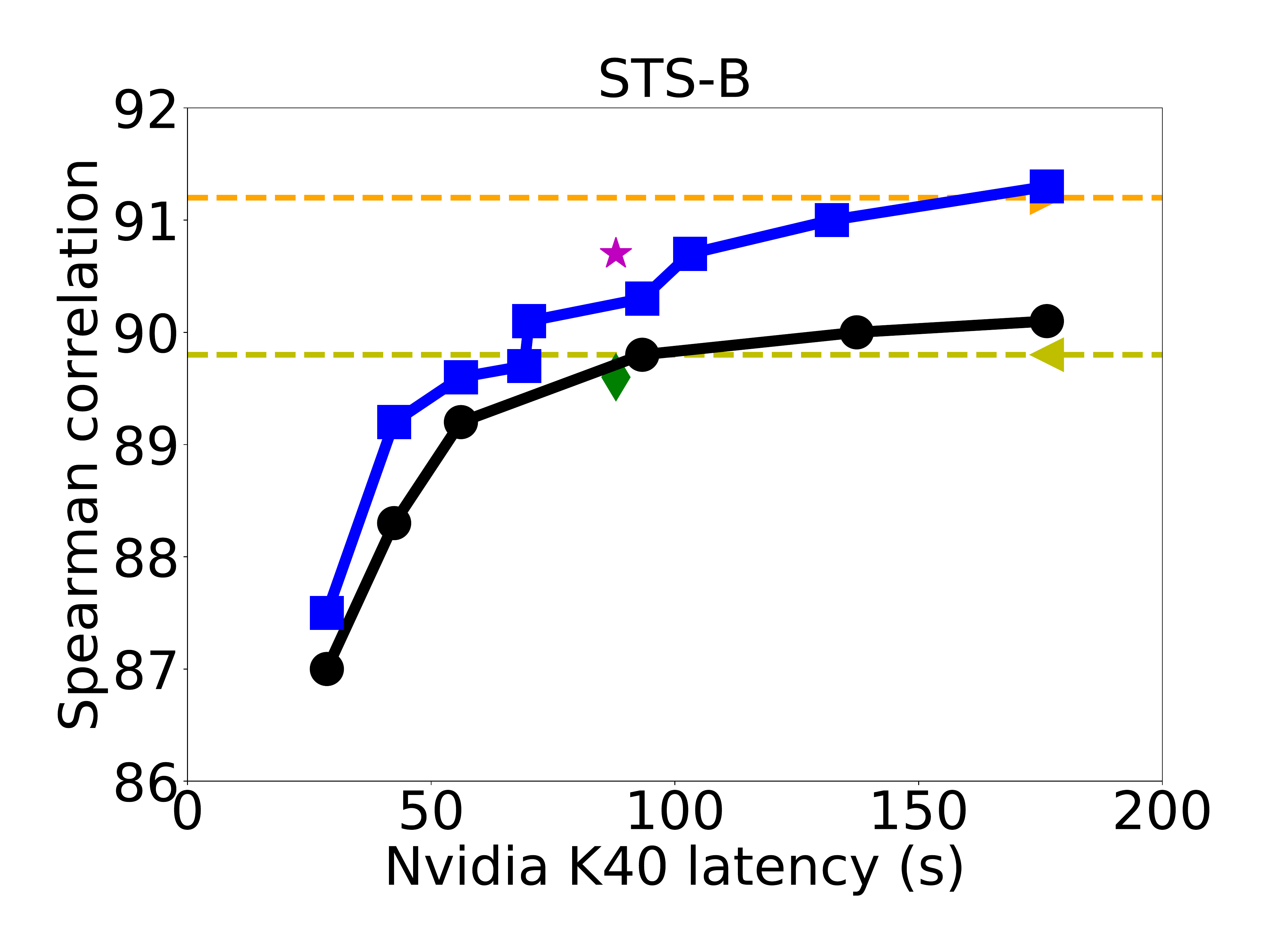}
		\includegraphics[width=0.23\textwidth]{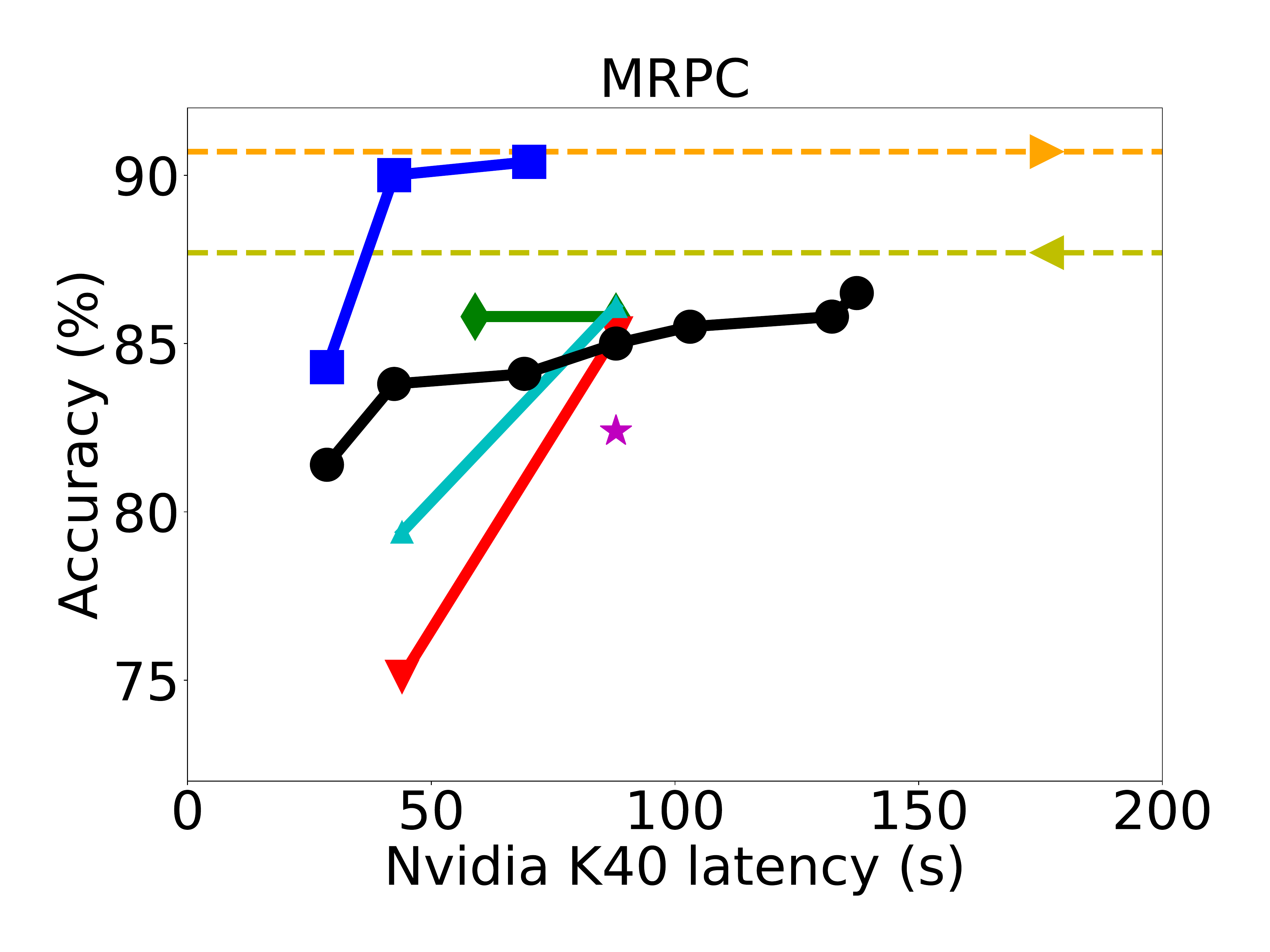}
		\includegraphics[width=0.23\textwidth]{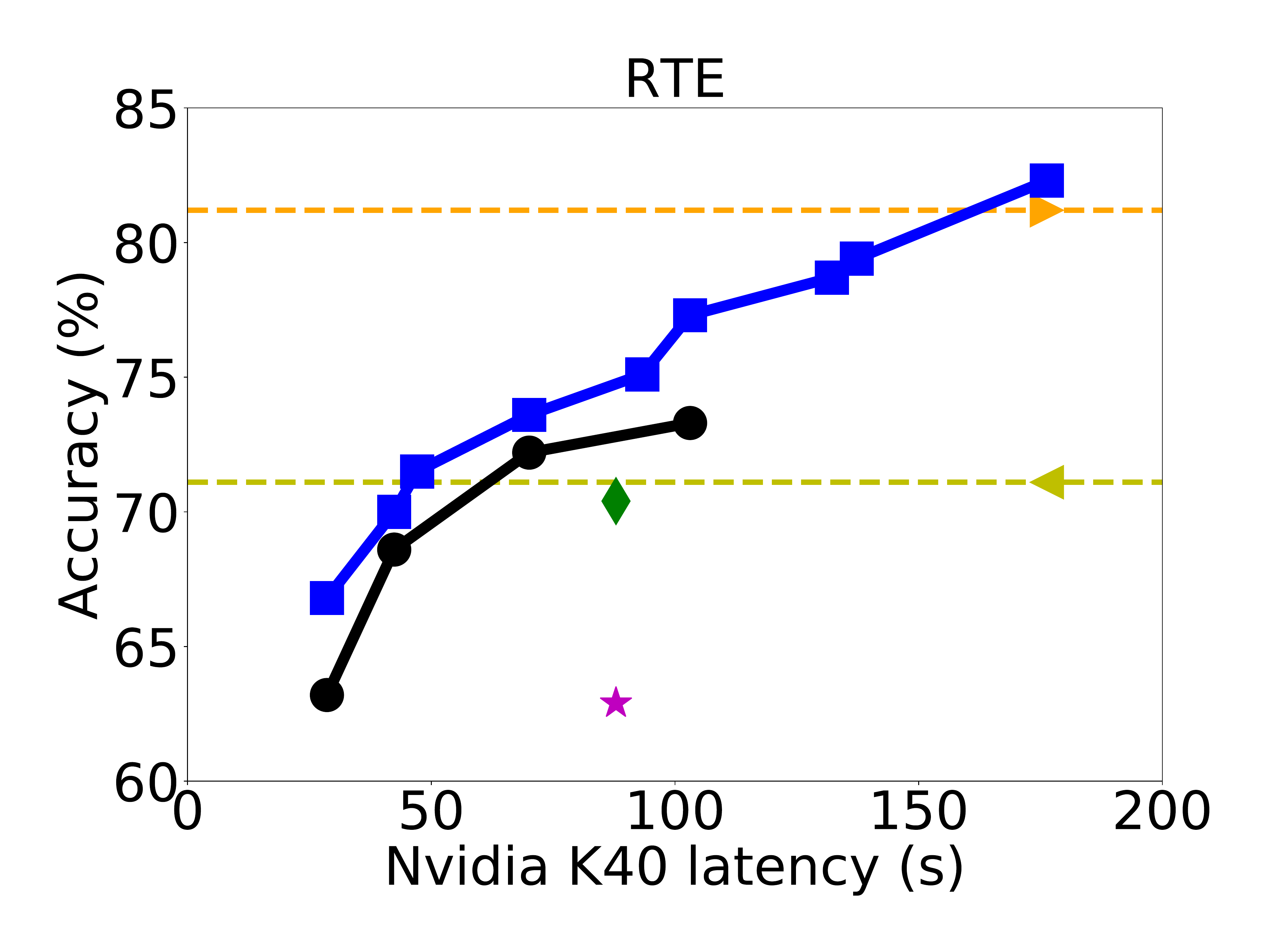}
	}
	\vspace{-0.15in}
	\\
	\subfloat[Nvidia K40 GPU latency(s).\label{fig:cola_gpu}]
	{
		\includegraphics[width=0.23\textwidth]{figures/comparison/mnli_val_gpu.pdf}
		\includegraphics[width=0.23\textwidth]{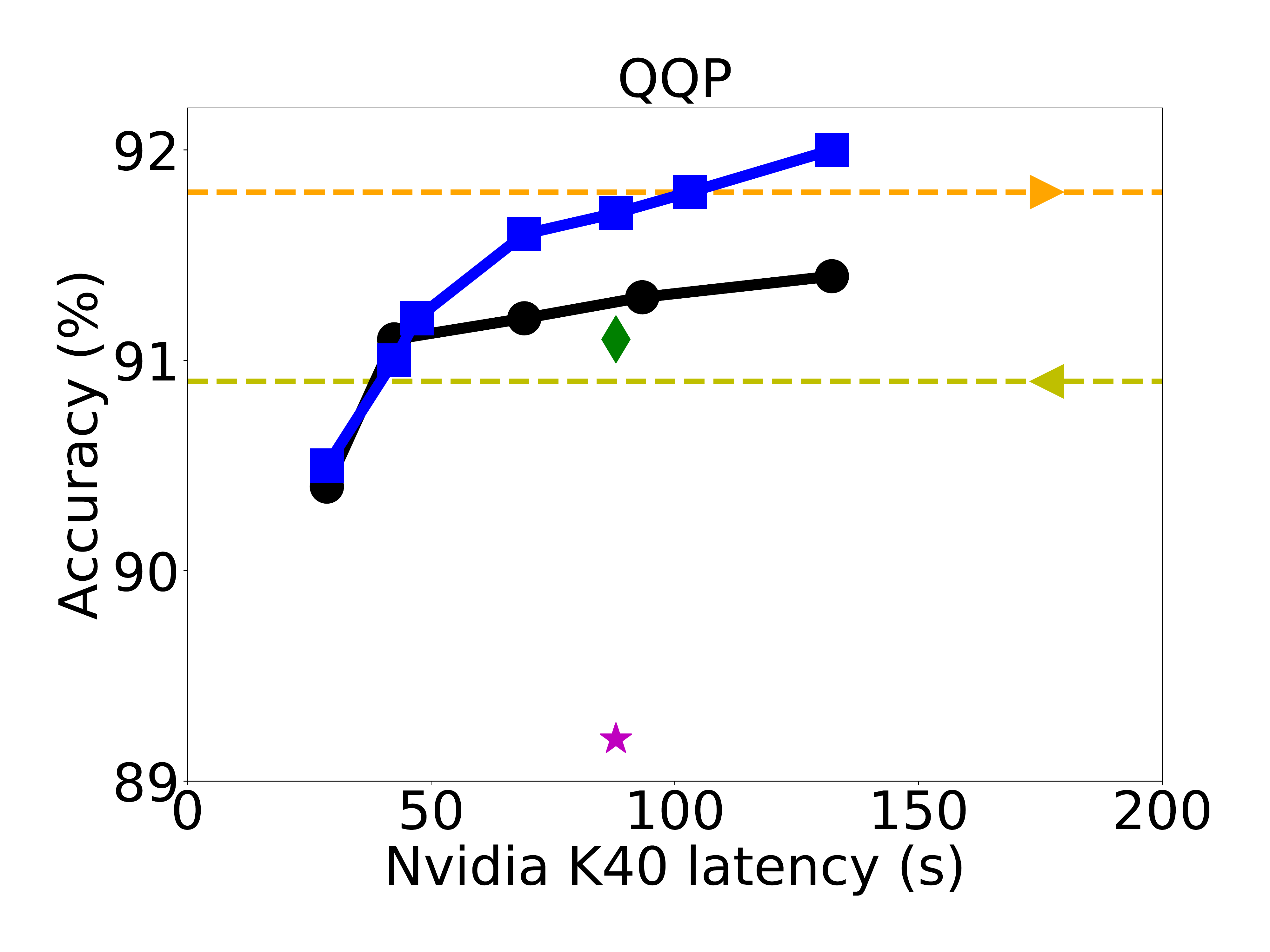}
		\includegraphics[width=0.23\textwidth]{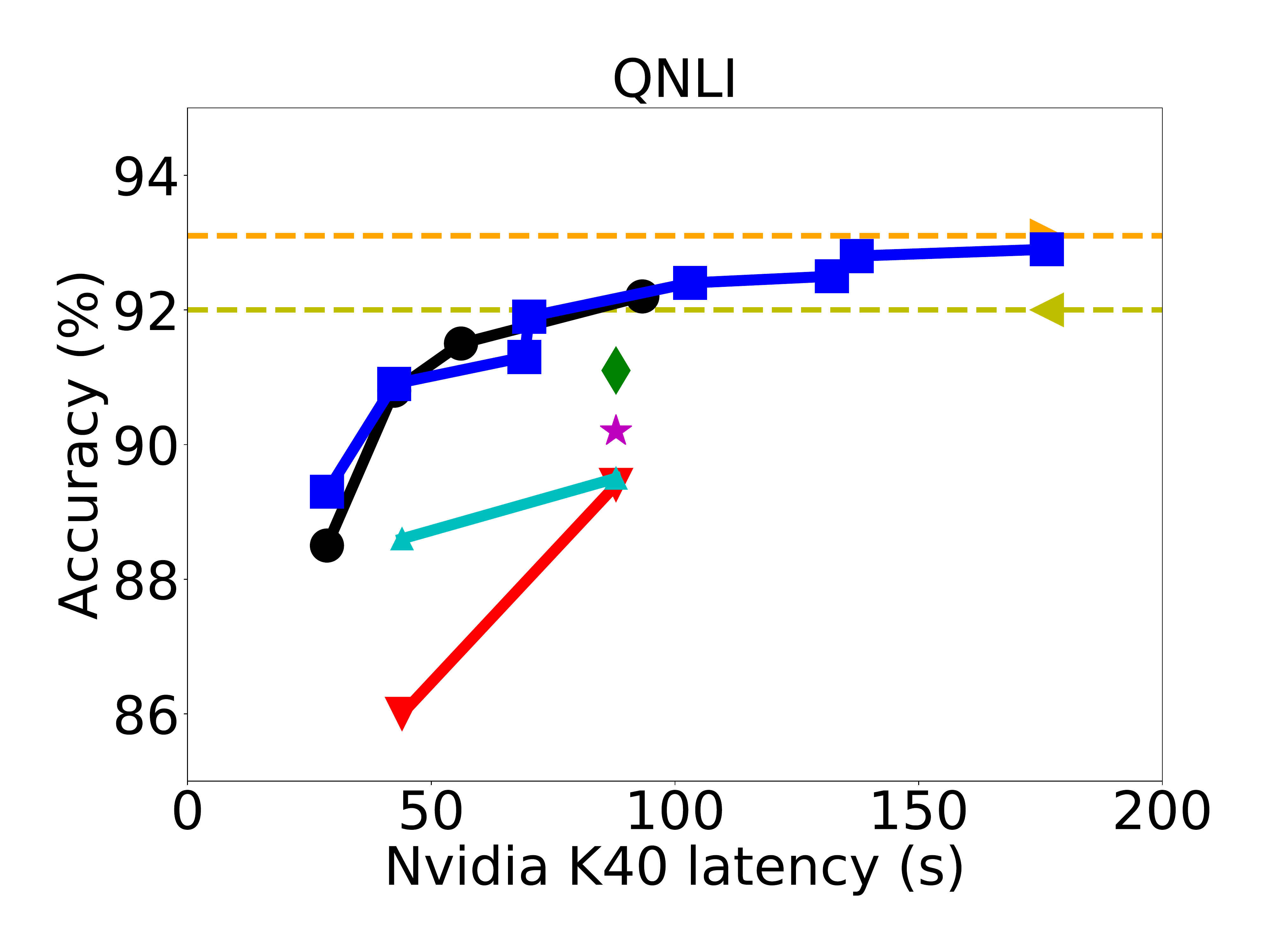}
		\includegraphics[width=0.23\textwidth]{figures/comparison/sst2_val_gpu.pdf}
	}
	\vspace{-0.15in}
	\addtocounter{subfigure}{-1}
	\\
	\subfloat{
		\includegraphics[width=0.23\textwidth]{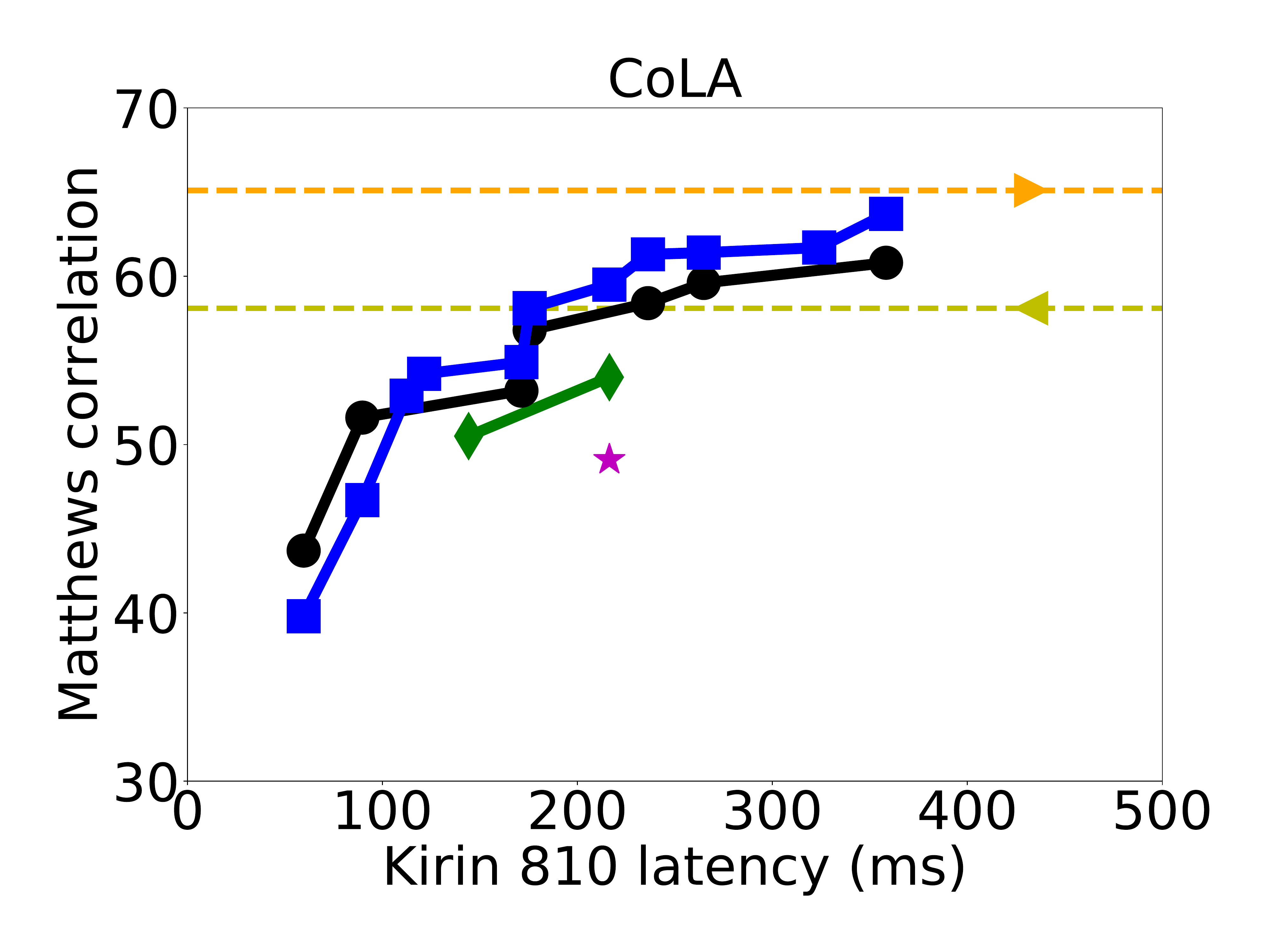}
		\includegraphics[width=0.23\textwidth]{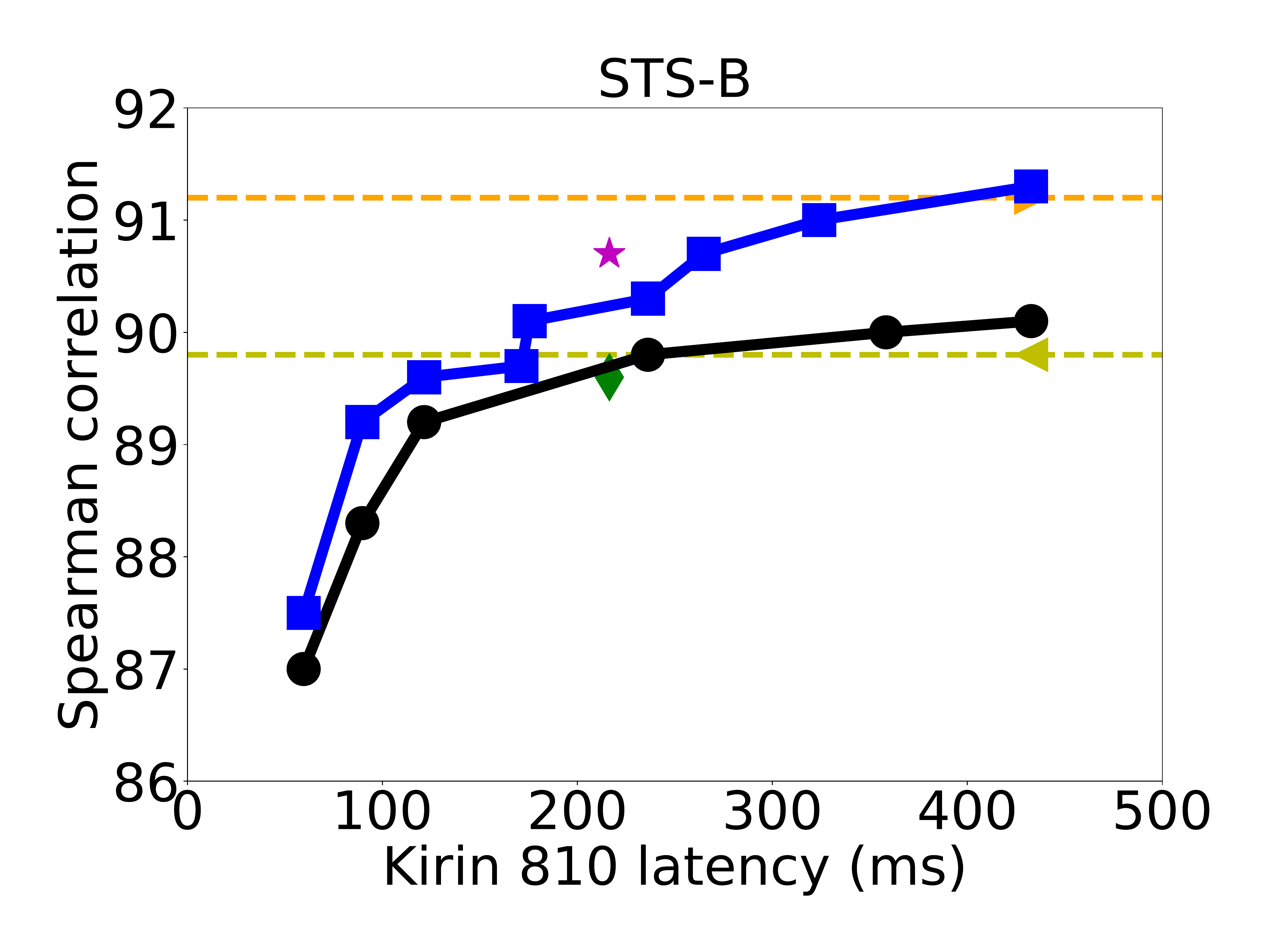}
		\includegraphics[width=0.23\textwidth]{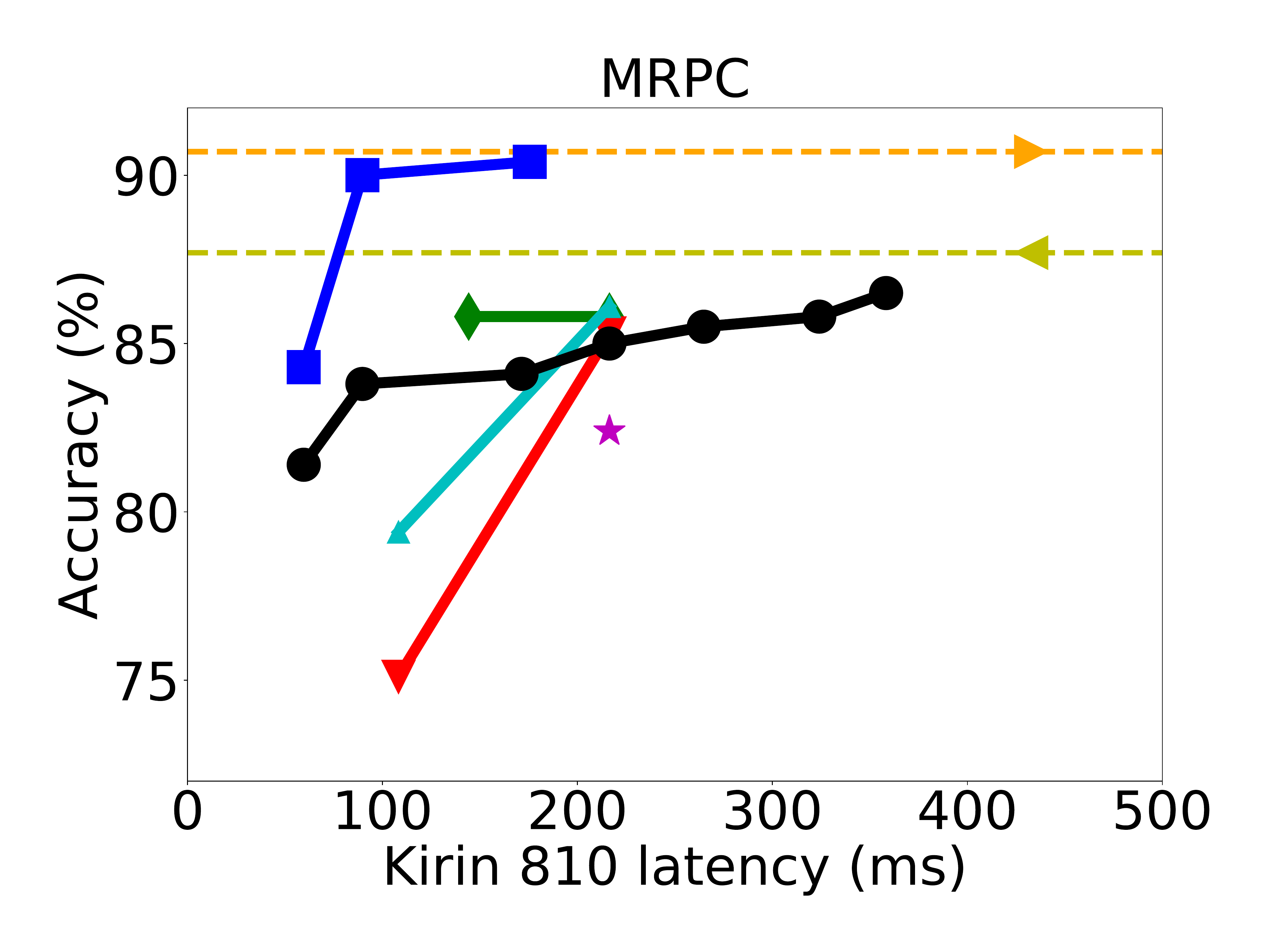}
		\includegraphics[width=0.23\textwidth]{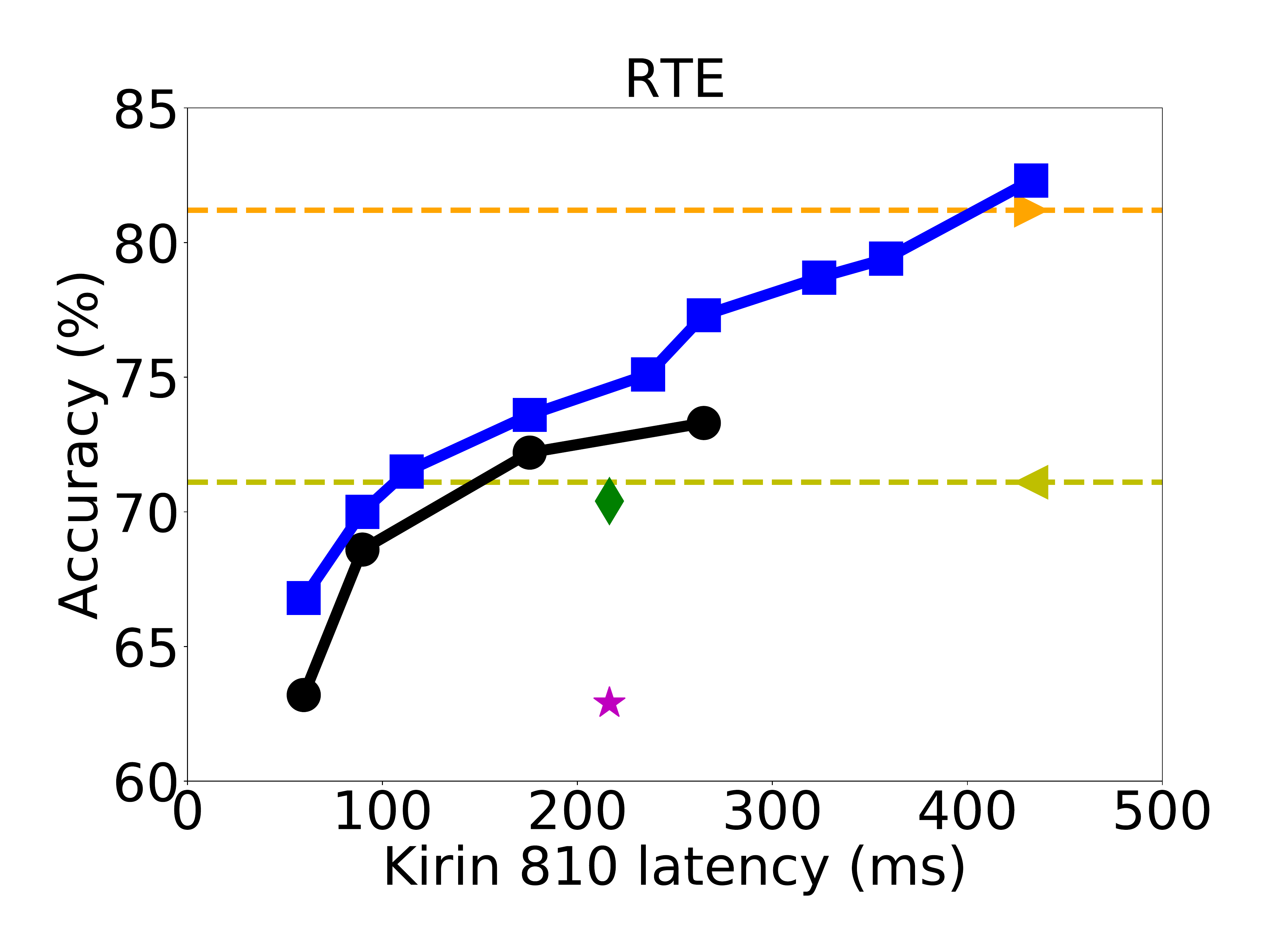}
	}
	\vspace{-0.15in}
	\\
	\subfloat[Kirin 810 ARM CPU latency(ms).\label{fig:cola_cpu}]
	{
		\includegraphics[width=0.23\textwidth]{figures/comparison/mnli_val_cpu.pdf}
		\includegraphics[width=0.23\textwidth]{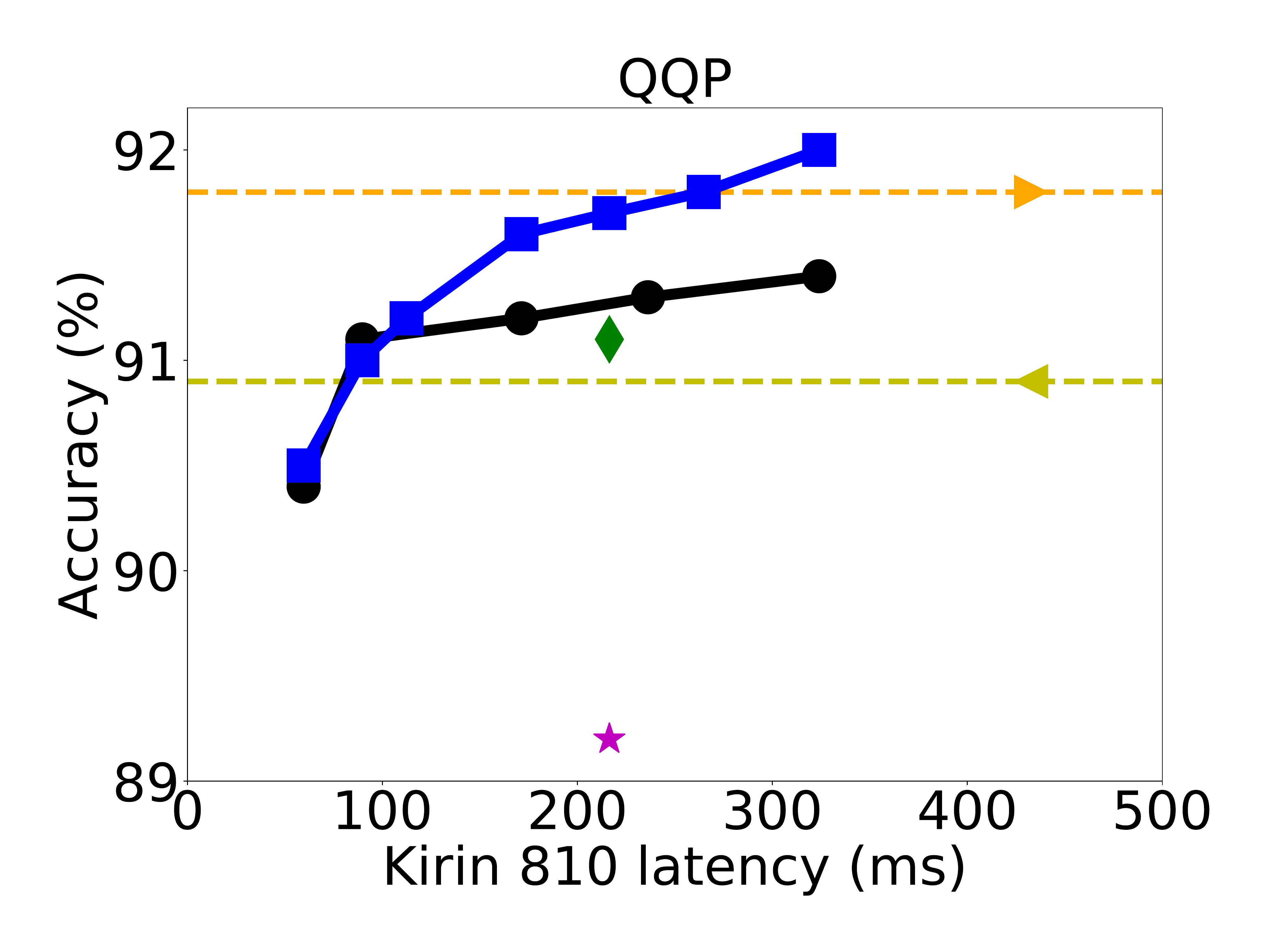}
		\includegraphics[width=0.23\textwidth]{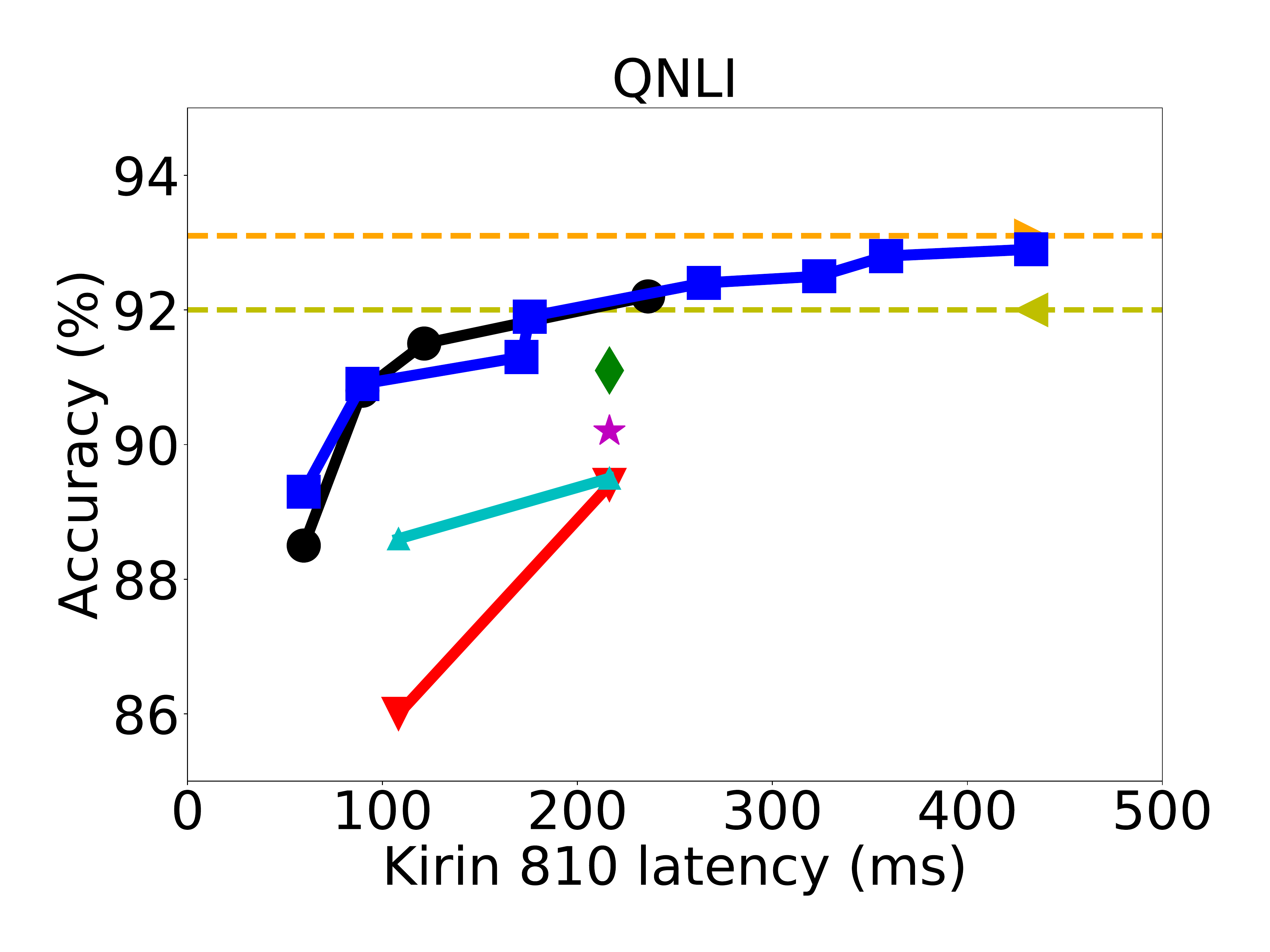}
		\includegraphics[width=0.23\textwidth]{figures/comparison/sst2_val_cpu.pdf}
	}
	\caption{Comparison of \#parameters(G), FLOPs(G), Nvidia K40 GPU latency(s) and Kirin 810 ARM CPU latency(ms)
		between our proposed DynaBERT and DynaRoBERTa and other methods on the GLUE benchmark. Average accuracy of \texttt{MNLI-m} and \texttt{MNLI-mm} is plotted.}
	\label{fig:comp_all}
\end{figure}

\paragraph{Comparison with Other Methods on All GLUE Tasks.}
Figure~\ref{fig:comp_all} shows the comparison of our proposed DynaBERT and DynaRoBERTa with other compression methods on all GLUE tasks, under different efficiency constraints, including \#parameters, FLOPs, latency on Nvidia K40 GPU and Kirin 810 A76 ARM CPU.

As can be seen, on all tasks,
the proposed DynaBERT and DynaRoBERTa  achieve comparable accuracy  as $\text{BERT}_\text{BASE}$ and $\text{RoBERTa}_\text{BASE}$, but often require fewer 
parameters, FLOPs or lower latency.
Similar to the observations in Section~\ref{expt:main},
under the same efficiency constraint, sub-networks extracted from our proposed DynaBERT outperform DistilBERT on all data sets except \texttt{STS-B} under \#parameters,
and outperforms TinyBERT on all data sets except \texttt{MRPC};
Sub-networks extracted from DynaRoBERTa outperform LayerDrop and even LayerDrop trained with much more data.

%

\subsection{Full Results of Ablation Study}
\label{apdx:ablation}
\paragraph{Training  $\text{DynaBERT}_\text{W}$ with Adaptive Width.}
Table~\ref{tbl:width_all} shows the accuracy for each width multiplier in the ablation study in 
the training of  $\text{DynaBERT}_\text{W}$.
As can be seen,  
$\text{DynaBERT}_\text{W}$ performs similarly as the separate network baseline at its largest width
and significantly better at smaller widths. The smaller the width, the more significant the accuracy gain.
From Table~\ref{tbl:width_all},  
after network rewiring, 
the average accuracy is over 2 points higher than the counterpart without rewiring. 
The accuracy gain is larger when the width of the model is smaller.

\begin{table}[h!]
	\vspace{-0.1in}
	\caption{Ablation study in the training of $\text{DynaBERT}_\text{W}$.
		Results on the development set are reported.
		The highest average accuracy of four width multipliers is highlighted.}
	\label{tbl:width_all}
	\centering
	\scalebox{0.74}{
		\begin{tabular}{ll|ccccccccc|c}
			\hline
			& $m_w$ & \texttt{MNLI-m} & \texttt{MNLI-mm} & \texttt{QQP}  & \texttt{QNLI} & \texttt{SST-2} & \texttt{CoLA} & \texttt{STS-B} & \texttt{MRPC} & \texttt{RTE}  & avg. \\ \hline
			& 1.0x  &      84.8       &       84.9       &     90.9      &      92.0       &      92.9      &     58.1      &      89.8      &     87.7      &     71.1      & 83.6 \\
			& 0.75x &      84.2       &       84.1       &     90.6      &     89.7      &      92.9      &     48.0      &      87.2      &     82.8      &     66.1      & 80.6 \\
			Separate 	network        & 0.5x  &      81.7       &       81.7       &     89.7      &      86       &      91.4      &     37.2      &      84.5      &     75.5      &     55.2      & 75.9 \\
			& 0.25x &      77.9       &       77.9       &     89.9      &     83.7      &      86.7      &     14.7      &      77.4      &     71.3      &     57.4      & 70.8 \\ \cline{2-12}
			& avg.  &      82.2       &       82.2       &     90.3      &     87.8      &      91.0      &     39.9      &      84.6      &     78.8      &     61.6      & 77.6 \\ \hline\hline
			& 1.0x  &      84.5       &       85.1       &     91.3      &     91.7      &      92.9      &     58.1      &      89.9      &     83.3      &     69.3      & 82.9 \\
			& 0.75x &      83.5       &        84.0        &     91.1      &     90.1      &      91.7      &     54.5      &      88.7      &     82.6      &     65.7      & 81.3 \\
			Vanilla  $\text{DynaBERT}_\text{W}$ 	& 0.5x  &      82.1       &       82.3       &     90.7      &     88.9      &      91.6      &     46.9      &      87.3      &     83.1      &      61       & 79.3 \\
			& 0.25x &      78.6       &       78.4       &     89.1      &     85.6      &      88.5      &     16.4      &      83.5      &     72.8      &     60.6      & 72.6 \\ \cline{2-12}
			& avg.  &      82.2       &       82.5       &     90.6      &     89.1      &      91.2      &     44.0      &      87.4      &     80.5      &     64.2      & 79.0 \\ \hline
			& 1.0x  &      84.9       &       84.9       &     91.4      &     91.6      &      91.9      &     56.3      &      90.0      &     84.6      &     70.0      & 82.8 \\
			& 0.75x &      84.3       &       84.2       &     91.3      &     91.7      &      92.4      &     56.4      &      89.9      &     86.0      &     71.1      & 83.0 \\
			\quad + Network rewiring          & 0.5x  &      82.9       &       82.9       &     91.0      &     90.6      &      91.9      &     47.7      &      89.2      &     84.1      &     71.5      & 81.3 \\
			& 0.25x &      80.4       &       80.0       &     90.0      &     87.8      &      90.4      &     45.1      &      87.3      &     80.4      &     66.0      & 78.6 \\ \cline{2-12}
			& avg.  &      83.1       &       83.0       &90.9 &     90.4      &      91.7      &     51.4      &      89.1      &     83.8      & \textbf{69.7} & 81.4 \\ \hline
			& 1.0x  &     85.1      &    85.4         &   91.1      &     92.5      &      92.9      &      59.0      &       90.0      &      86.0       &      70.0       & 83.5 \\
			& 0.75x &    84.9        &  85.6        &   91.1     &     92.4      &      93.1      &     57.9      &       90.0      &      87.0      &     70.8      & 83.6 \\
			\quad + Distillation and DA	& 0.5x  &   84.4        &    84.9        &   91.0     &     92.3      &      93.0      &     56.7      &      89.9      &     87.3      &     71.5      & 83.4 \\
			& 0.25x &       83.4      &    83.8      &    90.6    &     91.2      &      91.7      &     49.9      &       89.0       &     84.1      &     65.7      & 81.0 \\ \cline{2-12}
			& avg.  &  \textbf{84.5}  &  \textbf{84.9}   & \textbf{91.0}     & \textbf{92.1} & \textbf{92.7}  & \textbf{55.9} & \textbf{89.7}  & \textbf{86.1} &     69.5      & 82.9 \\ \hline
		\end{tabular}
	}
\end{table}

\paragraph{Training  $\text{DynaBERT}$ with Adaptive Width and Depth.}
Table~\ref{tbl:width_depth_all} shows the accuracy for each width and depth multiplier in the ablation study in 
the training of  $\text{DynaBERT}$.

\begin{table}[h!]
	\vspace{-0.1in}
	\caption{Ablation study in the training of DynaBERT. Results on the development set are reported.
		The highest average accuracy of four width multipliers for each depth multiplier is highlighted.}
	\label{tbl:width_depth_all}
	\centering
	\scalebox{0.75}{
		\begin{tabular}{ll|ccc|ccc|ccc|ccc}
			\hline
			&              &      \multicolumn{3}{c|}{\texttt{SST-2}}      &      \multicolumn{3}{c|}{\texttt{CoLA}}       &      \multicolumn{3}{c|}{\texttt{MRPC}}       &       \multicolumn{3}{c}{\texttt{RTE}}        \\ \hline
			& \diagbox[width=1.5cm,height=0.5cm]{$m_w$}{$m_d$}&     1.0x      &     0.75x     &     0.5x      &     1.0x      &     0.75x     &     0.5x      &     1.0x      &     0.75x     &     0.5x      &     1.0x      &     0.75x     &     0.5x      \\ \cline{2-14}
			& 1.0x         &     92.0      &     91.6      &     90.9      &     58.5      &     57.7      &     42.9      &     85.3      &     83.8      &     78.4      &     67.9      &     66.8      &     66.4      \\
			Vanilla  DynaBERT          & 0.75x        &     92.3      &     91.6      &     91.1      &     57.9      &     56.4      &     42.4      &     86.0      &     83.1      &     78.7      &     69.0      &     66.8      &     63.9      \\
			& 0.5x         &     91.9      &     91.9      &     90.6      &     55.9      &     53.3      &     40.6      &     86.0      &     83.1      &     79.7      &     68.2      &     65.0      &     63.9      \\
			& 0.25x        &     91.6      &     91.3      &     89.0      &     52.0      &     50.0      &     27.6      &     83.1      &     80.4      &     77.5      &     65.3      &     63.5      &     60.3      \\ \cline{2-14}
			& avg.         &     92.0      &     91.6      &     90.4      &     56.1      &     54.4      &     38.4      &     85.1      &     82.6      &     78.6      &     67.6      &     65.5      &     63.6      \\ \hline
			&\diagbox[width=1.5cm,height=0.5cm]{$m_w$}{$m_d$} &     1.0x      &     0.75x     &     0.5x      &     1.0x      &     0.75x     &     0.5x      &     1.0x      &     0.75x     &     0.5x      &     1.0x      &     0.75x     &     0.5x      \\ \cline{2-14}
			& 1.0x         &     92.9      &     93.3      &     92.7      &     57.1      &     56.7      &     52.6      &     86.3      &     85.8      &     85.0      &     72.2      &     70.4      &     66.1      \\
			\quad   + Distillation and & 0.75x        &     93.1      &     93.1      &     92.1      &     57.7      &     55.4      &     51.9      &     86.5      &     85.5      &     84.1      &     72.6      &     72.2      &     64.6      \\
			\quad Data augmentation    & 0.5x         &     92.9      &     92.1      &     91.3      &     54.1      &     53.7      &     47.5      &     84.8      &     84.1      &     83.1      &     72.9      &     72.6      &     66.1      \\
			& 0.25x        &     92.5      &     91.7      &     91.6      &     50.7      &     51.0      &     44.6      &     83.8      &     83.8      &     81.4      &     67.5      &     67.9      &     62.5      \\ \cline{2-14}
			& avg.         &     92.9      &     92.6      &     91.9      &     54.9      &     54.2      &     49.2      & \textbf{85.4} & \textbf{84.8} & \textbf{83.4} & \textbf{71.3} &     70.8      &     64.8      \\ \hline
			& \diagbox[width=1.5cm,height=0.5cm]{$m_w$}{$m_d$} &     1.0x      &     0.75x     &     0.5x      &     1.0x      &     0.75x     &     0.5x      &     1.0x      &     0.75x     &     0.5x      &     1.0x      &     0.75x     &     0.5x      \\ \cline{2-14}
			& 1.0x         &     93.2      &     93.3      &     92.7      &     59.7      &     59.1      &     54.6      &     84.1      &     83.6      &     82.6      &     72.2      &     71.8      &     66.1      \\
			\quad + Fine-tuning        & 0.75x        &     93.0      &     93.1      &     92.8      &     60.8      &     59.6      &     53.2      &     84.8      &     83.6      &     82.8      &     71.8      &     73.3      &     65.7      \\
			& 0.5x         &     93.3      &     92.7      &     91.6      &     58.4      &     56.8      &     48.5      &     83.6      &     83.3      &     82.6      &     72.2      &     72.2      &     67.9      \\
			& 0.25x        &     92.8      &     92.0      &     92.0      &     50.9      &     51.6      &     43.7      &     82.6      &     83.6      &     81.1      &     68.6      &     68.6      &     63.2      \\ \cline{2-14}
			& avg.         & \textbf{93.1} & \textbf{92.8} & \textbf{92.3} & \textbf{57.5} & \textbf{56.8} & \textbf{50.0} &     83.8      &     83.5      &     82.3      &     71.2      & \textbf{71.5} & \textbf{65.7} \\ \hline
		\end{tabular}
	}
\end{table}

\subsection{Full Results of Different Methods to Train  $\text{DynaBERT}_\text{W}$}
\label{apdx:pr_us}
\paragraph{Progressive Rewiring.}     
Instead of rewiring the network only once before training, ``progressive rewiring'' 
progressively rewires the network as more width multipliers are supported throughout the training.
Specifically, for four width multipliers $[1.0, 0.75,0.5,0.25]$, progressive rewiring first sorts the attention heads and neurons and rewires the corresponding connections before training to support width multipliers $[1.0,0.75]$. Then the attention heads and neurons are sorted and the network is rewired again before supporting $[1.0,0.75,0.5]$.
Finally, the network is again sorted and  rewired before supporting all four width multipliers.
For ``progressive rewiring'', we tune the initial learning rate from $\{2e-5, 1e-5, 2e-5, 5e-6, 2e-6 \}$  and pick the best-performing
initial learning rate $1e-5$.
Table~\ref{tbl:rewire} shows the development set accuracy on the GLUE benchmark for using progressive rewiring.
Since progressive rewiring requires progressive training and is time-consuming, we do not use data augmentation and distillation.
We use cross-entropy loss between predicted labels and the ground-truth labels as the training loss.
By comparing with Table~\ref{tbl:width} in Section~\ref{expt:ablation}, using progressive rewiring has no significant gain over rewiring only once.

\begin{table}[h!]
	\vspace{-0.1in}
	\centering
	\caption{Training $\text{DynaBERT}_\text{W}$ using progressive rewiring (PR).  }
	\label{tbl:rewire}
	\scalebox{0.9}{
		\begin{tabular}{l|ccccccccc|c}
			\hline
			$m_w$  &     \texttt{MNLI-m}       &      \texttt{MNLI-mm}       &      \texttt{QQP}      &     \texttt{QNLI}      &     \texttt{SST-2}     &    \texttt{CoLA}      &     \texttt{STS-B}     &     \texttt{MRPC}      &      \texttt{RTE}      & avg. \\ \hline
			1.0x   &  84.6  &  84.5   & 91.5 & 91.6 & 92.4  & 57.4 & 90.1  & 86.5 &  70.0  & 83.2    \\
			0.75x         &   83.6   &   84.0   &  91.2   & 91.4 & 91.7 & 56.6  & 89.7 & 84.8  & 70.8 & 82.6   \\
			0.5x          &   82.5   &  82.9  &   91.0    & 90.8 & 91.9 & 52.2  & 89.1 & 84.1  & 72.9 & 81.9   \\
			0.25x         &   78.3   &  79.7  &  89.9   & 87.9 & 90.4 & 45.1  & 87.6 & 82.4  & 67.5 & 78.8  \\ \hline
			avg.          &   82.3   &  82.8  &  90.9   & 90.4 & 91.6 & 52.8  & 89.1 & 84.5  & 70.3 & 81.6   \\ \hline
		\end{tabular}
	}
\end{table}

\paragraph{Universally Slimmable Training.}
Instead of using a pre-defined list of width multipliers, universally slimmable training  \cite{yu2019universally}
samples several width multipliers in each training iteration.
Following \cite{yu2019universally}, we also use inplace distillation for universally slimmable training.
For universally slimmable  training, we tune $(\lambda_1, \lambda_2)$ in $\{(1,1), (1,0), (0,1), (1, 0.1), (0.1,1), (0.1, 0.1)\}$ on \texttt{MRPC} and choose the best-performing one $(\lambda_1, \lambda_2)= (0.1, 0.1)$.
The corresponding results for can be found in Table~\ref{tbl:universal}.
For better comparison with using pre-defined width multipliers, we also report results when the width multipliers are $[1.0, 0.75,0.5,0.25]$.
We  do not use data augmentation here.
By comparing with Table~\ref{tbl:width} in Section~\ref{expt:ablation}, there is no significant difference between  using universally slimmable training and the  alternative training as used in Algorithm~\ref{alg:adaptive}.
\begin{table}[h!]
	\centering
	\caption{Training $\text{DynaBERT}_\text{W}$ using universally slimmable training (US).  }
	\label{tbl:universal}
	\scalebox{0.9}{
		\begin{tabular}{l|ccccccccc|c}
			\hline
			$m_w$        & \texttt{MNLI-m} & \texttt{MNLI-mm} & \texttt{QQP} & \texttt{QNLI} & \texttt{SST-2} & \texttt{CoLA} & \texttt{STS-B} & \texttt{MRPC} & \texttt{RTE} & avg. \\ \hline
			1.0x   &      84.6       &        85.0        &     91.2     &     91.7      &      92.4      &      59.7      &      90.0      &     85.3      &     69.0     & 83.2 \\
			0.75x   &      84.0       &       84.5       &     91.1     &     91.3      &      92.5      &      56.7      &      90.0      &     85.3      &     70.4     & 82.9 \\
			0.5x   &      82.2       &       82.6       &     90.7     &     90.5      &      91.1      &      52.1      &      89.2      &     85.3      &     71.5     & 81.7 \\
			0.25x   &      79.7       &       79.5       &     89.3     &     87.5      &      90.1      &      36.4      &      87.3      &     79.4      &     67.5     & 77.4 \\ \hline
			avg.   &      82.6       &       82.9       &     90.6     &     90.3      &      91.5      &      51.2      &      89.1      &     83.8      &     69.6     & 81.3 \\ \hline
		\end{tabular}
	}
\end{table}

\subsection{Looking into DynaBERT}
\label{apdx:attention_vis}

\begin{figure}[htbp]
	\centering
	\vspace{-0.1in}
	\subfloat[``the cat sat on the mat.''\label{fig:attn_1}
	]{
		\includegraphics[height=0.55\textwidth, width=0.75\textwidth]{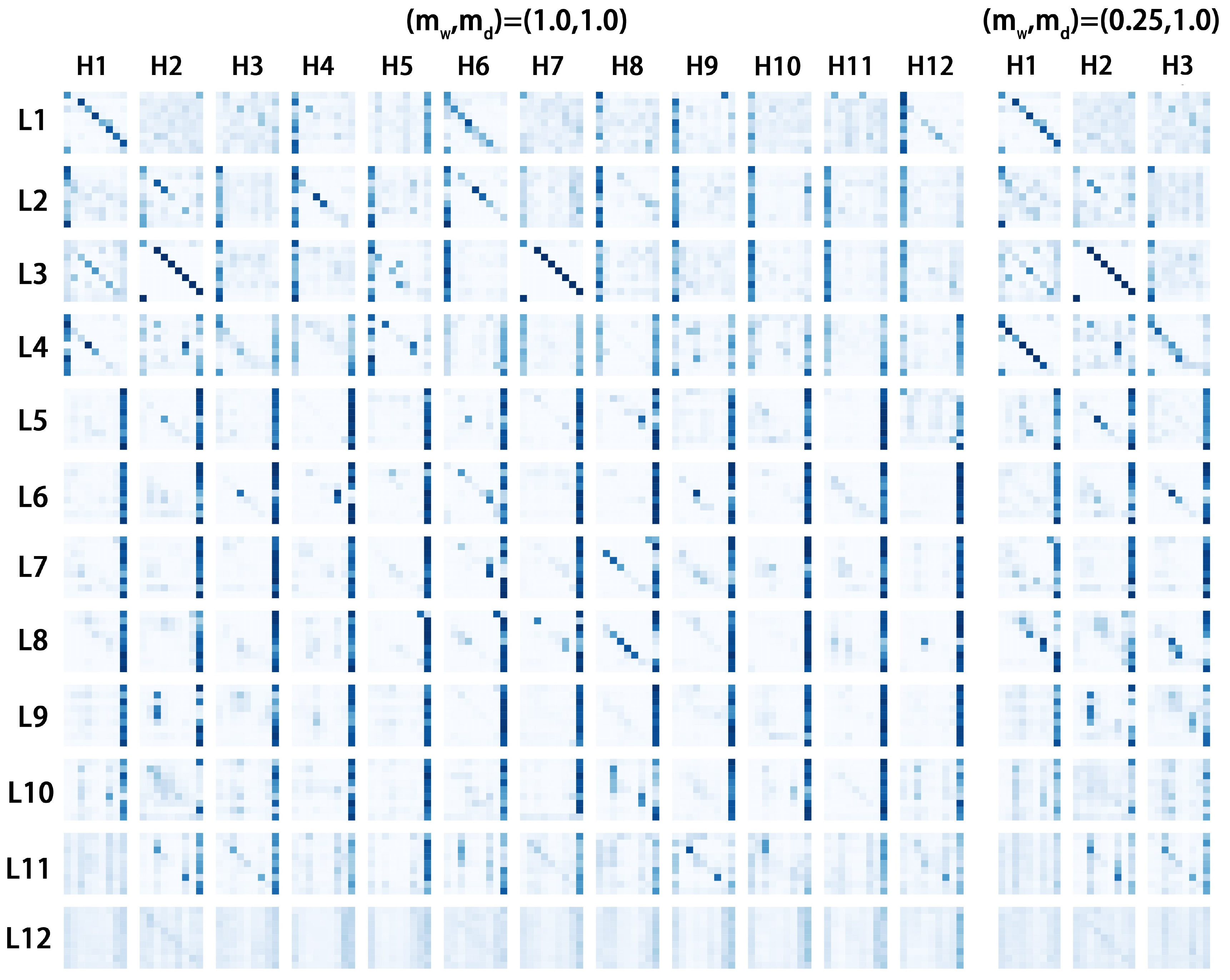}
	}
	\vspace{-0.1in}
	\subfloat[``.mat the on sat cat the''
	]{
		\includegraphics[height=0.55\textwidth, width=0.75\textwidth]{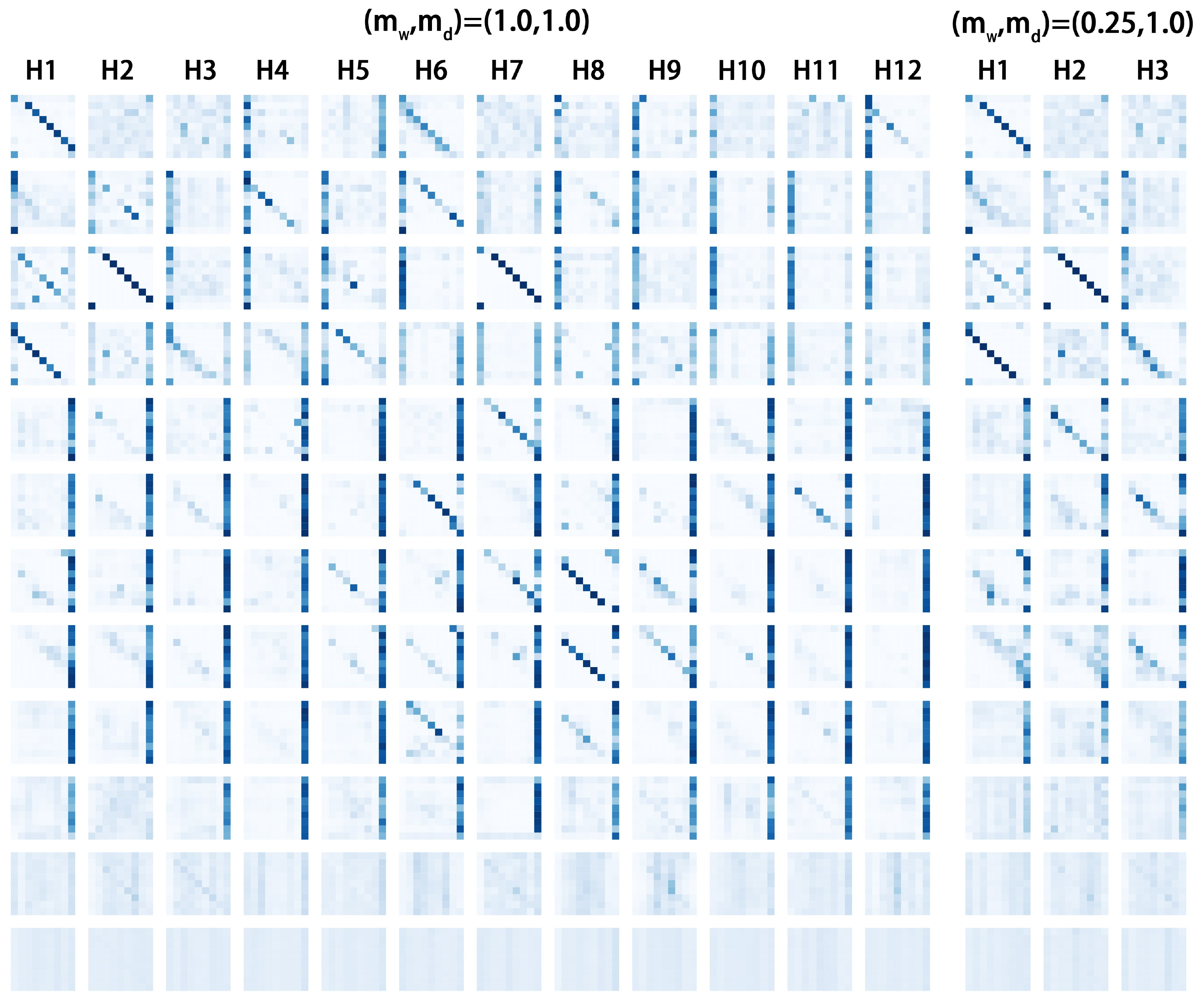}
	}
	\vspace{-0.05in}
	\caption{Attention maps in sub-networks with different widths in DynaBERT 
		trained on \texttt{CoLA}. 
	}
	\vspace{-0.1in}
	\label{fig:cola}
\end{figure}

\begin{figure}[htbp]
	\centering
	\subfloat[``the cat sat on the mat.'']{
		\includegraphics[height=0.6\textwidth, width=0.6\textwidth]{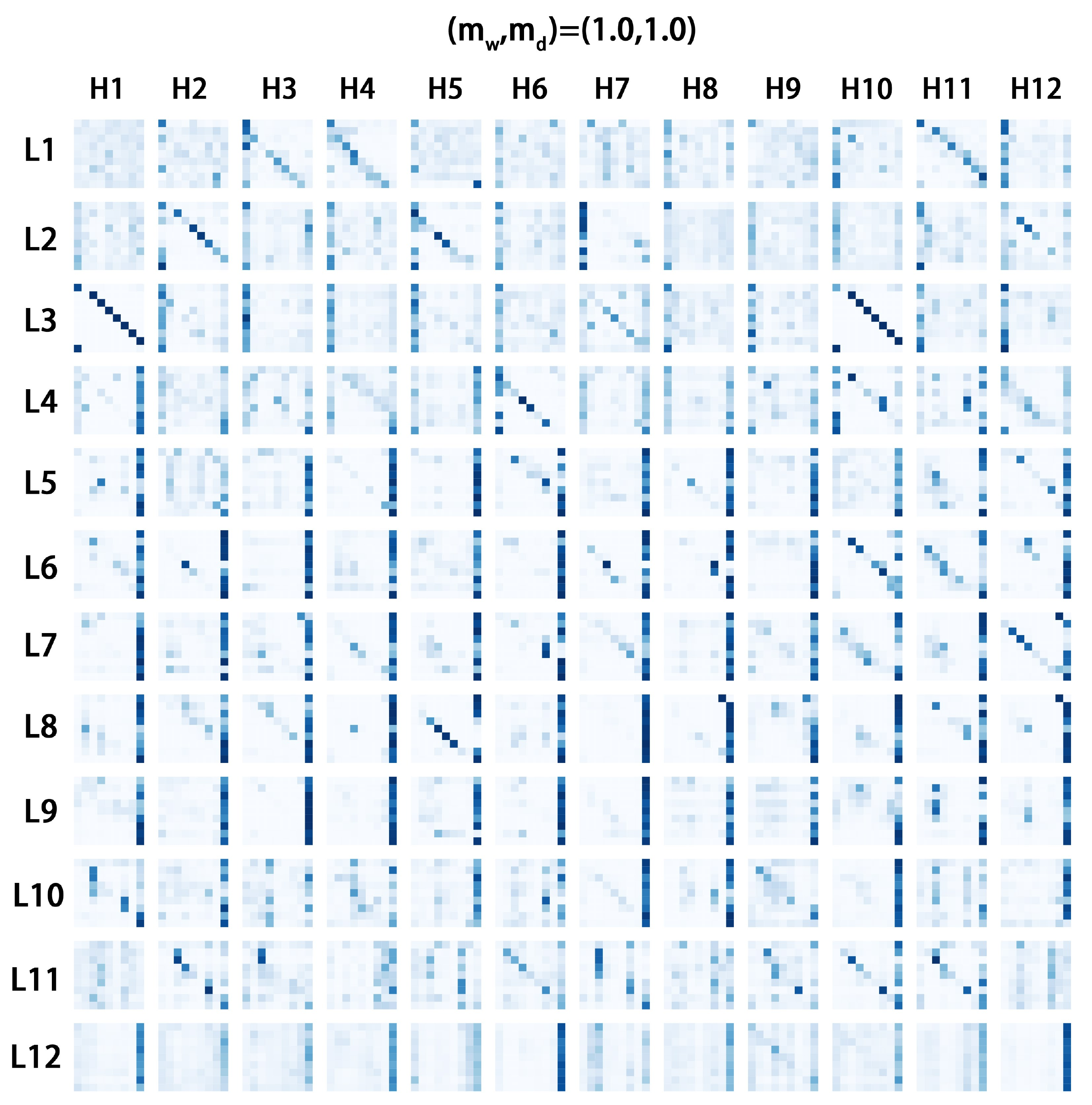}}
	\\
	\subfloat[``.mat the on sat cat the'']{
		\includegraphics[height=0.6\textwidth, width=0.6\textwidth]{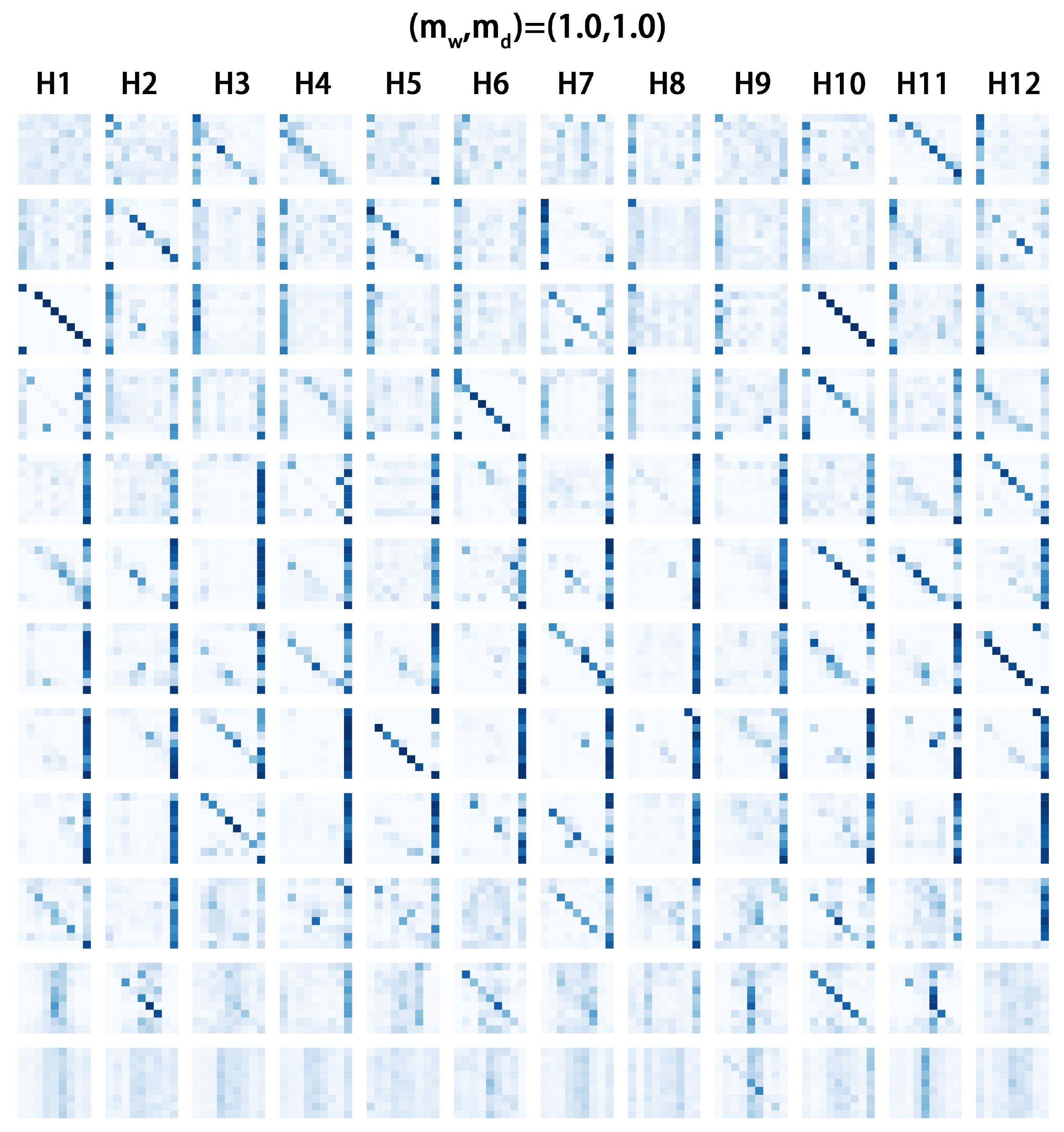}}
	\vspace{-0.05in}
	\caption{Attention maps in $\text{BERT}_\text{BASE}$
		fine-tuned on \texttt{CoLA}. 
	}
	\label{fig:cola_base}
\end{figure}

\paragraph{CoLA.}
\texttt{CoLA} is abbreviated for the ``Corpus of Linguistic Acceptability'' and  is a binary single-sentence classification task, where the goal is to predict whether an English sentence is linguistically “acceptable”.
Figure~\ref{fig:cola}  shows the attention maps of the learned DynaBERT with two different width multipliers $m_w=1.0$ and $0.25$. 
We use both a linguistically acceptable sentence ``the cat sat on the mat.'' and a non-acceptable one ``.mat the on sat cat the'' whose words are in the reverse order.
As can be seen,  in the last two Transformer layers of DynaBERT of both widths,
for the linguistic non-acceptable sentence, the attention heads do not encode useful information, with each word attending to every other word with almost equal probability. 
Figure~\ref{fig:cola_base} shows the attention maps obtained by $\text{BERT}_\text{BASE}$
fine-tuned on \texttt{CoLA}, with the same linguistic acceptable and non-acceptable sentence as in Figure~\ref{fig:cola}. As can be seen, unlike DynaBERT,
the attention maps in the final two layers still show positional or syntactic patterns.
This observation reveals the enhanced ability of
the proposed DynaBERT in distinguishing linguistic acceptable and non-acceptable sentences.
Similar observations are also found in other samples in \texttt{CoLA} data set.

\begin{figure}[htbp]
	\centering
	\subfloat[``a smile on your face.'']{
		\includegraphics[height=0.55\textwidth, width=0.75\textwidth]{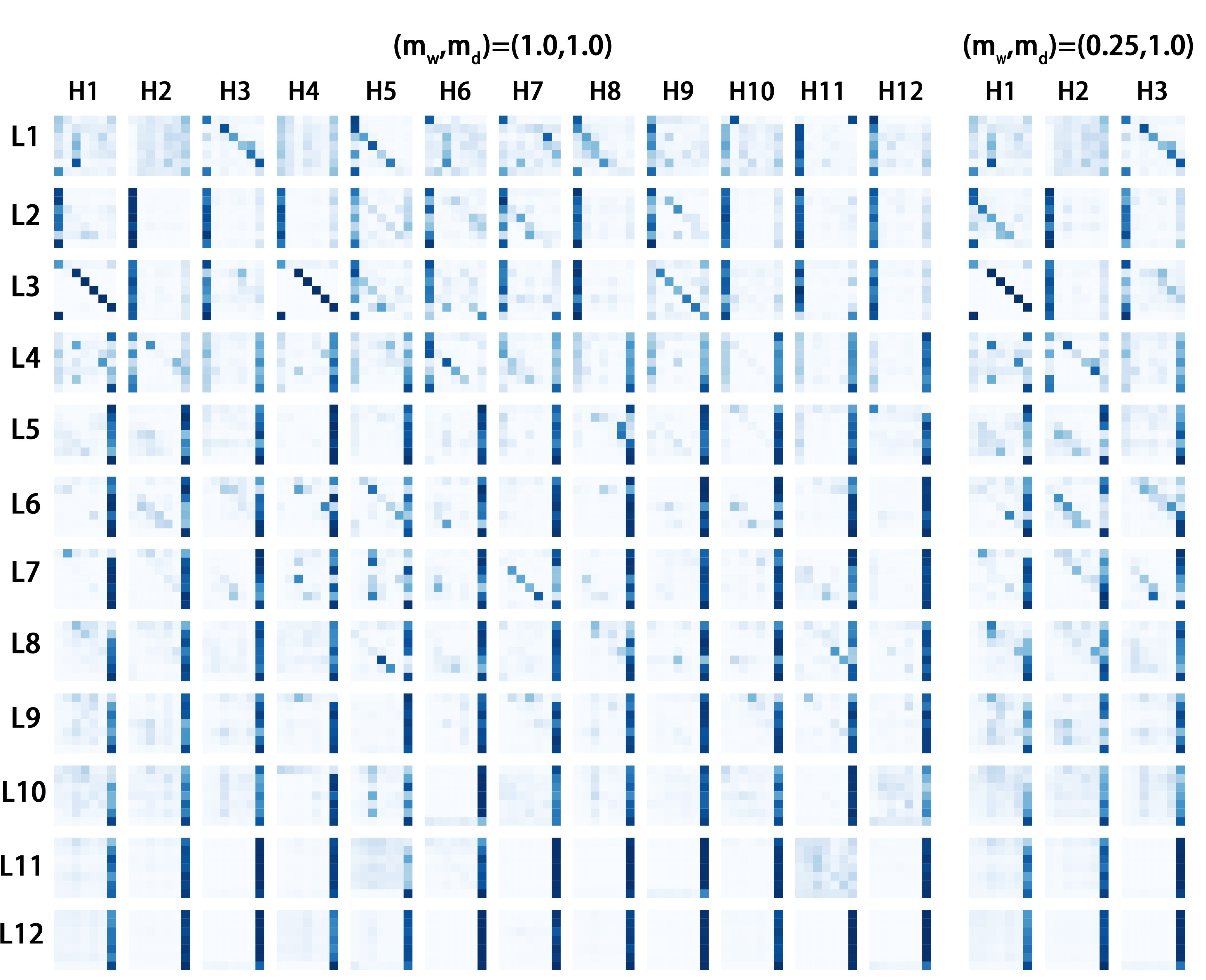}}
	\hfill
	\subfloat[``an extremely unpleasant film.'']{
		\includegraphics[height=0.55\textwidth, width=0.75\textwidth]{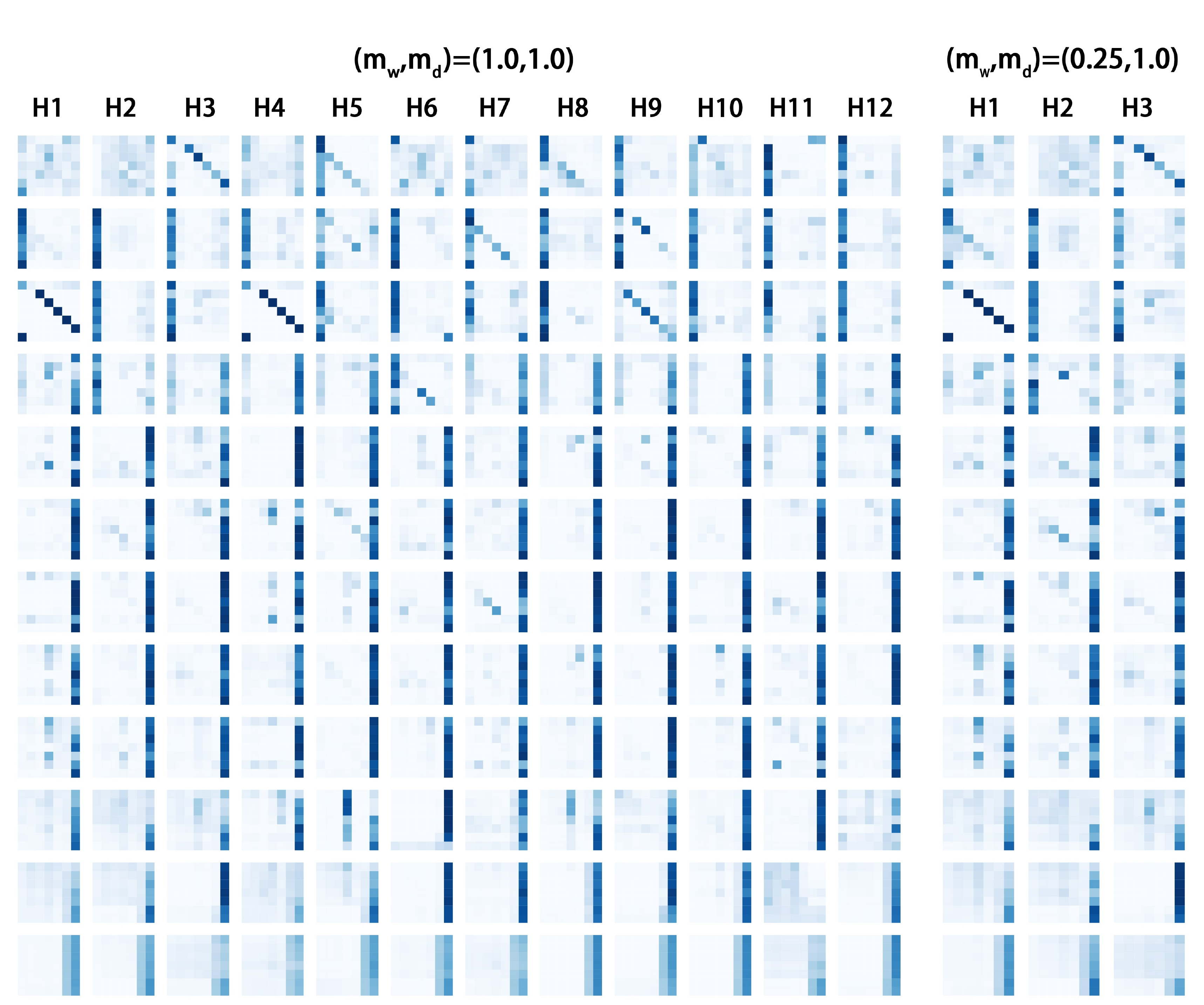}}
	\caption{Attention maps in sub-networks with different widths in DynaBERT 
		trained on \texttt{SST-2}. 
	}
	\label{fig:sst2}
\end{figure}

\paragraph{SST-2.}
\texttt{SST-2} (the Stanford Sentiment Treebank) is a binary single-sentence classification task consisting of sentences extracted from movie reviews with human annotations of their sentiment.
Figure~\ref{fig:sst2} shows the attention maps obtained by DynaBERT with  annotations of both positive and negative sentiment. The sentence with positive sentiment is ``a smile on your face.''. The sentence with negative sentiment is ``an extremely unpleasant film .''.
As can be seen, for both $m_w=1$ and $0.25$, most  attention maps in the final few layers point  to the last token ``[SEP]'', which is not used in the downstream task. This indicates that there is redundancy in the Transformer layers.
This  is also consistent with the 
finding  in Section~\ref{expt:glue}  that, even when the depth multiplier is only $m_d=0.5$ (i.e., 6 Transformer layers), the model has only less than 1 point of accuracy degradation for both widths.

\section{Related Work on the Capacity of Language Models}
There are also related works that study the relationship between the  capacity and performance of language models. 
It is shown in \cite{kovaleva2019revealing,rogers2020primer} that  considerable redundancy and over-parametrization exists in BERT models. 
In \cite{jawahar2019does}, it is shown that the BERT's  layers encode a hierarchy of linguistic information, with surface features at the bottom, syntactic features in the middle and semantic features at the top. 
The capacity of other language models besides BERT like character CNN and recurrent networks
are also studied in \cite{jozefowicz2016exploring,melis2018state}.
In \cite{subramani2019can}, it is shown that pre-trained language models with moderate sized representations are able to recover arbitrary sentences.

\section{Preliminary Results of Applying DynaBERT in the Pretraining Phase }
In this section, we use the proposed method for pre-training a BERT with adaptive width and depth.
We use a pre-trained 6-layer BERT downloaded from the official Google BERT  repository \url{https://github.com/google-research/bert} as the backbone model.
To make sub-networks of DynaBERT  the same size as those small models,  for width, we also adapt the hidden state size $H=128,256,512,768$ besides attention heads and intermediate layer neurons. For depth, we adjust the number of layers to be $L=4,6$. 
Distillation loss over the hidden states in the last layer is used as the training objective.
The number of training epochs is 5.
After pre-training DynaBERT, we fine-tune each separate sub-network with the original task-specific data on MNLI-m and report the development set results in Table~\ref{tbl:pretrain}.
We compare with separately pre-trained small models in Google BERT repository.
As can be seen, sub-networks of the pre-trained DynaBERT outperform separately pre-trained small networks.
\begin{table}[htbp]
	\vspace{-0.12in}
	\caption{Development set accuracy on MNLI-m of separately pre-trained BERT  models and sub-networks of a pre-trained DynaBERT.}
	\label{tbl:pretrain}
	\centering
	\scalebox{0.83}{
		\begin{tabular}{l|c|c|c|c|c|c|c|c}
			\hline
			(L, H)                   & (6, 768) & (6, 512) & (6, 256) & (6, 128) & (4, 768) & (4, 512) & (4, 256) & (4, 128) \\ \hline
			Separate small networks &    81.8     &   80.3      &  76.0       &      72.4   &  80.1       &  78.6       &   74.9      &    70.7        \\ \hline
			Sub-networks of DynaBERT    &    82.0     &    81.0     &   77.8      &   73.0      &      81.5   &   80.4      &  76.1       &     71.4        \\ \hline
		\end{tabular}
	}
	\vspace{-0.12in}
\end{table}

\end{document}